\documentclass[10pt,journal,compsoc]{IEEEtran}
\usepackage{setspace}
\usepackage{amssymb}
\usepackage{graphicx}
\usepackage{subfigure}
\usepackage{mathrsfs}
\usepackage{amsmath}
\usepackage{url}
\usepackage{algorithm}
\usepackage{algorithmic}
\usepackage{color}
\usepackage{tikz}
\usepackage{booktabs}
\usepackage{comment}
\usepackage{subeqnarray}

\DeclareMathOperator*{\argmin}{arg\,min}

\newcommand{\eat}[1]{}
\newcommand{\tensor}[1]{\boldsymbol{\mathcal{#1}}}

\newcommand{\revision}{\color{black}}

\ifCLASSOPTIONcompsoc
  \usepackage[nocompress]{cite}
\else
  \usepackage{cite}
\fi
\begin{document}

\title{Understanding Urban Dynamics via Context-aware Tensor Factorization with Neighboring Regularization}

\author{
Jingyuan~Wang, %~\IEEEmembership{Member,~IEEE,}
        Junjie~Wu, %~\IEEEmembership{Member,~IEEE,}
        Ze~Wang, %~\IEEEmembership{Member,~IEEE,}
        Fei~Gao,
        and~Zhang~Xiong
        % <-this % stops a space
\IEEEcompsocitemizethanks{
	\IEEEcompsocthanksitem J. Wang, Z. Wang and Z. Xioing are with the School of Computer Science and Engineering, Beihang Unversity, Beijing 100191, China. E-mail: \{jywang,ze.w,xiongz\}@buaa.edu.cn.
	\IEEEcompsocthanksitem J. Wu (corresponding author) is with the School of Economics and Management, Beihang University, Beijing 100191, China. E-mail: wujj@buaa.edu.cn.
	\IEEEcompsocthanksitem F. Gao is with Microsoft Research Asia, Beijing, China.
}% <-this % stops a space
%\thanks{Manuscript received XXX, 2017; revised XXX, 2014.}
}


%\markboth{IEEE Transactions on Knowledge and Data Engineering,~Vol.~xx, No.~x, xxx~xxxx}%
%{Shell \MakeLowercase{\textit{et al.}}: Bare Advanced Demo of IEEEtran.cls for Journals}

\IEEEtitleabstractindextext{%
\begin{abstract}
Recent years have witnessed the world-wide emergence of mega-metropolises with incredibly huge populations. Understanding residents¡¯ mobility patterns, or urban dynamics, thus becomes crucial for building modern smart cities. In this paper, we propose a Neighbor-Regularized and context-aware Non-negative Tensor Factorization model (NR-cNTF) to discover interpretable urban dynamics from urban heterogeneous data. Different from many existing studies concerned with prediction tasks via tensor completion, NR-cNTF focuses on gaining urban managerial insights from spatial, temporal, and spatio-temporal patterns. This is enabled by high-quality Tucker factorizations regularized by both POI-based urban contexts and geographically neighboring relations. NR-cNTF is also capable of unveiling long-term evolutions of urban dynamics via a pipeline initialization approach. We apply NR-cNTF to a real-life data set containing rich taxi GPS trajectories and POI records of Beijing. The results indicate: 1) NR-cNTF accurately captures four kinds of city rhythms and seventeen spatial communities; 2) the rapid development of Beijing, epitomized by the CBD area, indeed intensifies the job-housing imbalance; 3) the southern areas with recent government investments have shown more healthy development tendency. Finally, NR-cNTF is compared with some baselines on traffic prediction, which further justifies the importance of urban contexts awareness and neighboring regulations.

%Huge urban populations bring traffic jams, job-housing imbalance, and other challenges to metropolises. Understanding mobility patterns of residents in a city, {\em i.e. Understanding Urban Dynamics}, is very essential from urban planning and policy making. In this paper, we proposed a Neighboring Regularized and context-aware Non-negative Tensor Factorization (NR-cNTF) model to discover explainable urban dynamics from multi-source data. Specifically, we formulated the urban dynamics mining as a probabilistic non-negative tensor factorization, and then introduced a context-aware item to merge urban context data with the tensor factorization. Moreover, a neighboring regularization is introduced in the model optimization to incorporate geographical neighboring relations into our model, and a pipeline initialization approach are adopted to capture long-term evolution of urban dynamics. The NR-cNTF model could discover urban dynamics from  four aspects: spatial patterns, temporal patterns, spatio-temporal interactions, and long-term evolutions. All dynamics discovered by the model have definite statistical meanings and therefore very easy to understand. We applied the NR-cNTF model over a real-life data set that contains GPS trajectories of more than 20,000 taxis and 400 thousands POI records of Beijing. From the urban dynamics learn by the model, we can got many managerial insights for urban planning. We also compare our methods with some baselines on traffic prediction, which justifies the modeling of urban context and neighboring regulation in NR-cNTF.
\end{abstract}

}

\maketitle

% is not selected <OR> if conference mode is selected - because compsoc
% conference papers position the abstract like regular (non-compsoc)
\IEEEdisplaynontitleabstractindextext

% \fi
\IEEEpeerreviewmaketitle

\ifCLASSOPTIONcompsoc
\IEEEraisesectionheading{\section{Introduction}\label{sec:introduction}}
\else
\section{Introduction}
\label{sec:introduction}
\fi

As reported by the World Bank\footnote{\url{http://data.worldbank.org/}}, {\revision at the end of 2016 more than 53\% population of the world, {\it i.e.}, about 3.7 billion people, lived in cities; about 36 mega-metropolises worldwide had a population of more than 10 million.} Huge urban populations bring great challenges such as traffic jams, {\revision educational/medical resource scarcity}, environmental pollution, {\it etc}. Understanding the behavioral patterns of residents in a city, or {\it urban dynamics} for short, therefore becomes an important yet urgent demand for urban planning and public policy making from a smart city perspective. Fortunately, the widely adopted {\it mobile crowd sensing} (MCS) technologies~\cite{mcs}, such as GPS, mobile phones, and location-based services, {\revision give us an unprecedented opportunity to access to enormous and perhaps unbounded human mobility data}, {\revision which combined with urban infrastructure data offer a ``rich ore'' for discovery of urban dynamics.}

In general, mining urban dynamics from MCS data has three requirements. The first one is to model {\it multi-source heterogeneous data}, which consist of mobility records of residents such as the origins and destinations, the travel time, the purposes, and the surroundings hidden in different data sources such as GPS trajectories, urban contexts, and city maps. The second requirement is to capture {\it long-term evolutions}, which is critically important for urban planners to understand the evolving rules of cities so as to make proper urban planning. The last one is to find urban dynamics with {\it good interpretability} --- an obscure urban dynamic is useless {\revision to decision making in real-world application scenarios}. Despite of rich literature in applying matrix/tensor factorizations to model urban heterogeneous data, {\revision most of them aim to generate patterns to improve the predictive accuracy of traffic volumes~\cite{tensor_its,gas_consum,travel_time}, but leave pattern explanation to luck}. It is not until recently that a few works begin to take {\revision the understanding of urban dynamics as the primary research task}, and the representative ones include the earlier rNTD model using Tucker factorizations~\cite{apweb}, the city spectrum modeling using CP factorizations~\cite{cityspectrum}, {\revision and still some using single source data~\cite{PLOSONE_matrix,Kang2016Understanding,TRB} or for discovering urban functional zones only~\cite{planning,yuan2012discovering}}. These excellent works, however, cannot meet all the {\revision above-mentioned} requirements simultaneously.

%could satisfy above three requirements at the same time. For instance, city spectrum~\cite{cityspectrum} and works in~\cite{PLOSONE_matrix,Kang2016Understanding,TRB} use tensor and matrix factorizations to mine urban dynamics from signal human mobility data set. The method proposed in~\cite{planning,yuan2012discovering} could discover urban functional zones from multi-source data. The rNTD model~\cite{apweb} could discover short-term multi-aspects dynamics from multi-sources data. More importantly, very few works introduced special mechanisms to ensure the interpretability of discovered dynamics, which causes whether discovered patterns are explainable depends mainly on luck.

In this paper, we propose a {\bf N}eighbor-{\bf R}egularized {\bf c}ontext-aware {\bf N}on-negative {\bf T}ensor {\bf F}actorization model (NR-cNTF) to discover explainable and evolving urban dynamics from multi-source heterogeneous urban data. In the NR-cNTF model, we introduce the concepts of data space and pattern space and describe the relations between urban data and urban dynamics. The Tucker factorization is then introduced with the {\revision POI-based (Point-Of-Interests)} urban contexts to factorize the {\revision ODT (Origin-Destination-Time)} tensor into spatial, temporal, and spatio-temporal patterns of good interpretability. Moreover, a neighboring regularization that incorporates geographically neighboring relations is introduced into our model to further improve the explainability of spatial patterns. Finally, a simple yet effective pipeline initialization approach is designed to capture the long-term evolutions of urban dynamics.

We conduct extensive experiments on a real-life data set that contains the GPS trajectories of over 20,000 taxies and over {\revision 400,000} POI records of Beijing from 2008 to 2015. The first scenario of the experiments is to verify the ability of NR-cNTF in disclosing true urban dynamics and obtain managerial insights via NR-cNTF. The results indicate that: 1) NR-cNTF accurately captures four kinds of mobility rhythms and seventeen spatial communities of Beijing; 2) the rapid development of Beijing in the CBD area, is indeed at the expense of severer job-housing imbalance and therefore is unsustainable in a long run; 3) the southern areas of Beijing are experiencing unprecedented growth with the recent government investments, and most importantly they have shown more healthy development tendency. The second scenario of the experiments is to testify the prediction power of NR-cNTF, which is compared with some baselines on traffic prediction. The results demonstrate the superiority of NR-cNTF in tensor completion, which further justifies the importance of adopting urban contexts and neighboring regulations in NR-cNTF.

\section{Problem Formulation}
\label{sec:problem}
%{\color{blue} In this section, we formulate \emph{urban dynamics discovery} as a context-aware tensor factorization problem. We first give some math notation as follows.
%}

\begin{spacing}{1.1}
\begin{table}\caption{Notation Definition}\label{tab:notation_def}
%  \vspace{-0.2cm}
  \footnotesize
  \centering
  \begin{tabular}{c|c|l}
    \hline \hline
    \bf{Space} & \bf{Variable} &  \bf{Definition}  \\ \hline
     & $\tensor{R} $ &  the data tensor\\ \cline{2-3}
    Data      &  $r_{xyz}$ &  the $(x,y,z)$ element of $\tensor{R}$\\ \cline{2-3}
    Space & $\mathbf{W}$ & the urban context matrix\\ \cline{2-3}
     & ${w}_{pq}$ &  the $(p,q)$ element of $\mathbf{W}$\\ \hline
    & $\tensor{C}$ &  the pattern tensor\\ \cline{2-3}
    & $c_{ijk}$ &  the $(i,j,k)$ element of $\tensor{C}$\\ \cline{2-3}
    Pattern & $\mathbf{O,D,T}$ &  the pattern projection matrices\\ \cline{2-3}
    Space  & $\mathbf{o}_{x},\mathbf{d}_{x},\mathbf{t}_{x}$ & the $x$-th row vectors of $\mathbf{O,D,T}$\\ \cline{2-3}
    & $\mathbf{o}_{:i},\mathbf{d}_{:i},\mathbf{t}_{:i}$ & the $i$-th column vectors of $\mathbf{O,D,T}$\\ \cline{2-3}
     & $o_{xi},d_{xi},t_{xi}$ & the $(x,i)$ elements of $\mathbf{O,D,T}$\\
    \hline \hline
  \end{tabular}
%  \vspace{-0.2cm}
\end{table}
\end{spacing}

In this section, we formulate \emph{urban dynamics discovery} as a context-aware tensor factorization problem. Table~\ref{tab:notation_def} lists the math variables to be used, which are divided into two categories, \textit{i.e.}, data-space variables and pattern-space variables, according to their observability. Variables in the data space are observable from real-world human mobility, while variables in the pattern space are latent but crucial for understanding urban dynamics.

%data and used to describe urban dynamics in the original data space. In contrast, variables in pattern space are unobservable from real-world data and used to characterize urban dynamics as spatio-temporal patterns in the underlying pattern space. More details are given as follows.

Throughout the paper, we use lowercase symbols such as $a$, $b$ to denote scalars, bold lowercase symbols such as $\mathbf{a}$, $\mathbf{b}$ for vectors, bold uppercase symbols such as $\mathbf{A}$, $\mathbf{B}$ for matrices, and calligraphy symbols such as $\tensor{A}$, $\tensor{B}$ for tensors.

{\bf  Data-space variables:} The primary variable in data space is a {\it data tensor}. Assume there are $M$ urban zones in a city, and $N$ time slices in a day. Let $r_{xyz}$ denote the {\it resident travel intensity} from an origin zone $x~\in~\{1, \cdots, M\}$ to a destination zone $y~\in~\{1, \cdots, M\}$ within a time slice $z~\in~\{1, \cdots, N\}$. A third-order tensor $\tensor{R} \in \mathbb{R}^{M \times M \times N}$ is then defined by having $r_{xyz}$ as the $(x, y, z)$ element. Intuitively, $\tensor{R}$ contains the original information about urban dynamics, which can be obtained from urban vehicle and resident trajectory data. Another variable in data space is an {\it urban-context similarity matrix} $\mathbf{W} \in \mathbb{R}^{M \times M}$. The $(p, q)$ element of $\mathbf{W}$, {\it i.e.}, ${w}_{pq}$,  is a coefficient that describes the similarity between urban zones $p$ and $q$ using, \textit{e.g.}, points of interest (POI) data.

{\bf Pattern-space variables:} The variables in pattern space include a {\it core tensor} and three {\it pattern projection matrices}. Assume there are $I$ origin spatial patterns (OSP), $J$ destination spatial patterns (DSP), and $K$ temporal patterns (TP) hidden inside the data tensor $\tensor{R}$. We define $\mathbf{O}\in \mathbb{R}^{M \times I}$ as a spatial projection matrix that projects $M$ origin zones into $I$ OSP's. Similarly, $\mathbf{D} \in \mathbb{R}^{M \times J}$ is defined as another spatial projection matrix that projects $M$ destination zones into $J$ DSP's. The matrix $\mathbf{T}\in \mathbb{R}^{N \times K}$ is a temporal projection matrix that projects $N$ time slices to $K$ TP's. The elements of $\mathbf{O}$, $\mathbf{D}$ and $\mathbf{T}$ are denoted as $o_{xi}$, $d_{yj}$ and $t_{zk}$, respectively, indicating the projection intensities from the urban zones $x$, $y$ and time slice $z$ to OSP $i$, DSP $j$ and TP $k$, $1\leq i\leq I$, $1\leq j\leq J$, $1\leq k\leq K$. We define a third-order tensor $\tensor{C}$ as a core tensor that describes the dynamics of resident travels among temporal and spatial patterns. The $(i,j,k)$ element of $\tensor{C}$, {\it i.e.}, $c_{ijk}$, denotes the intensity of resident travels from OSP $i$ to DSP $j$ within TP $k$.

\subsection{Construction of Data Tensor}

We here explain how to construct the data tensor $\tensor{R}$ using real-life GPS trajectory data of Beijing Taxies. To this end, we first segment the Beijing city map  into $M$ urban zones. In the literature, quite a few methods including the grid based, morphology based, road networks based, and administrative boundaries based methods~\cite{zheng2011urban, yuan2012segmentation} can fulfill this task. Here we adopt a Traffic Analysis Zones (TAZ) map provided by Beijing Municipal Committee of Transport\footnote{http://www.bjjtw.gov.cn/} to segment Beijing into $M=651$ zones. Finally, since resident behaviors in city life are often cyclical every day, we divide one day into $N=24$ time slices (one hour per slice). The above procedure determines the three modes of $\tensor{R}$.

We then compute the element values of $\tensor{R}$. Note that the taxi GPS data are often organized as a set of quintuples in the form as $\langle vid,~time,~longitude,~latitude,~state\rangle$, where $vid$ is the unique ID of a taxi, $(longitude,~latitude)$ is the location of the taxi, and $state$ informs whether the taxi is carrying any passengers at time $time$. We first obtain all taxi-based passenger travels by removing the records with ``no passengers'' state. Then an {\em origin-destination-time} (ODT) record is constructed for each travel by picking up the first and last records of the travel and then extracting the origin and destination coordinates and the travel starting time. We collect the travel ODT records of all workdays in a month as a data set. The monthly total amount of travels that depart from TAZ $x$ in time slice $z$ and arrive at TAZ $y$ is recorded as $\tilde{r}_{xyz}$. As reported in~\cite{xuke}, the travel volumes between different urban zones usually follow a long-tail distribution. Therefore, we adopt the $\log$ function to rescale $\tilde{r}_{xyz}$ as
\begin{equation}\label{equ:rescal_org_tesnor}
    r_{xyz} = \log \left( 1 + \tilde{r}_{xyz} \right),
\end{equation}
which is finally used as the $(x,y,z)$ element of $\tensor{R}$.

\subsection{Definition of Pattern Tensor}
\label{subsec:pattern}

Variables in pattern space include $\tensor{C}$, $\mathbf{O}$, $\mathbf{D}$, and $\mathbf{T}$, where $\tensor{C}$ is the core tensor that models the dynamic relations among spatio-temporal patterns in the pattern space, and $\mathbf{O}$, $\mathbf{D}$ and $\mathbf{T}$ are the matrices that project the  data tensor $\tensor{R}$ into the core tensor $\tensor{C}$. To better understand this, we give formal definitions to the spatial and temporal patterns as follows.

{\bf Definition 1} (Spatial Pattern): A spatial pattern is a vector containing the membership score of each urban zone to this pattern. Assume there are $I$ spatial patterns and $M$ urban zones. The $i$th spatial pattern is denoted as a vector $\mathbf{v}_{:i} = (v_{1i}, \ldots, v_{Mi})^{\top}$, where $v_{mi}$ is the membership score of the $m$th zone to the $i$th spatial pattern. The spatial projection matrix $\mathbf{V}$ that projects $M$ urban zones to $I$ spatial patterns is then defined as $\mathbf{V} = [\mathbf{v}_{:1}, \ldots, \mathbf{v}_{:I}]$.~\hfill~$\blacksquare$

The $x$th row vector of $\mathbf{V}$, denoted as $\mathbf{v}_x$, is a vector that depicts the membership scores of urban zone $x$ to $I$ different spatial patterns. We assign $x$ to spatial pattern $i$ if $i\in\arg\max_{1\leq j\leq I} v_{xj}$. In this way, we can cluster all urban zones into the $I$ spatial patterns. %, as shown in Fig.~\ref{fig:urban_communities}.
This implies that a spatial pattern is essentially a \emph{spatial community} consisting of urban zones that function similarly in urban dynamics. For example, most of residents in a residential community leave in the morning and return in the evening. In contrast, for a business community, people arrive in the morning and leave in the evening. Spatial patterns can be further divided into origin spatial patterns (OSP) and destination spatial patterns (DSP). The projection matrix $\mathbf{V}$ is denoted as $\mathbf{O}$ for OSP's and $\mathbf{D}$ for DSP's for differentiation. While $\mathbf{O}$ and $\mathbf{D}$ share the same $M$ urban zones, they might have different numbers of spatial patterns.

\begin{comment}
\begin{figure}\centering
	\includegraphics[width=\columnwidth]{fig/urban_communities-eps-converted-to.pdf}\\
	\caption{Clustering urban zones as communities using the spatial pattern projection matrices.}\label{fig:urban_communities}
\end{figure}
\end{comment}

{\bf Definition 2} (Temporal Pattern): A temporal pattern is a vector containing the membership score of each time slice within a day to this pattern. Assume there are $K$ temporal patterns and $N$ time slices in a day. The $k$th temporal pattern is denoted as a vector $\mathbf{t}_{:k} = (t_{1k}, \ldots, t_{Nk})^\top$, where $t_{nk}$ is the membership score of the $n$th time slice to the $k$th temporal pattern. The temporal projection matrix $\mathbf{T}$ that projects $N$ times slices into $K$ temporal patterns is then defined as $\mathbf{T} = [\mathbf{t}_{:1}, \ldots, \mathbf{t}_{:K}]$.~\hfill~$\blacksquare$

%Fig.~\ref{fig:rhythm} illustrates how to use the projection matrix $\mathbf{T}$ to decompose the temporal intensity of $N$ time slices into $K$ temporal patterns.
In essence, a temporal pattern describes a \emph{temporal rhythm} of urban dynamics, which might correspond to an event that occurs recurrently everyday, {\it e.g.}, the morning peak and evening peak in a city. Accordingly, the vector $\mathbf{t}_{:k}$ indicates the dynamic intensity of the rhythm $k$ within a day. %Note that temporal rhythm is an important feature of urban dynamics. For example, in morning peaks of workdays, residents go to workplaces and then go back home in evening peaks.

%The temporal patterns and projection matrix can model the rhythm feature of urban dynamic.

%Besides spatial and temporal patterns% explained by projection matrices $\mathbf{O}$, $\mathbf{D}$, and $\mathbf{T}$,
Next, we define a pattern tensor to describe the interrelationships among spatio-temporal patterns.

\begin{comment}
\begin{figure}\centering
  \includegraphics[width=\columnwidth]{fig/travel_rhythms-eps-converted-to.pdf}\\
  \caption{Mapping time slices of a day as rhythms using the temporal pattern projection matrix.}\label{fig:rhythm}
\end{figure}
\end{comment}

{\bf Definition 3} (Pattern Tensor): A tensor $\boldsymbol{\mathcal{C}}\in \mathbb{R}^{I\times J \times K}$ is a third-order pattern tensor, if its $(i,j,k)$ element $c_{ijk}$ indicates the intensity of resident travels from OSP $i$ to DSP $j$ in TP $k$, $1\leq i\leq I, 1\leq j\leq J, 1\leq k\leq K$. ~\hfill~$\blacksquare$

Human behaviors in city life usually have synchronism, which can be described by urban dynamic patterns in $\boldsymbol{\mathcal{C}}$. For example, intuitively, residents living in a residential community commute to business regions synchronously in every morning peak of workdays. So an element $c_{ijk}$ has a high value when the origin spatial pattern $i$ corresponds to a residence community, the destination spatial pattern $j$ corresponds to a business community, and the temporal pattern $k$ corresponds to a morning-peak rhythm.

\subsection{Definition of Urban Context}

Travel behaviors of residents not only have relations with urban spatial and temporal patterns but also have close relations with the so-called {\it urban context}~\cite{yuan2012discovering,yuan2015discovering}. Urban context refers to the surroundings inside an urban zone that can affect the travel behaviors of that zone. One typical type of urban context is the so-called {\it points of interests} (POI) including residential buildings, office buildings, shopping malls, \textit{etc}. We have the following definition.

{\bf Definition 4} (Urban-Context Similarity Matrix): A matrix $\mathbf{W} \in \mathbb{R}^{M \times M}$ is called an urban-context similarity matrix, whose $(p,q)$ element $w_{pq}$ is a coefficient that measures the POI context similarity between zones $p$ and $q$, $1\leq p,q\leq M$.~\hfill~$\blacksquare$

In general, $\mathbf{W}$ is a nonnegative and symmetric matrix, which could be used to validate the effectiveness of the spatial patterns found purely from trajectory data. For example, it is intuitive that the travel patterns of urban zones with a mass of office buildings should be very similar, but differ sharply from that of zones filled with residential buildings.

%as important supplements to $\mathbf{O}, \mathbf{D}, \mathbf{T}$ and $\boldsymbol{\mathcal{C}}$ for the better characterization of urban dynamics.

\begin{figure}[t]\centering
	\begin{center}\centering
		\subfigure[Non-negative Tensor Factorization]{\label{fig:frameworkNTF}\includegraphics[width=0.85\columnwidth]{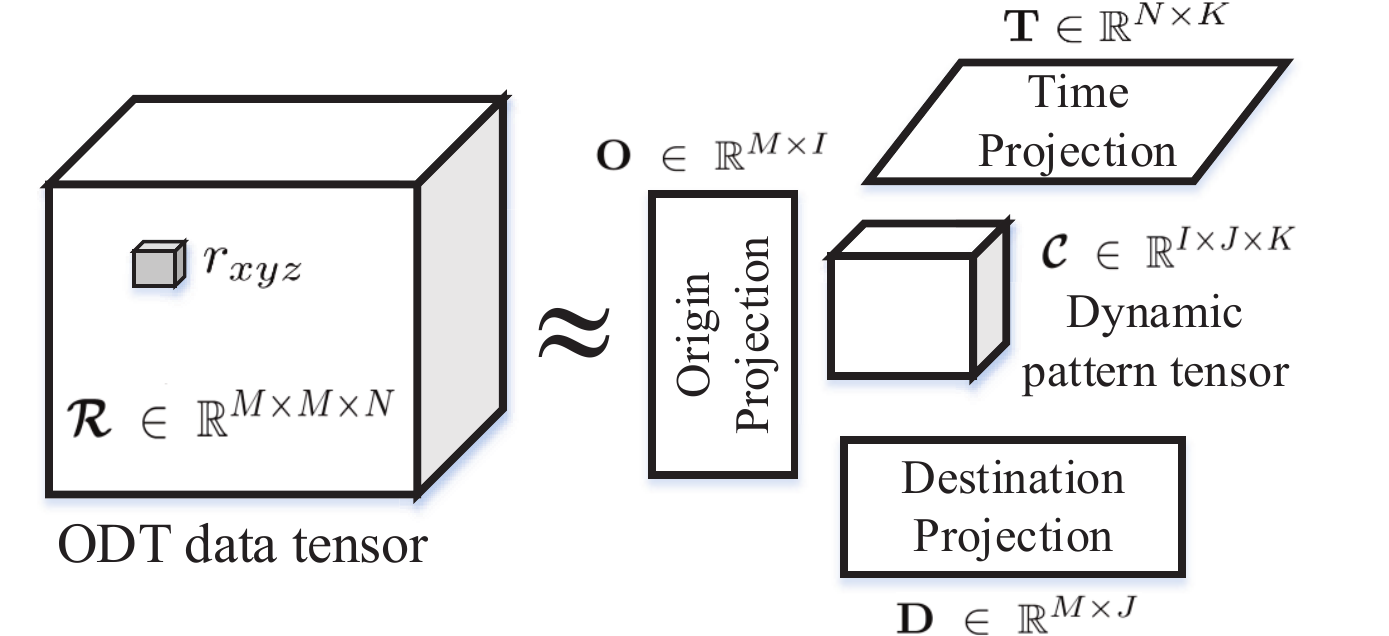}}
		\subfigure[Contexts Awareness ]{\label{fig:frameworkNMF}\includegraphics[width=0.85\columnwidth]{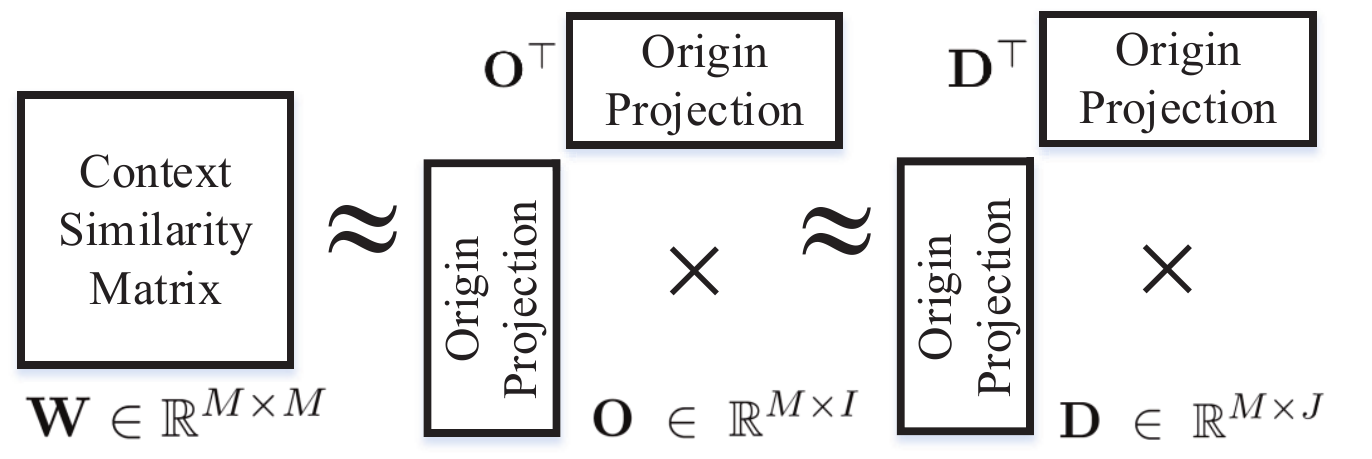}}
	\end{center}
	\vspace{-0.2cm}
	\caption{Model framework of cNTF.}
	\vspace{-0.2cm}
	\label{fig:model_framework}
\end{figure}

\subsection{Problem Definition}
\label{subsec:problem}

We here formulate the {\it urban dynamics discovery} problem as a tensor factorization problem. The model framework is given in Fig.~\ref{fig:model_framework}, where the ODT data tensor ${\tensor{R}}$, pattern tensor $ {\tensor{C}}$, and projection matrices $\mathbf{O}$, $\mathbf{D}$, and $\mathbf{T}$ have the following relationship:
\begin{equation}\label{eq:main_problem}
 {\tensor{R}} =  {\tensor{C}} \times_{o} \mathbf{O} \times_{d} \mathbf{D} \times_{t} \mathbf{T} + {\tensor{E}},
\end{equation}
where $\boldsymbol{\mathcal{E}}\in \mathbb{R}^{M \times M \times N}$ is a random error tensor, and $\times_{n}$ denotes the tensor $n$-mode product. Eq.~(\ref{eq:main_problem}) implies that the resident travel dynamics hidden inside data tensor ${\tensor{R}}$ can be well explained by the latent dynamic patterns given by pattern tensor $\tensor{C}$. The matrices $\mathbf{O}$, $\mathbf{D}$, and $\mathbf{T}$ express the projection relations between $\tensor{R}$ and $\tensor{C}$.

Note that while $\tensor{R}$ is observable from resident travels data, the pattern tensor $\tensor{C}$ as well as the projection matrices $\mathbf{O}$, $\mathbf{D}$ and $\mathbf{T}$ are unknown variables. Hence, our task is:
\begin{itemize}
	\item To infer $\tensor{C}, $$\mathbf{O}$, $\mathbf{D}$ and $\mathbf{T}$ from $\tensor{R}$;
	\item To understand urban dynamics using $\tensor{C}$, $\mathbf{O}$, $\mathbf{D}$, $\mathbf{T}$.
\end{itemize}

The urban-context similarity matrix $\mathbf{W}$ offers additional information to tensor factorization. Recall the row vector $\mathbf{o}_x$ of the projection matrix $\mathbf{O}$, which contains the membership scores of urban zone $x$ to all the OSP's. It is intuitive that similar urban zones should exhibit similar spatial patterns. Hence, we can measure the similarity of zones $x$ and $y$ by simply having $\mathbf{o}_x \mathbf{o}_y^{\top}$. Analogously, we can also measure the similarity of zones $x$ and $y$ by employing the information of DSP's in $\mathbf{D}$, {\it i.e.}, $\mathbf{d}_x \mathbf{d}_y^{\top}$. Since $\mathbf{W}$ evaluates the similarity between $x$ and $y$ as $w_{xy}$ according to the urban context, we finally have the following relationships between $\mathbf{W}$ and projection matrices $\mathbf{O}$ and $\mathbf{D}$:
\begin{equation}\label{eq:main_regularization}
%\begin{aligned}
\mathbf{W} = \mathbf{O} \mathbf{O}^{\top} + \mathbf{E}_O,~\textup{and}~~
\mathbf{W} = \mathbf{D} \mathbf{D}^{\top} + \mathbf{E}_D,
%\end{aligned}
\end{equation}
where $\mathbf{E}_O$ and $\mathbf{E}_D$ are random error matrices. Note that in Eq.~(\ref{eq:main_regularization}), $\mathbf{W}$ is an observable variable and $\mathbf{O}$ and $\mathbf{D}$ are latent ones. In other words, we can use urban context to fine-tune OSP's and DSP's in $\mathbf{O}$ and $\mathbf{D}$, respectively.

In summary, Eq.~(\ref{eq:main_problem}) and Eq.~(\ref{eq:main_regularization}) together define a context-aware Non-negative Tensor Factorization (cNTF) problem. Our task is to infer urban dynamics given cNTF.

\subsection{Extension to Long-Term Evolution}
\label{subsec:dynamic}
Long-term evolution is an important characteristic of urban dynamics, which refers to the evolution of urban spatial, temporal and spatio-temporal patterns over time. For example, temporal rhythms of resident travels in a city might change with the developments of public transport, economics, migration, {\it etc.}

We use {\it tensor sequence} to describe the evolution of urban dynamics in both data and pattern spaces. In the data space, we define $\tensor{R}|_{l=1}^{L} = \{ \tensor{R}_{1},\ldots,\tensor{R}_{L}\}$ as a data tensor sequence of length $L$, where $\tensor{R}_{l}$ is the data tensor of the $l$-th year. Suppose we factorize $\tensor{R}_{l}$ into $\mathbf{O}_l$, $\mathbf{D}_l$, $\mathbf{T}_l$ and $\tensor{C}_l$ according to Eq.~ (\ref{eq:main_problem}) and Eq.~(\ref{eq:main_regularization}), then we have the pattern tensor sequence $\tensor{C}|_{l=1}^{L} = \{ \tensor{C}_{1},\ldots,\tensor{C}_{L}\}$, and the corresponding projection matrix sequences $\mathbf{O}|_{l=1}^{L}$, $\mathbf{D}|_{l=1}^{L}$ and $\mathbf{T}|_{l=1}^{L}$, respectively.

The problem is, for any two subsequent years $l$ and $l+1$, the patterns inferred from $\tensor{R}_l$ might not be comparable to that from $\tensor{R}_{l+1}$, for they are inferred {\it separately} to optimize the objectives in Eq.~(\ref{eq:main_problem}) and Eq.~(\ref{eq:main_regularization}). Therefore, another task of this study is to infer the long-term evolution of urban dynamics given a data tensor sequence.

\section{Model}
\label{sec:model}

In this section, we reformulate the cNTF problem from a probabilistic perspective, which results in the exact objective function for urban dynamics discovery.

\subsection{Probabilistic Non-negative Tensor Factorization}

We assume the random error of observation $\tensor{E}$ follows a Gaussian distribution: $\mathcal{N}(0,\sigma_{\mathcal{R}}^2)$, then the conditional distribution over the observed entries in ${\tensor{R}}$ is defined as
\begin{equation}%\small
\begin{aligned}\label{}
&{P}( {\tensor{R}}|  {\tensor{C}}, \mathbf{O}, \mathbf{D}, \mathbf{T}, \sigma^2_{ {\mathcal{R}}}) \\
&= \prod_{x=1}^{M} \prod_{y=1}^{M} \prod_{z=1}^{N} \mathcal{N}(r_{xyz}| \tensor{C}\times_{o} \mathbf{o}_{x} \times_{d} \mathbf{d}_{y} \times_{t} \mathbf{t}_{z},\sigma^2_{ {\mathcal{R}}}).
\end{aligned}
\end{equation}

In order to obtain more evident patterns, we should introduce sparse priors to the variables in pattern space. As a result, we adopt zero-mean Laplace priors for projection matrices:
\begin{equation}%\small
\begin{aligned}
    &P(\mathbf{O}|\sigma_O) = \prod_{x=1}^M \mathcal{L}(\mathbf{o}_x|\mathbf{0}, \sigma_O \mathbf{I}_I),\\%~~~
    &P(\mathbf{D}|\sigma_D) = \prod_{y=1}^M \mathcal{L}(\mathbf{d}_y|\mathbf{0}, \sigma_D \mathbf{I}_J),\\
    &P(\mathbf{T}|\sigma_T) = \prod_{z=1}^N \mathcal{L}(\mathbf{t}_z|\mathbf{0}, \sigma_T \mathbf{I}_K),
\end{aligned}\label{}
\end{equation}
and assume zero-mean Laplace priors for the pattern tensor:
\begin{equation}%\small
P( {\tensor{C}}|\sigma_{ {\mathcal{C}}}) = \prod_{x=1}^{I} \prod_{y=1}^{J} \prod_{z=1}^{K}
  \mathcal{L}({c}_{xyz}|0, \sigma_{\mathcal{C}}).
\end{equation}
Then the posterior distribution of the pattern space variables is given by
\begin{equation}%\small
\begin{aligned}
&P({\tensor{C}}, \mathbf{O,D,T}| \tensor{R}, \sigma^2_{\mathcal{R}}, \sigma_{\mathcal{C}}, \sigma_O, \sigma_D, \sigma_T) \\
&=\frac{P(\tensor{R}| \tensor{C}, \mathbf{O}, \mathbf{D}, \mathbf{T}, \sigma^2_{\mathcal{R}}) P(\tensor{C}|\sigma_{\mathcal{C}})P(\mathbf{O}|\sigma_O) P(\mathbf{D}|\sigma_D) P(\mathbf{T}|\sigma_T) }{P(\tensor{R}|\sigma_{\mathcal{R}}^2)}, \\
\end{aligned}
\end{equation}
and the log posterior distribution is then calculated by
\begin{equation}%\small
\begin{aligned}
    \ln&~P(\tensor{C}, \mathbf{O,D,T}| \tensor{R}, \sigma^2_{\mathcal{R}}, \sigma_{\mathcal{C}}, \sigma_O, \sigma_D, \sigma_T)& \\
    \propto&-\frac{1}{2\sigma^2_{\mathcal{R}}}\sum_{xyz}(r_{xyz} - \tensor{C} \times_{o} \mathbf{o}_{x} \times_{d} \mathbf{d}_{y} \times_{t} \mathbf{t}_{z})^2 \\
    &-\frac{1}{\sigma_O}\sum_{x}\|\mathbf{o}_{x}\|_1-\frac{1}{\sigma_D}\sum_{y}\|\mathbf{d}_{y}\|_1-\frac{1}{\sigma_T}\sum_{z}\|\mathbf{t}_{z}\|_1\\
    &-\frac{1}{\sigma_{{\mathcal{C}}}}\sum_{xyz}|c_{xyz}|.
\end{aligned}\label{}
\end{equation}
Therefore, to obtain the {\em Maximum A Posteriori (MAP)} estimation of $\mathbf{O}$, $\mathbf{D}$, $\mathbf{T}$ and $\boldsymbol{\mathcal{C}}$ is equivalent to minimizing the object function
\begin{equation}\label{eq:obj_main}%\small
\begin{aligned}
    \tilde{\mathcal{J}}=&\frac{1}{2\sigma^2_{{\mathcal{R}}}}\| \tensor{R} - \tensor{C} \times_{o} \mathbf{O} \times_{d} \mathbf{D} \times_{t} \mathbf{T} \|_{F}^2 \\
    &+ \frac{1}{\sigma_O}\left\|\mathbf{O}\right\|_1+\frac{1}{\sigma_D}\|\mathbf{D}\|_1+\frac{1}{\sigma_T}\|\mathbf{T}\|_1+\frac{1}{\sigma_{\mathcal{C}}}\|\tensor{C}\|_1,
\end{aligned}
\end{equation}
where $\|.\|_{F}$ is the Frobenius-norm, $\|.\|_1$ is the L1-norm.

\begin{table}[t!]
\caption{Information of POI categories}
\vspace{-0.5cm}
\begin{center}
\begin{footnotesize}
\begin{tabular}{c|c|c|c}
\hline
\hline
\ \ ID \ \ &\ \  POI category \ \ &\ \  ID \ \  &\ \  POI category\ \  \\
\hline
\hline
1 & food \& beverage Service & 8 & education and culture\\
2 & hotel & 9 & business building\\
3 & scenic spot & 10 & residence\\
4 & finance \& insurance & 11 & living service\\
5 & corporate business & 12 & sports \& entertainments\\
6 & shopping service & 13 & medical care\\
7 & transportation facilities & 14 & government agencies \\
\hline
\hline
\end{tabular}
\end{footnotesize}
\end{center}
\vspace{-0.2cm}
\label{table:poi}
\end{table}

\subsection{Modeling Urban Contexts}

We here introduce urban contextual factors into the probabilistic non-negative tensor factorization model. We use a Beijing POI dataset, with the categories given in Table~\ref{table:poi}.

\subsubsection{Computing Urban Contextual Factors}

\begin{figure}[t]
  \centering
  \includegraphics[width=0.6\columnwidth]{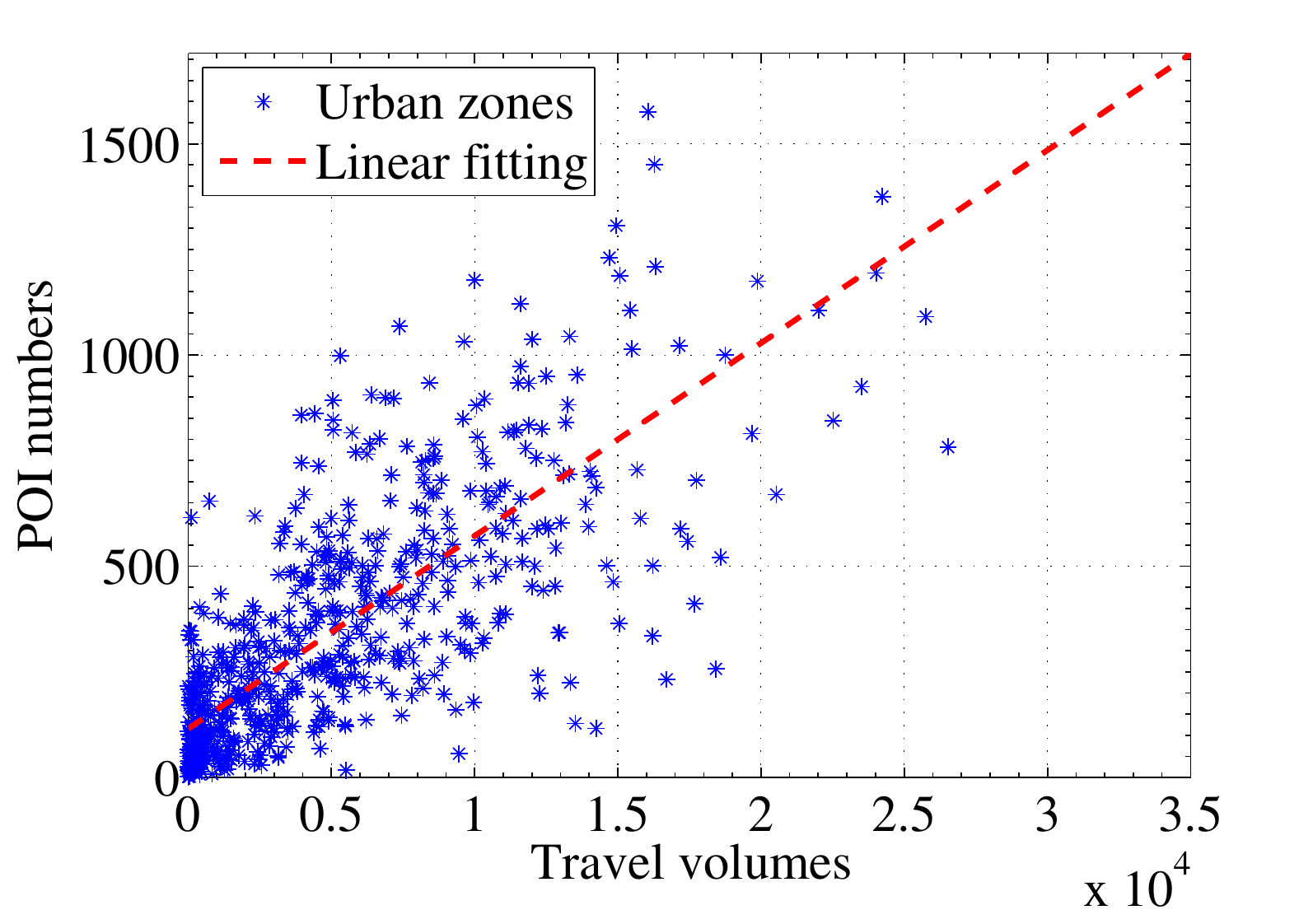}\\
  \caption{Validation of urban context correlations.}\label{fig:quantity_correlation}
\end{figure}

{\revision Fig.~\ref{fig:quantity_correlation} shows a clear positive correlation between POI quantity and the resident travel volume (including inflow and outflow) for all urban zones of Beijing. Moreover, urban zones in the same community have similar categories of POI's (see Section III of {\em Supplementary Materials}\footnote{The companion file with the supplementary materials of this paper.} for the details). Therefore, we use quantity and categories of POI's in an urban zone to describe urban contextual factors.

Suppose altogether we have $H$ POI categories, and denote $n_{ph}$ as the number of POI's in category $h$ for urban zone $p$. The fraction of the $h$-th category POI in the zone $p$ is defined as
\begin{equation}\label{}
  c_{ph} = \frac{n_{ph}}{\sum_{p=1}^P n_{ph}},
\end{equation}
The fraction of all category of POI in the zone $p$ is then defined as
\begin{equation}\label{}
  n_p=\frac{\sum_{h=1}^H n_{ph}}{\sum_{p=1}^P \sum_{h=1}^H n_{ph}},
\end{equation}
We use the vector $\mathbf{u}_p = (c_{p1}, \ldots, c_{ph}, \ldots, c_{pH} , n_p)^{\top}$ to describe the POI context of the zone $p$.}

Given the POI context vectors, the similarity of two urban zones $p$ and $q$ can be computed as
\begin{equation}\label{}
    w_{pq} = \frac{\mathbf{u}_p \cdot \mathbf{u}_q}{\|\mathbf{u}_p\| \cdot \|\mathbf{u}_q\|},
\end{equation}
which is the $(p,q)$ element of $\mathbf{W}$.

%\begin{figure}[t]\centering
%  \begin{center}\centering
%    \subfigure[Quantity Correlation]{\label{fig:quantity_correlation}\includegraphics[width=0.65\columnwidth]{fig/quantity_correlation-eps-converted-to.pdf}}\\
%    \hspace{-0.05\columnwidth}
%    \subfigure[Category Correlation]{\label{fig:category correlation} \includegraphics[width=0.65\columnwidth]{fig/category_correlation-eps-converted-to.pdf}}
%  \end{center}
%  \vspace{-0.2cm}
%  \caption{Validation of urban context correlations.}
%  \vspace{-0.2cm}
%  \label{fig:correlations}
%\end{figure}

\subsubsection{Incorporating Urban Contextual Factors}

Context-aware regularization is an effective tool to fusion contextual information into tensor and matrix factorizations~\cite{zhang2016multicalib,newyork}. We introduce urban contextual factors as context-aware regularization using a  maximum a posteriori method. Assume the elements of $\mathbf{E}_O$ and $\mathbf{E}_D$ in Eq.~\eqref{eq:main_regularization} follow zero-mean Gaussian distributions, then we have
\begin{equation}%\small
P(\mathbf{W}|\mathbf{O}, \sigma^2_{WO}) = \prod_{p=1}^{M} \prod_{q=1}^{M} \mathcal{N}(w_{pq}| \mathbf{o}_p \mathbf{o}_q^\top, \sigma^2_{WO}),
\end{equation}
and
\begin{equation}%\small
P(\mathbf{W}|\mathbf{D}, \sigma^2_{WD}) = \prod_{p=1}^{M} \prod_{q=1}^{M} \mathcal{N}(w_{pq}| \mathbf{d}_p \mathbf{d}_q^\top, \sigma^2_{WD}).
\end{equation}
Let $\Omega = \{\sigma^2_{\mathcal{R}},  \sigma^2_{WO},  \sigma^2_{WD}, \sigma_{O},  \sigma_{D}, \sigma_{T}, \sigma_{\mathcal{C}}\}$. Given the data tensor ${\tensor{R}}$ and urban context matrix $\mathbf{W}$, the posterior distribution of $\mathbf{O}$, $\mathbf{D}$, $\mathbf{T}$ and $\tensor{C}$ is given by
\begin{equation}%\small
\begin{aligned}
    &P(\mathbf{O}, \mathbf{D}, \mathbf{T}, \tensor{C} | \tensor{R}, \mathbf{W}, \Omega) \\
    &\propto P(\tensor{R} | \mathbf{O}, \mathbf{D}, \mathbf{T}, \tensor{C}, \Omega) P(\mathbf{W} | \mathbf{O}, \Omega) P(\mathbf{W} | \mathbf{D}, \Omega) \\
    &~~~~P(\mathbf{O}|0,\Omega)  P(\mathbf{D}|0,\Omega) P(\mathbf{T}|0,\Omega) P(\tensor{C}|0,\Omega),
\end{aligned}
\end{equation}
and the log posterior distribution is
\begin{equation}%\small
\begin{aligned}
    &\ln P(\mathbf{O}, \mathbf{D}, \mathbf{T}, \tensor{C} | \tensor{R}, \mathbf{W}, \Omega) \\
    &\propto-\frac{1}{2\sigma^2_{\mathcal{R}}}\sum_{xyz}(r_{xyz} - \tensor{C} \times_{o} \mathbf{o}_{x} \times_{d} \mathbf{d}_{y} \times_{t} \mathbf{t}_{z})^2 \\
    &-\frac{1}{2\sigma^2_{WO}}\sum_{pq}(w_{pq} - \mathbf{o}_{p} \mathbf{o}^{\top}_{q})^2  -\frac{1}{2\sigma^2_{WD}}\sum_{pq}(w_{pq} - \mathbf{d}_{p} \mathbf{d}^{\top}_{q})^2 \\
    &-\frac{1}{\sigma_O}\sum_{x}\|\mathbf{o}_{x}\|_1-\frac{1}{\sigma_D}\sum_{y}\|\mathbf{d}_{y}\|_1
    -\frac{1}{\sigma_T}\sum_{z}\|\mathbf{t}_{z}\|_1\\
    &-\frac{1}{\sigma_{\mathcal{C}}}\sum_{ijk}|c_{ijk}|.
\end{aligned}
\end{equation}

To maximize the posterior distribution is equivalent to minimizing the sum-of-squared errors function with hybrid quadratic regularization terms, {\it i.e.},
\begin{equation}%\small
\label{eq:main_obj}
\begin{aligned}
  \min_{\mathbf{O},\mathbf{D},\mathbf{T},\tensor{C}}~&  {\mathcal{J}}=\| \tensor{R} - \tensor{C} \times_{o} \mathbf{O} \times_{d} \mathbf{D} \times_{t} \mathbf{T} \|_{F}^2 \\
    &+ \alpha \| \mathbf{W} - \mathbf{OO}^\top \|_{F}^2 + \beta \| \mathbf{W} - \mathbf{DD}^\top \|_{F}^2 \\
    &+\gamma \left\|\mathbf{O}\right\|_1+\delta \|\mathbf{D}\|_1+\epsilon \|\mathbf{T}\|_1+\varepsilon\|\tensor{C}\|_1\\
   \textbf{\emph{s.t.}} &~~\mathbf{O} \geq 0, \mathbf{D}\geq 0, \mathbf{T}\geq 0, \tensor{C}\geq 0,
\end{aligned}
\end{equation}
where $\alpha = \frac{\sigma^2_{\boldsymbol{\mathcal{R}}}}{\sigma^2_{WO}}$, $\beta = \frac{\sigma^2_{\boldsymbol{\mathcal{R}}}}{\sigma^2_{WD}}$, $\gamma = \frac{2\sigma^2_{\boldsymbol{\mathcal{R}}}}{\sigma_{O}}$, $\delta = \frac{2\sigma^2_{\boldsymbol{\mathcal{R}}}}{\sigma_{D}}$, $\epsilon = \frac{2\sigma^2_{\boldsymbol{\mathcal{R}}}}{\sigma_{T}}$, $\varepsilon = \frac{2\sigma^2_{\boldsymbol{\mathcal{R}}}}{\sigma_{\boldsymbol{\mathcal{C}}}}$.
Note that we introduce non-negativity constraints on the variables so as to avoid perplexing negative travel volumes. Eq.~\eqref{eq:main_obj} indeed formulates the cNTF problem defined in Sect.~\ref{subsec:problem}.

\subsection{Neighboring Regularization}
\label{subsec:neighbor}
Let $\mathcal{SP}_i = \{x: v_{xi} = \max_{1\leq j\leq I} v_{xj}\}$ denote the $i$th {\it urban community} corresponding to the spatial pattern $\mathbf{v}_{:i}$ in the spatial projection matrix $\mathbf{V}$.  For the urban zones in $\mathcal{SP}_i$, it is natural to expect that: $i$) they are geographically neighboring to each other, and $ii$) their resident mobility behaviors are similar to one another and different from that in other communities. These, however, have not been considered in the above-mentioned cNTF model.

To address these, we here introduce the so-called Neighboring Regularization (NR), which is inspired by the conditional random field based image segmentation method in~\cite{Kr2012Efficient}. Specifically, we model urban community discovery as an image segmentation problem; that is, the community labels of urban zones are modeled as a Markov random field $G(\mathbb{V}, \mathbb{E})$, where $\nu_x \in \mathbb{V}$ is the community label of urban zone $x$, and $e_{xy} \in \mathbb{E}$ is an undirectional dependency between urban zone $x$ and $y$. For the latent $\nu_x$, we have an observable matrix $\mathbf{R}_{x::}$ for the origin order of $\tensor{R}$, or $\mathbf{R}_{:y:}$ for the destination order.

Without loss of generality, in what follows, we use the origin order as an example to introduce the neighboring regularization. Suppose $G(\mathbb{V}, \mathbb{E})$ and $\mathbf{R}_{x::}$, $x\in \{1 \dots M\}$, satisfy the conditional random field hypothesis. Similar to the classical image segmentation task in~\cite{Kr2012Efficient}, the optimization objective for community discovery is to maximize a potential function as
\begin{equation}\label{equ:potential}
\zeta = \sum_{x=1}^{M}\psi^{u}_x(\nu_x) +  \sum_{x=1}^{M} \sum_{y \in M_x} \psi^{p}_{xy}(\nu_x, \nu_y),
\end{equation}
where $M_x$ is the set of neighbor zones of zone $x$. $\psi^{u}_x(\nu_x)$ is the unary potential of the CRF in zone $x$ when the community label of $x$ is set to $\nu_x$, which is defined as
\begin{equation}\label{}
\psi^{u}_x(\nu_x) = -\log \frac{o_{x\nu_x}}{\sum_{i=1}^{I} o_{xi}}.
\end{equation}
$\psi^{p}_{xy}(\nu_x, \nu_y)$ is the pairwise potential between zones $x$ and $y$ when the community labels of $x$ and $y$ are set to $\nu_x$ and $\nu_y$, respectively; that is,
\begin{equation}
\begin{aligned}
\psi^{p}_{xy}(\nu_x, \nu_y) =
\begin{cases}
0, &\mathrm{if}~\nu_x = \nu_y,\\
g(x,y), &\mathrm{otherwise}.
\end{cases}
\end{aligned}
\end{equation}
Note that $g(x,y)$ is a function of the difference between $\mathbf{R}_{x::}$ and $\mathbf{R}_{y::}$, which is defined as a Gaussian kernel as follows:
\begin{equation}
g(x, y) =  \exp\left(-\frac{\left \| \mathbf{R}_{x::} - \mathbf{R}_{y::} \right \|_F^{2}}{2 \sigma_{\mathrm{NR}}^{2}}\right),
\end{equation}
where $\sigma_{\mathrm{NR}}$ is a parameter suggested in~\cite{Kr2012Efficient}. This actually introduces a penalty for the zones that are adjacent and have similar resident mobility behaviors but are assigned to different communities.

In a nutshell, Eq.~\eqref{equ:potential} introduces the spatial community discovery problem, which could be regarded as a neighboring regularization to cNTF, and thus form the so-called {\bf NR-cNTF} model.

%Considering both the tensor factorization task and the community segmentation task, the optimization objective of our model becomes to minimize the objective function~\eqref{eq:main_obj} for cNTF and the potential function ~\eqref{equ:potential} for community segmentation at the same time. Therefore, the community segmentation task could be considered as a neighboring regularization of the cNTF model, which is named as NR-cNTF.

%Because the CRF regularization can not be written as an explicit form with the objective function of cNTF, we call it as an implicit regularization. The cNTF model with the implicit CRF regularization is called as IR-cNTF in this study.

\begin{figure}[t]\centering
	\begin{tikzpicture}[->, auto, node distance=0.7in, thick]
	\node (G0) {$\mathcal{G}_0$};
	\node (G1) [right of=G0] {$\mathcal{G}_1$};
	\node (G2) [right of=G1] {$\mathcal{G}_2$};
	\node (Gdot) [right of=G2] {$\dots \mathcal{G}_l \dots$};
	\node (GL) [right of=Gdot] {$\mathcal{G}_L$};
	\node (R1) [below of=G1] {$\tensor{R}_1\mathbf{W}_1$};
	\node (R2) [below of=G2] {$\tensor{R}_2\mathbf{W}_2$};
	\node (Rdot) [below of=Gdot] {$\dots \tensor{R}_l\mathbf{W}_l \dots$};
	\node (RL) [below of=GL] {$\tensor{R}_L\mathbf{W}_L$};
	\draw (G0) -- (G1) node[midway, above] {init};
	\draw (G1) -- (G2) node[midway, above] {init};
	\draw (G2) -- (Gdot) node[midway, above] {init};
	\draw (Gdot) -- (GL) node[midway, above] {init};
	\draw (R1) -- (G1) node[midway, right] {};
	\draw (R2) -- (G2) node[midway, right] {};
	\draw (Rdot) -- (Gdot) node[midway, right] {};
	\draw (RL) -- (GL) node[midway, right] {};
	\end{tikzpicture}
	\vspace{-0.2cm}
	\caption{Pipeline initialization for tensor sequence analysis.}\label{fig:pipe_init}
	\vspace{-0.2cm}
\end{figure}
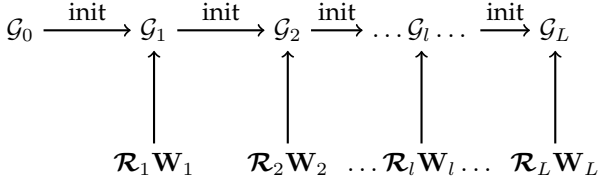

\subsection{Modeling Long-Term Evolution}
We here introduce a simple yet effective way to model the long-term evolution of spatio-temporal patterns. Let $\tensor{R}_l$ and $\mathbf{W}_l$ denote the data tensor and POI similarity matrix in the $l$-th year, and $\mathcal{G}_l = \left\{ \tensor{C}_l, \mathbf{O}_l, \mathbf{D}_l , \mathbf{T}_l\right\}$ denote the set of latent patterns learnt from the $l$-th year's data, $l=1,2,\cdots,L$.

As described in Sect.~\ref{subsec:dynamic}, to factorize every $\tensor{R}_{l}$ independently for $\mathcal{G}|_{l=1}^{L}$ is often inappropriate for generating incomparable patterns in successive years. The Dynamic Tensor Analysis (DTA) scheme suggested in~\cite{DTA,fema}  cannot fulfill our task either for using $\tensor{R}_l$ as well as historical data tensors to obtain a ``hybrid'' $\mathcal{G}_l$, which is not the genuine $\mathcal{G}_l$ we aim to analyze in practice.

We here propose a simple Pipeline Initialization based Tensor Sequence Analysis (PI-TSA) method. In PI-TSA, the factorization results in $\mathcal{G}_{l}$ are expressed as
\begin{equation}\label{eq:pipe_init}
\mathcal{G}_{l} = {f}_{\textrm{NR-cNTF}}\left(\tensor{R}_l, \mathbf{W}_l, \mathcal{G}_{l-1}\right),
\end{equation}
where $f_{\textrm{NR-cNTF}}$ denotes the optimization algorithm for NR-cNTF. %, $\tensor{R}_l$ and $\mathbf{W}_l$ are the data space input variables of NR-cNTF, and $\mathcal{G}_{l-1}$ is the factorization result of the step $l-1$, which is used to set model initial values of the $l$-th step.
Fig.~\ref{fig:pipe_init} further illustrates PI-TSA via a flow chart. As can be seen, the key of PI-TSA is to set the initial values of the $l$-th year's optimization as the outputs in the ($l$-1)-th step ({\it i.e.}, $\mathcal{G}_{l-1}$). In this way, the patterns in the ($l$-1)-th year can be ``inherited'' by the patterns in the $l$-th year, and only the information of $\tensor{R}_l$ and $\mathbf{W}_l$ is used for pattern discovery in the $l$-th year.

\section{Inference}
\label{sec:inference}

\subsection{Basic Optimization}

\begin{algorithm}[t]%\small
	\caption{Block Coordinate Descent Procedure}
	\begin{algorithmic}\label{alg:BCD}
		\REQUIRE Data sets $\left\{\tensor{R}, \mathbf{W}\right\}$, parameters $\left\{ \gamma, \delta, \epsilon, \varepsilon \right\}$
		\STATE \textbf{Initialization}: $\left(\tensor{C}^{(0)}, \mathbf{O}^{(0)}, \mathbf{D}^{(0)}, \mathbf{T}^{(0)}\right)$
		\FOR {$s=1, 2, \ldots$}
		\STATE Update $\tensor{C}^{(s)}$ by solving the problem (\ref{eq:bcd_c}).
		\STATE Update $\mathbf{O}^{(s)}$ by solving the problem (\ref{eq:bcd_o}).
		\STATE Update $\mathbf{D}^{(s)}$ by solving the problem (\ref{eq:bcd_d}).
		\STATE Update $\mathbf{T}^{(s)}$ by solving the problem (\ref{eq:bcd_t}).
		\STATE Apply Algorithm~\ref{alg:crf} to $\mathbf{O}^{(s)}$.
		\STATE Apply Algorithm~\ref{alg:crf} to $\mathbf{D}^{(s)}$.
		\IF {convergence}
		\STATE return $\left(\tensor{C}^{(s)}, \mathbf{O}^{(s)}, \mathbf{D}^{(s)}, \mathbf{T}^{(s)}\right)$.
		\ENDIF
		\ENDFOR
	\end{algorithmic}
\end{algorithm}

We adopt the Block Coordinate Descent-Proximal Gradient (BCD-PG) algorithm~\cite{xu2015alternating,xu2013block} to solve the cNTF problem in Eq.~(\ref{eq:main_obj}). While this function is not jointly convex with respect to $\tensor{C}$, $\mathbf{O}$, $\mathbf{D}$, and $\mathbf{T}$, it is {\em block multiconvex} with each one when the other three are fixed. Therefore, as shown in Algorithm~\ref{alg:BCD}, we adopt a Block Coordinate Descent (BCD) procedure, which starts from an initialization on $\mathcal{G}^{(0)}$, and then iteratively updates $\mathcal{G}^{(s)}$, $s=1,2,\cdots$, by
\begin{subequations}\small
\begin{align}
&\tensor{C}^{(s)} = \argmin_{\tensor{C}}{\mathcal{J}\left(\tensor{C}, \mathbf{O}^{(s-1)}, \mathbf{D}^{(s-1)}, \mathbf{T}^{(s-1)}\right)} + \gamma\|\tensor{C}\|_1,  \label{eq:bcd_c} \\
&\mathbf{O}^{(s)} = \argmin_{\mathbf{O}}{\mathcal{J}\left(\tensor{C}^{(s)}, \mathbf{O}, \mathbf{D}^{(s-1)}, \mathbf{T}^{(s-1)}\right)} + \delta\|\mathbf{O}\|_1, \label{eq:bcd_o}\\
&\mathbf{D}^{(s)} = \argmin_{\mathbf{D}}{\mathcal{J}\left(\tensor{C}^{(s)}, \mathbf{O}^{(s)}, \mathbf{D}, \mathbf{T}^{(s-1)}\right)} + \epsilon\|\mathbf{D}\|_1, \label{eq:bcd_d}\\
&\mathbf{T}^{(s)} = \argmin_{\mathbf{T}}{\mathcal{J}\left(\tensor{C}^{(s)}, \mathbf{O}^{(s)}, \mathbf{D}^{(s)}, \mathbf{T}\right)} + \varepsilon\|\mathbf{T}\|_1. \label{eq:bcd_t}
\end{align}
\end{subequations}

Let $\left( \mathbf{g}_1, \mathbf{g}_{2}, \mathbf{g}_{3}, \mathbf{g}_{4}\right)$ denote $\left(\tensor{C}, \mathbf{O}, \mathbf{D}, \mathbf{T}\right)$ for concision. Using a Proximal Gradient (PG) method, the algorithm updates the $i$-th variable of $\mathcal{G}$ in the $s$-th round as
\begin{equation}\small
\label{eq:apg}
\begin{split}
\mathbf{g}_{i}^{(s)} =& \argmin_{\mathbf{g}_i \geq 0 } \left\langle \frac{\partial\mathcal{J}\left(\mathbf{g}_{<i}^{(s)}, \tilde{\mathbf{g}}_{i}^{(s)},\mathbf{g}_{>i}^{(s-1)}\right)}{\partial \mathbf{g}_{i}}, \mathbf{g}_{i} - \tilde{\mathbf{g}}_{i}^{(s)} \right\rangle \\
+& \frac{\tau_{i}}{2} \left \| \mathbf{g}_{i} - \tilde{\mathbf{g}}_{i}^{(s)} \right \|^2_F+ \lambda_{i}\|\mathbf{g}_i\|_1\\
=& \max \left\{0,  \tilde{\mathbf{g}}_{i}^{(s)}- \frac{1}{\tau_i} \frac{\partial\mathcal{J}\left(\mathbf{g}_{<i}^{(s)}, \tilde{\mathbf{g}}_{i}^{(s)},\mathbf{g}_{>i}^{(s-1)}\right)}{\partial \mathbf{g}_{i}} - \frac{\lambda_i}{\tau_i}\right\},
\end{split}
\end{equation}
where $\langle\cdot\rangle$ denotes the inner product, $\mathbf{g}_{<i}^{(s)}$ denotes $\{\mathbf{g}_{1}^{(s)} \ldots \mathbf{g}_{i-1}^{(s)}\}$, and $\mathbf{g}_{>i}^{(s-1)}$ denotes $\{\mathbf{g}_{i+1}^{(s-1)} \ldots \mathbf{g}_{4}^{(s-1)}\}$. The variable $\tilde{\mathbf{g}}_{i}^{(s)}$ is a linear extrapolated point as follows:
\begin{equation}\small
\tilde{\mathbf{g}}_{i}^{(s)} = \mathbf{g}_{i}^{(s-1)} + \omega_{i}^{(s)}\left(\mathbf{g}_{i}^{(s-1)} - \mathbf{g}_{i}^{(s-2)} \right),
\end{equation}
where $\omega_{i}^{(s)}$ is an extrapolation weight set according to~\cite{xu2013block}. The parameter $\tau_i$ in \eqref{eq:apg} is a Lipschitz constant of $\frac{\partial\mathcal{J}\left(\mathbf{g}_{i}\right)}{\partial \mathbf{g}_{i}}$ with respect to $\mathbf{g}_{i}$, namely,
\begin{equation}\small
\label{eq:Lipschitz}
\left \| \frac{\partial\mathcal{J}\left(\mathbf{g}_{i_1}\right)}{\partial \mathbf{g}_{i_1}} - \frac{\partial\mathcal{J}\left(\mathbf{g}_{i_2}\right)}{\partial \mathbf{g}_{i_2}}\right\|_F \le \tau_i\|\mathbf{g}_{i_1} - \mathbf{g}_{i_2} \|_F, \forall~{\mathbf{g}_{i_1},\mathbf{g}_{i_2}},
\end{equation}
and $\lambda_i$ is the regularization parameter of $\mathbf{g}_i$. Specifically, the gradients of $\mathcal{J}$ with respect to each component are calculated as
\begin{equation}\small
\label{equ:gradient_C}
\begin{split}
\frac{\partial{\mathcal{J}}}{\partial{\tensor{C}}}  &=  2\ \Big( {\tensor{C}} \times_o \left(\mathbf{O}^{\top}\mathbf{O}\right) \times_d \left(\mathbf{D}^{\top}\mathbf{D}\right) \times_t \left(\mathbf{T}^{\top}\mathbf{T}\right) \\
& - {\tensor{R}} \times_o \mathbf{O}^{\top} \times_d \mathbf{D}^{\top} \times_t \mathbf{T}^{\top} \Big),\\
%\end{split}
%\end{equation}
%\begin{equation}%\small
%\label{equ:gradient_O}
%\begin{split}
\frac{\partial{\mathcal{J}}}{\partial{\mathbf{O}}} &=  2\ \Big( \mathbf{O}\left({\tensor{C}} \times_d \left(\mathbf{D}^{\top}\mathbf{D}\right) \times_t \left(\mathbf{T}^{\top}\mathbf{T}\right)\right)_{\left(o\right)}\tensor{C}_{\left(o\right)}^{\top} \\
& - \left({\tensor{R}} \times_d \mathbf{D}^{\top} \times_t \mathbf{T}^{\top}\right)_{\left(o\right)}\tensor{C}_{\left(o\right)}^{\top} -  \alpha \left( \mathbf{W} - \mathbf{O}\mathbf{O}^{\top}\right)\mathbf{O} \Big),\\
%\end{split}
%\end{equation}
%\begin{equation}%\small
%\label{equ:gradient_D}
%\begin{split}
\frac{\partial{\mathcal{J}}}{\partial{\mathbf{D}}} &= 2\ \Big( \mathbf{D}\left({\tensor{C}} \times_o \left(\mathbf{O}^{\top}\mathbf{O}\right) \times_t \left(\mathbf{T}^{\top}\mathbf{T}\right)\right)_{\left(d\right)}\tensor{C}_{\left(d\right)}^{\top} \\
& - \left({\tensor{R}} \times_o \mathbf{O}^{\top} \times_t \mathbf{T}^{\top}\right)_{\left(d\right)}\tensor{C}_{\left(d\right)}^{\top} - \beta \left( \mathbf{W} - \mathbf{D}\mathbf{D}^{\top}\right)\mathbf{D}\Big),\\
%\end{split}
%\end{equation}
%and
%\begin{equation}%\small
%\label{equ:gradient_T}
%\begin{split}
\frac{\partial{\mathcal{J}}}{\partial{\mathbf{T}}} &= 2\ \Big(\mathbf{T}\left({\tensor{C}} \times_o \left(\mathbf{O}^{\top}\mathbf{O}\right) \times_d \left(\mathbf{D}^{\top}\mathbf{D}\right)\right)_{\left(t\right)}\tensor{C}_{\left(t\right)}^{\top} \\
& - \left({\tensor{R}} \times_o \mathbf{O}^{\top} \times_d \mathbf{D}^{\top}\right)_{\left(t\right)}\tensor{C}_{\left(t\right)}^{\top} \Big),
\end{split}
\end{equation}
where $\tensor{X}_{\left(x\right)}$ denotes the mode-$x$ matricization of tensor $\tensor{X}$.

\subsection{Neighboring Regularization Optimization}

Algorithm~\ref{alg:crf} shows the optimization process of neighboring regularization. Without loss of generality, we still take the origin order for illustration. In each cNTF optimization iteration, Algorithm~\ref{alg:crf} regularizes the projection matrix $\mathbf{O}$ through the following steps:

1) {\bf Calculate Unary Potentials}: We first normalize $\mathbf{O}$ as
\begin{equation}\label{}
o'_{xi} = \frac{o_{xi}}{\sum_{j=1}^{I} o_{xj}}.
\end{equation}
Then the unary potential of $o_{xi}$ is $\psi^{u}_x(i) = -\log o'_{xi}$.

2) {\bf Calculate Pairwise Potentials}: We then calculate the average pairwise potential of $\nu_x = i$ to $\nu_y \in \{j|j\neq i\}$ as
\begin{equation}\label{eq:pairwise}
{Q}_{xi} =  \sum_{j \neq i} \sum_{y\in M_x} P_{yj} \cdot \psi_{xy}^p(i,j),
\end{equation}
where $M_x$ is the set of neighbor zones for zone $x$. $P_{yj}$ in~Eq.~\eqref{eq:pairwise} is a probability of
$v_y = j$, which is defined as
\begin{equation}\label{}
P_{yj} = \frac{\exp(-\psi^{u}_y(j))}{Z_y} = o'_{yj},
\end{equation}
where $1/Z_{x}$ denotes the partition function.

3) {\bf Update the Projection Matrix}: Finally, we calculate the total potential of $o_{xi}$ as
\begin{equation}\label{}
\zeta_{xi} = \psi^{u}_x(i) + {Q}_{xi}.
\end{equation}
The regularized element is then defined as
\begin{equation}\label{}
\tilde{o}_{xi} = \exp(-\zeta_{xi})\cdot\sum_{j=1}^{I} o_{xj}.
\end{equation}
For the $s$-th round of iteration in Algorithm~\ref{alg:BCD}, we define $\Delta_{NR} = \tilde{o}_{xi}^{(s)} - o_{xi}^{(s)}$, and $\Delta_{cNTF} = o_{xi}^{(s)} - o_{xi}^{(s-1)}$. Algorithm~\ref{alg:crf} then updates $o_{xi}^{(s)}$ as
\begin{equation}\label{eq:deta}%\small
o_{xi}^{(s)} =
\begin{cases}
\max\{0, o_{xi}^{(s-1)} + \Delta^{cNTF} +¡¡\Delta^{NR}\}, & \mathrm{if}~\Delta^{cNTF} \leq 0, \\
o_{xi}^{(s-1)} + \max\{0, \Delta^{cNTF} +\Delta^{NR}\}, & \mathrm{otherwise}.
\end{cases}
\end{equation}

Note that $\tilde{o}_{xi}^{(s)} \leq o_{xi}^{(s)} \Rightarrow \Delta^{NR} \leq 0$, so the update of $o_{xi}$ in Eq.~\eqref{eq:deta} is in the same direction with the gradient of $o_{xi}^{(s-1)}$. Algorithm~\ref{alg:crf} therefore ensures that the reconstruction error in each iteration is always the same or lower than that in the previous iteration.

\begin{algorithm}[t]
	\caption{Neighboring Regularization Optimization}
	\begin{algorithmic}\label{alg:crf}
		\STATE {\bf Unary Potentials}: $o'_{xi} \leftarrow \frac{o_{xi}}{\sum_{j=1}^{I} o_{xj}},~~\psi^{u}_x(i) \leftarrow -\log o'_{xi}.$
		\STATE {\bf Pairwise Potentials}: $\tilde{Q}_{xi} \leftarrow  \sum_{j \neq i} \sum_{y\in M_x} \psi_{xy}^p(i,j) o'_{yj}.$
		\STATE {\bf Update the Projection Matrix.}
	\end{algorithmic}
\end{algorithm}

\section{Experimental Results}
\label{sec:experiments}

In this section, we conduct extensive experiments to evaluate the effectiveness of our methods in learning urban dynamics and gaining managerial insights for urban planning. We also compare our methods with some baselines on traffic prediction, which justifies the modeling of urban contexts and neighboring regulation in NR-cNTF.

\subsection{Experimental Setup}

%\subsection{Parameter setting}
%
%In this section, we report the sensitivity of parameters our method involves, the dimensionality of pattern space $M$, $N$, and the tradeoff parameter for urban contextual regularization $\alpha$, $\beta$, and L1 regularization $\gamma$, $\delta$ and $\epsilon$.

\subsubsection{Data Sets}

Three types of data sets were used in our experiments including taxi trajectory data, POI data, and Traffic Analysis Zone data. The taxi trajectory data set contains the GPS trajectories of 20,000 Beijing taxis collected in November 2008 and November 2015, from which we extracted more than 6 million trips of taxi passengers to present the daily mobility behaviors of residents in Beijing. The POI data set contains more than 400 thousands POI records of Beijing in the years of 2008 and 2015. The Traffic Analysis Zone (TAZ) data set, offered by Beijing Municipal Commission of Transportation, divides the Beijing area within the 5-$th$ Ring Road into 651 zones. %, as shown in Fig.~\ref{fig:dataset_beijingmap}.
Using the three data sets, we built two data tensors $(651 \times 651 \times 24)$ and two POI context matrices $(651 \times 651)$ for the years of 2008 and 2015, respectively. {\revision In the experiments, we only use data of workdays to construct the data tensor $\tensor{R}$, so the discovered patterns reflect resident mobility in workdays. People¡¯s leisure patterns in holiday could be very different from their workday patterns. We have conducted extra experiments on holiday data, and included the results to {\em Supplementary Materials} for readers with interests.}

\begin{comment}
\begin{figure}[t]\centering
	\includegraphics[width=0.6\columnwidth]{fig/dataset_beijingmap-eps-converted-to.pdf}\\
	\caption{Areas in 5-th Ring Road of Beijing split by Traffic Analysis Zones}\label{fig:dataset_beijingmap}
\end{figure}
\end{comment}

\subsubsection{Setting of Dimensionality of Pattern Space}

The goal of the NR-cNTF model is to find an $I \times J \times K$-dimensional pattern space. How to set $I, J, K$ appropriately, however, is a ``tricky'' issue. If the dimensionality is too small, we might omit some urban dynamics; if too large, we might obtain many trivial patterns (for the extreme case, if the dimensionality of the pattern space is the same as the data space, the patterns will be meaningless).

\begin{figure}[t]
	\centering
	\subfigure[Setting of $I,J$]{\includegraphics[width=0.4\columnwidth]{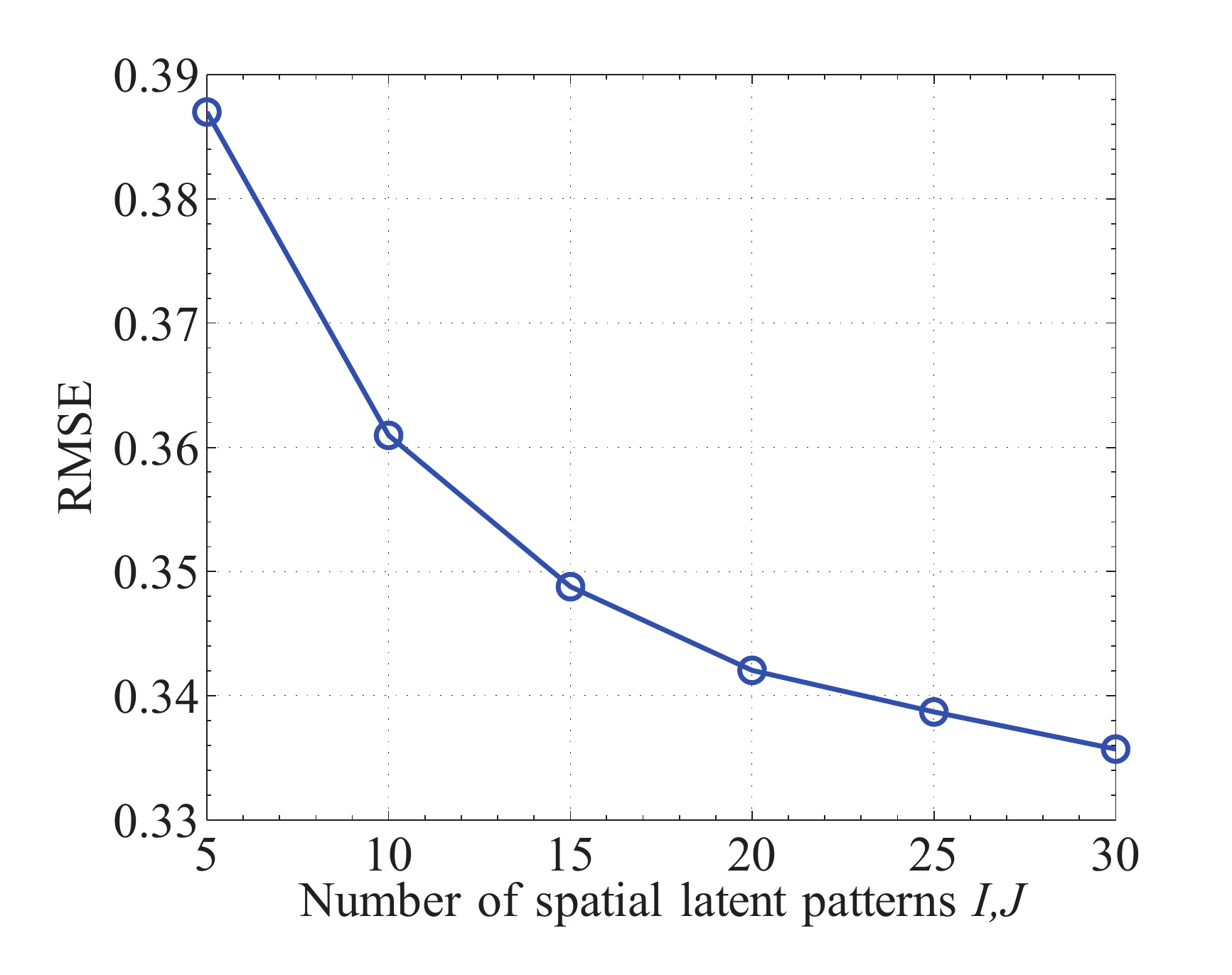} }~~~~~~
	\subfigure[Setting of $K$]{\includegraphics[width=0.4\columnwidth]{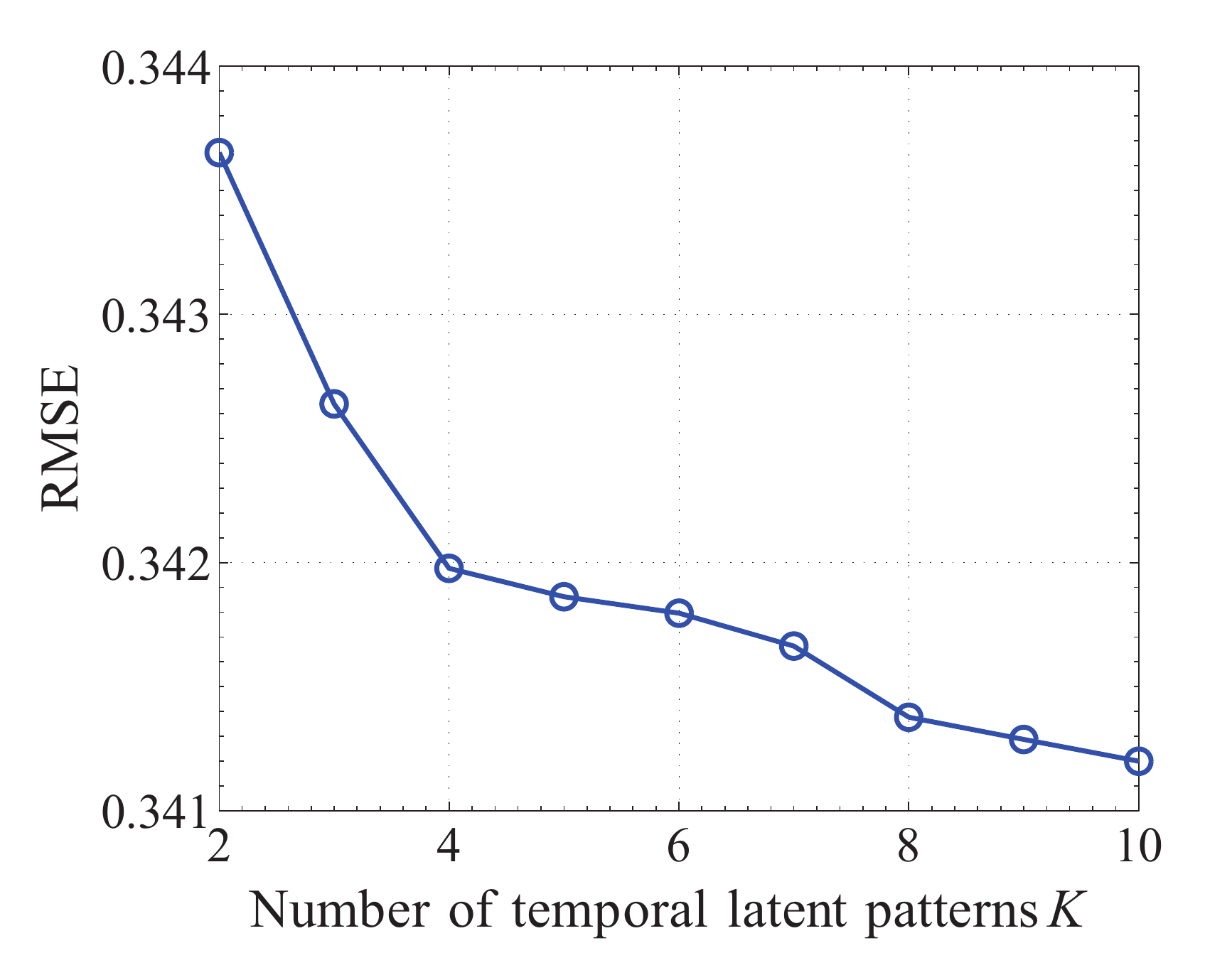} }
%	\vspace{-0.2cm}
	\caption{Performance with varying dimensionality of pattern space.}
%	\vspace{-0.2cm}
	\label{fig:dimension}
\end{figure}

In our experiments, we set the parameters carefully so as to make a tradeoff between the reconstruction error and the dimension reduction. The reconstruction error is evaluated by {\em Root Mean Square Error} (RMSE) defined as follows:
\begin{equation}\label{eq:rmse}%\small
\mathrm{RMSE} = \sqrt{ \frac{\sum_{x=1}^{M}\sum_{y=1}^{M}\sum_{z=1}^{N} \left(r_{xyz} - \hat{r}_{xyz}\right)^2} {M \times M \times N} },
\end{equation}
where $\hat{r}_{xyz}$ is the ($x,y,z$) element of the reconstructed data tensor. We repeated experiments 10 times with $I=J$ ranging from 5 to 30 and $K$ ranging from 2 to 10. Fig.~\ref{fig:dimension} gives the resultant average reconstruction errors with different parameters, where RMSE reduces sharply at the very beginning but slows down when $I,J \ge 20$ and $K \ge 4$. We therefore set $I = J = 20$ and $K = 4$ as defaults.

\begin{figure}[t]
	\centering
	\subfigure[POI Regularization]{\label{fig:lambda_vary_a}\includegraphics[width=0.4\columnwidth]{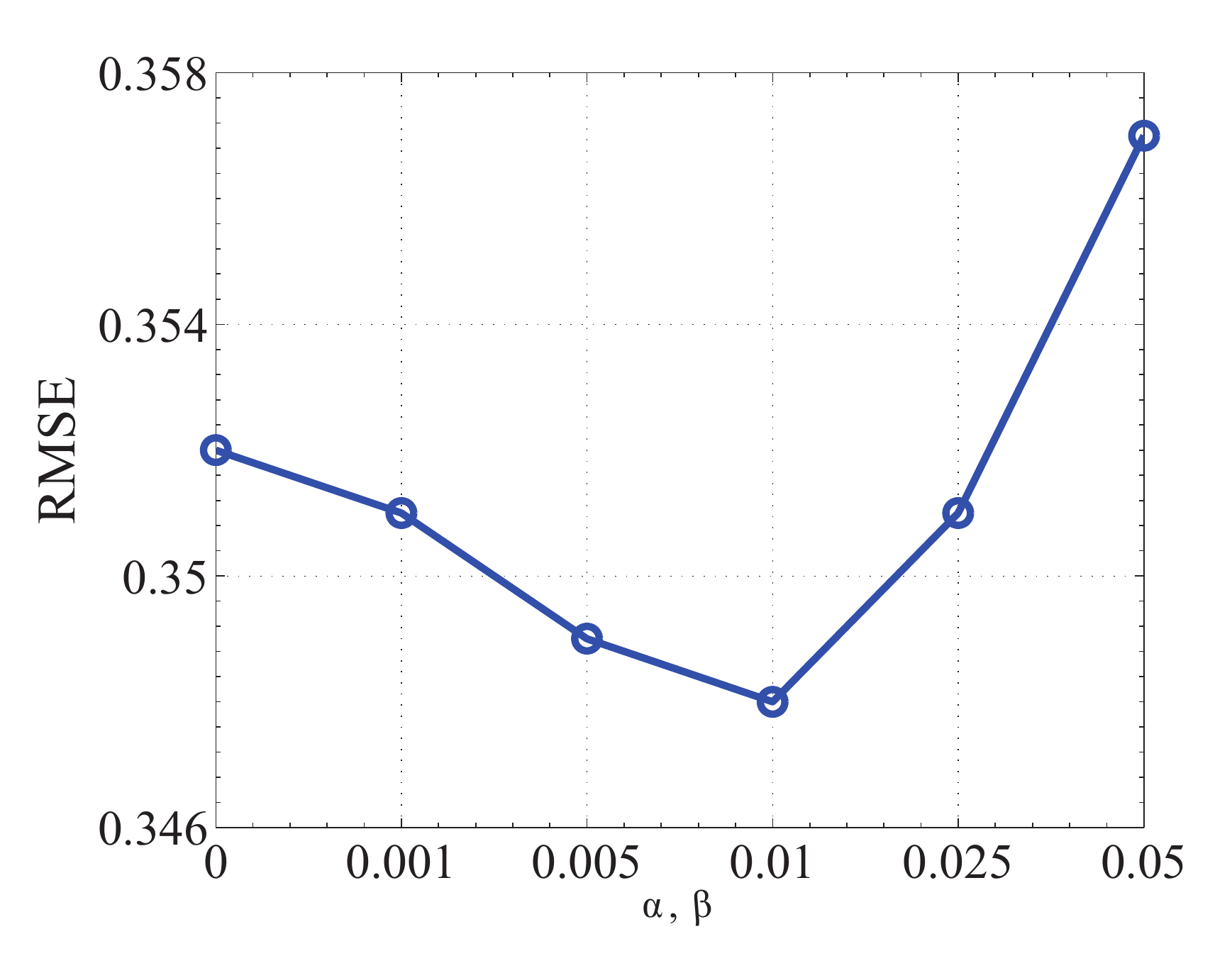}}~~~~~~
	\subfigure[L1 Regularization]{\label{fig:lambda_vary_b}\includegraphics[width=0.4\columnwidth]{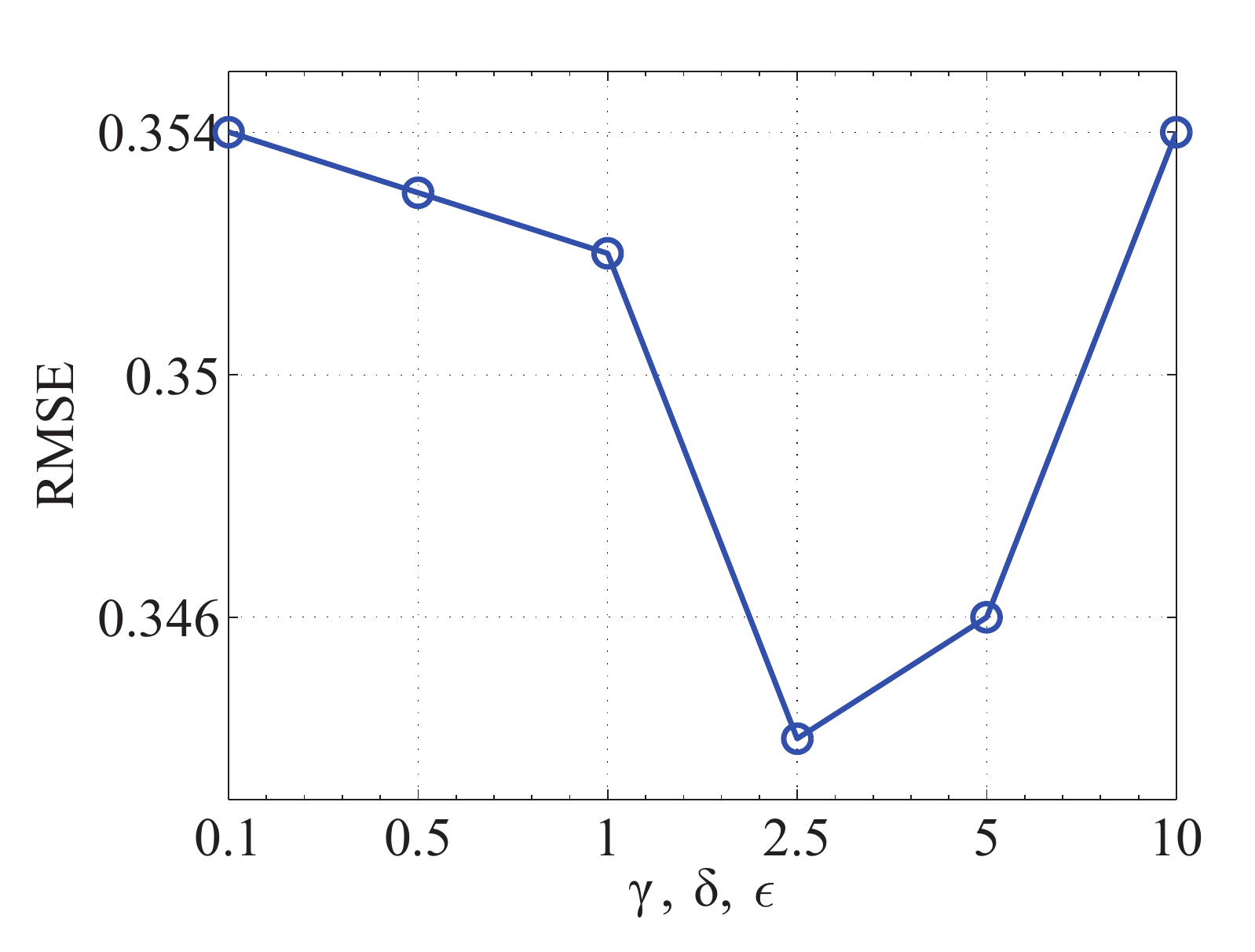} }
	\vspace{-0.2cm}
	\caption{Performance with varying POI and L1 regularization coefficients.}
	\vspace{-0.2cm}
	\label{fig:lambda_vary}
\end{figure}

\begin{figure}[t]\centering
	\begin{center}\centering
		\subfigure[2008]{\label{fig:T2008} \includegraphics[width=0.47\columnwidth]{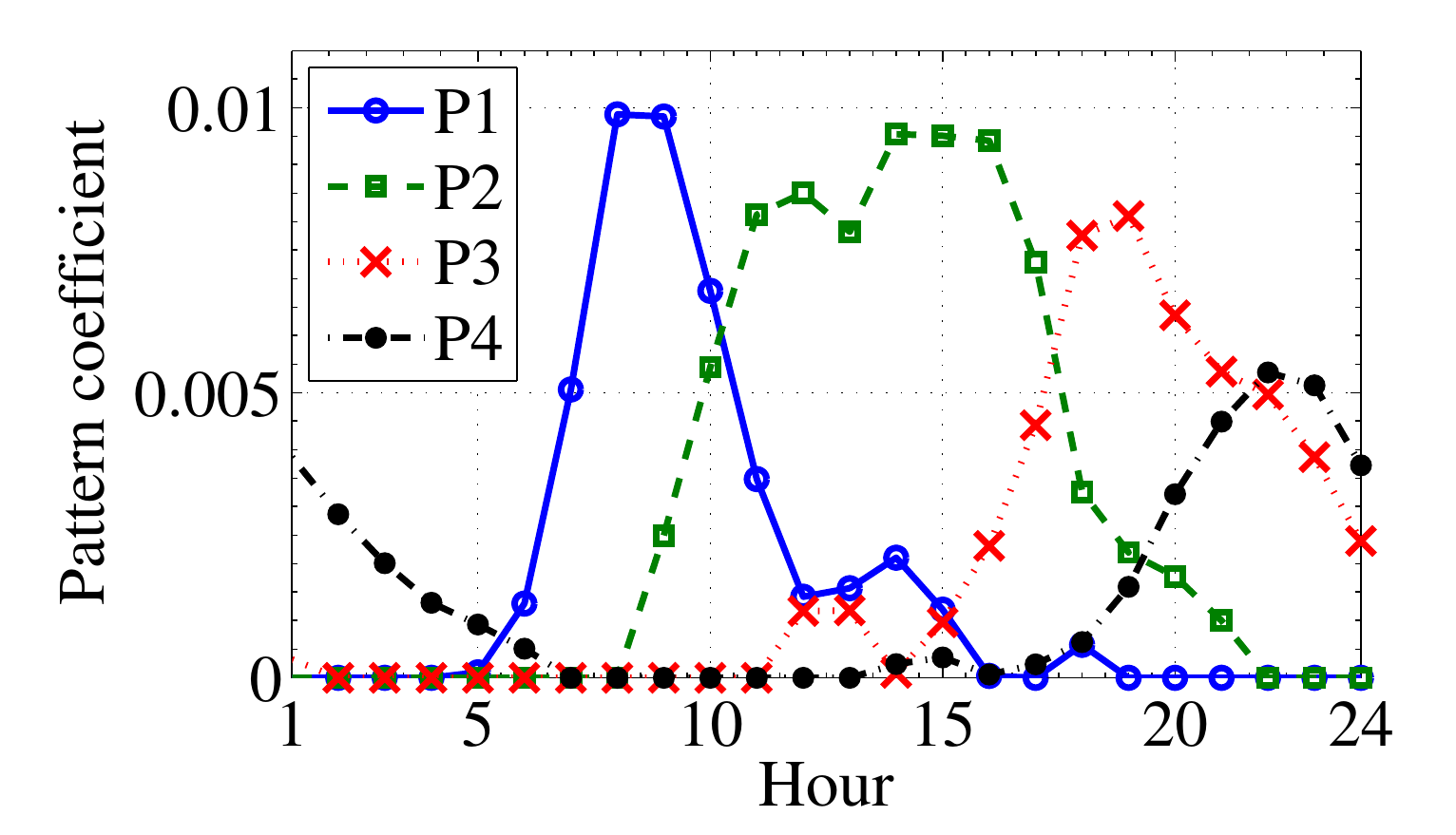}}	\subfigure[2015]{\label{fig:T2015}\includegraphics[width=0.47\columnwidth]{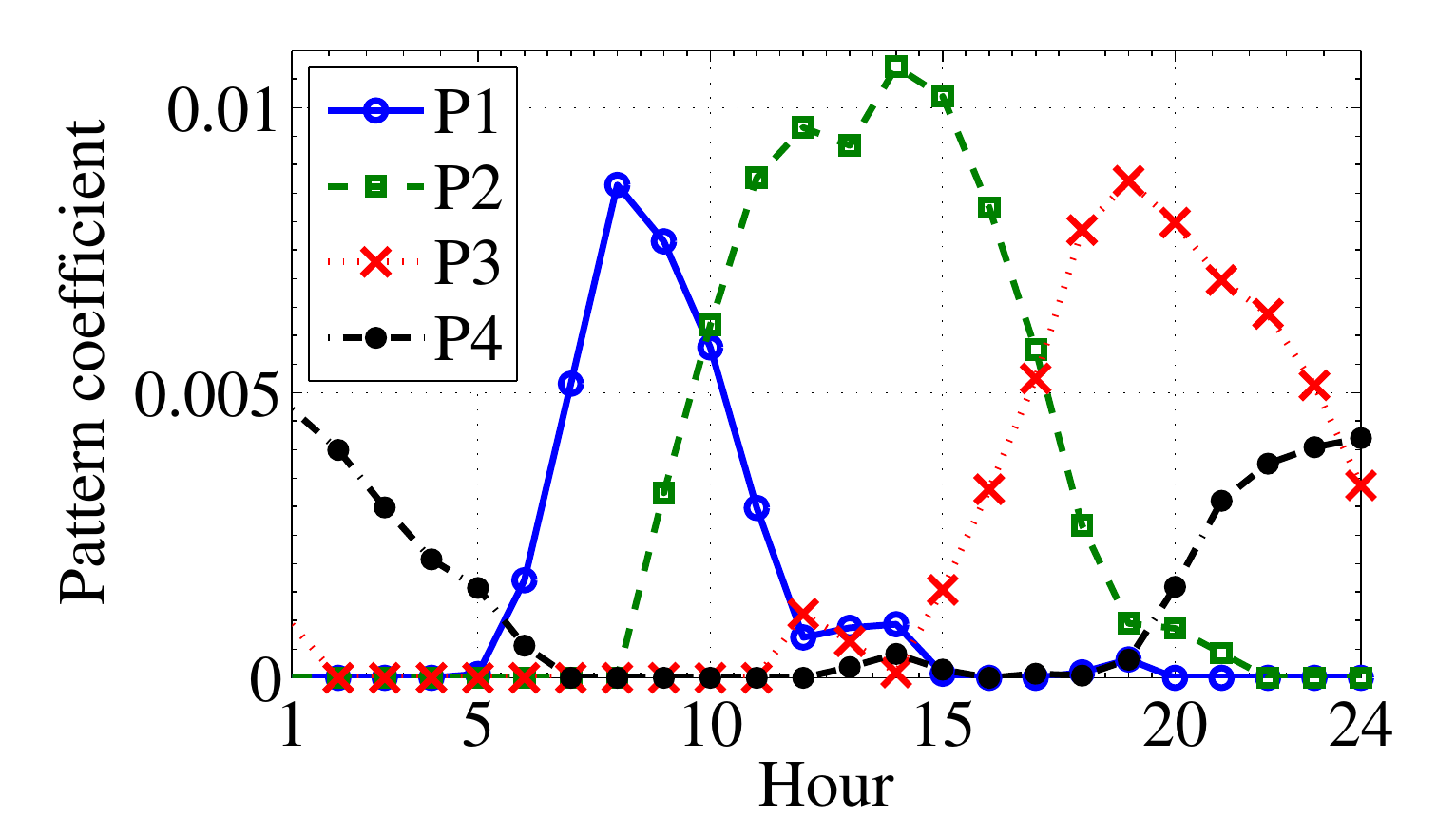}}
	\end{center}
%	\vspace{-0.4cm}
	\caption{Temporal patterns in 2008 and 2015.}
%	\vspace{-0.2cm}
	\label{fig:temporal_patterns}
\end{figure}

\subsubsection{Setting of Tradeoff Parameters}

In NR-cNTF, the tradeoff parameters $\alpha$ and $\beta$ are for adjusting the strength of urban context terms, and $\gamma$, $\delta$ and $\epsilon$ for adjusting the strength of sparsity regularization terms. %If the tradeoff parameters are set too small, the performance of our model will be close to normal NTF. On the contrary, if the parameters are relatively large, the optimization in Equation (\ref{eq:main_obj}) may be dominated by the regularization terms, therefore the reconstructed loss term is not properly optimized.
In our experiment, we set the tradeoff parameters using a traverse approach. We vary $\alpha$ and $\beta$ from 0 to 0.05 and $\gamma$, $\delta$ and $\epsilon$ from 0.1 to 10, respectively, aiming to choose the parameters with the best performances. Fig.~\ref{fig:lambda_vary} exhibits the experimental reconstruction errors with different tradeoff parameters, where each point is averaged on 10 runs. As can be seen, the best performance appears when $\alpha=\beta=0.01$ and $\gamma=\delta=\epsilon = 2.5$, which become the default settings.

\subsection{Discovery of Temporal Patterns}

\begin{figure*}[t]\centering
	\begin{center}\centering
		\subfigure[Morning Peak]{\label{fig:morning} \includegraphics[width=0.47\columnwidth]{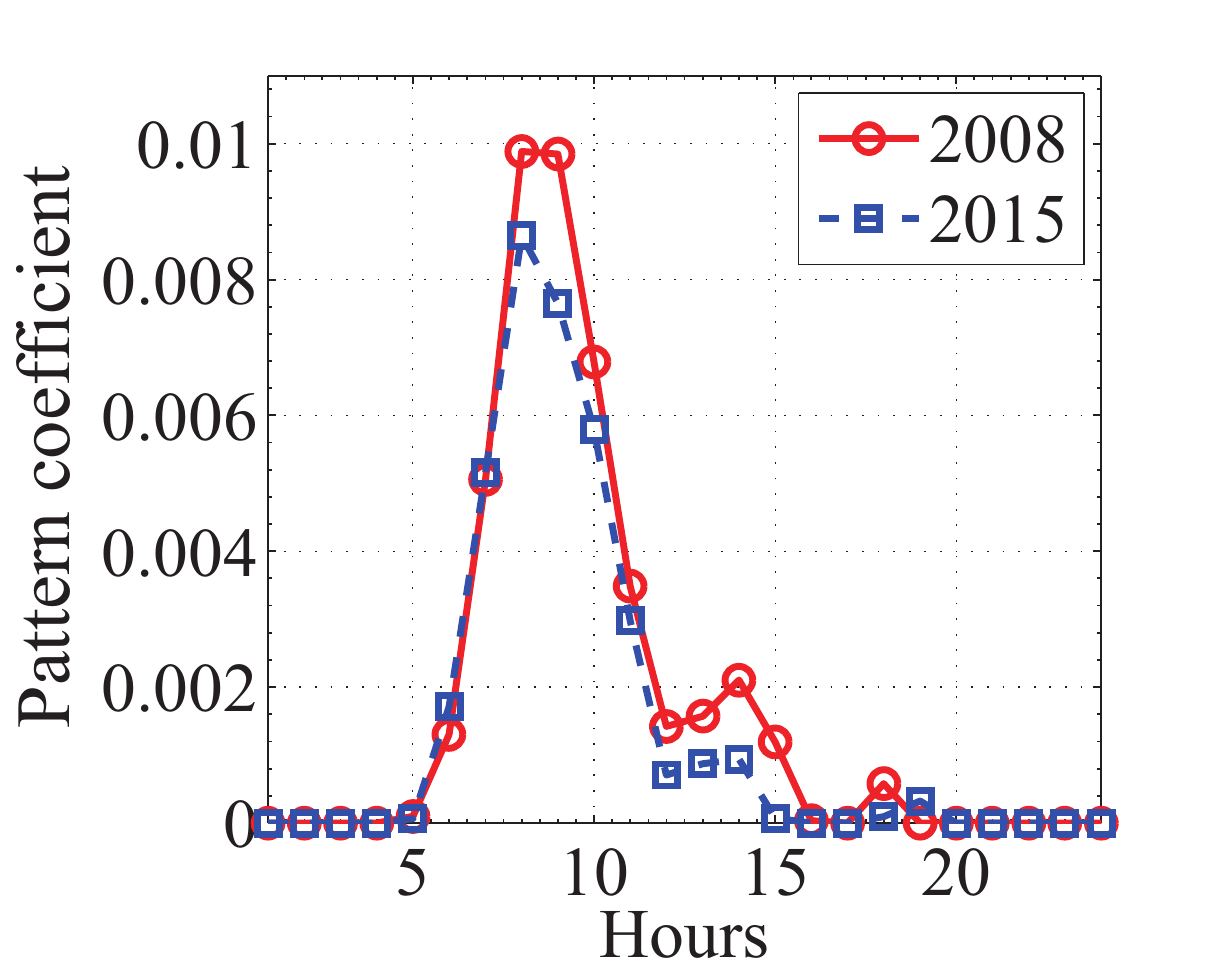}}~ \subfigure[Midday]{\label{fig:midday}\includegraphics[width=0.47\columnwidth]{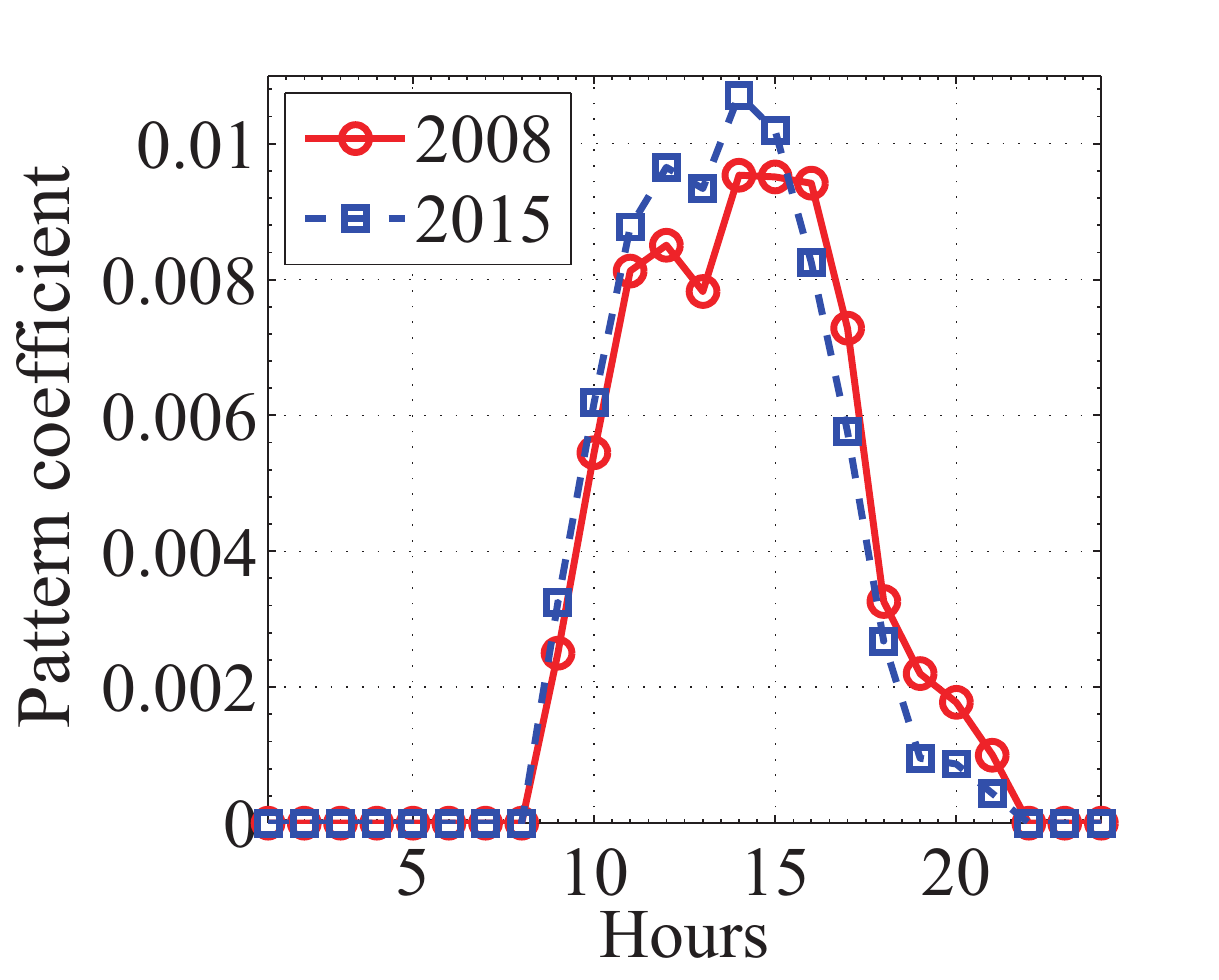}}~
		\subfigure[Evening Peak]{\label{fig:evening} \includegraphics[width=0.47\columnwidth]{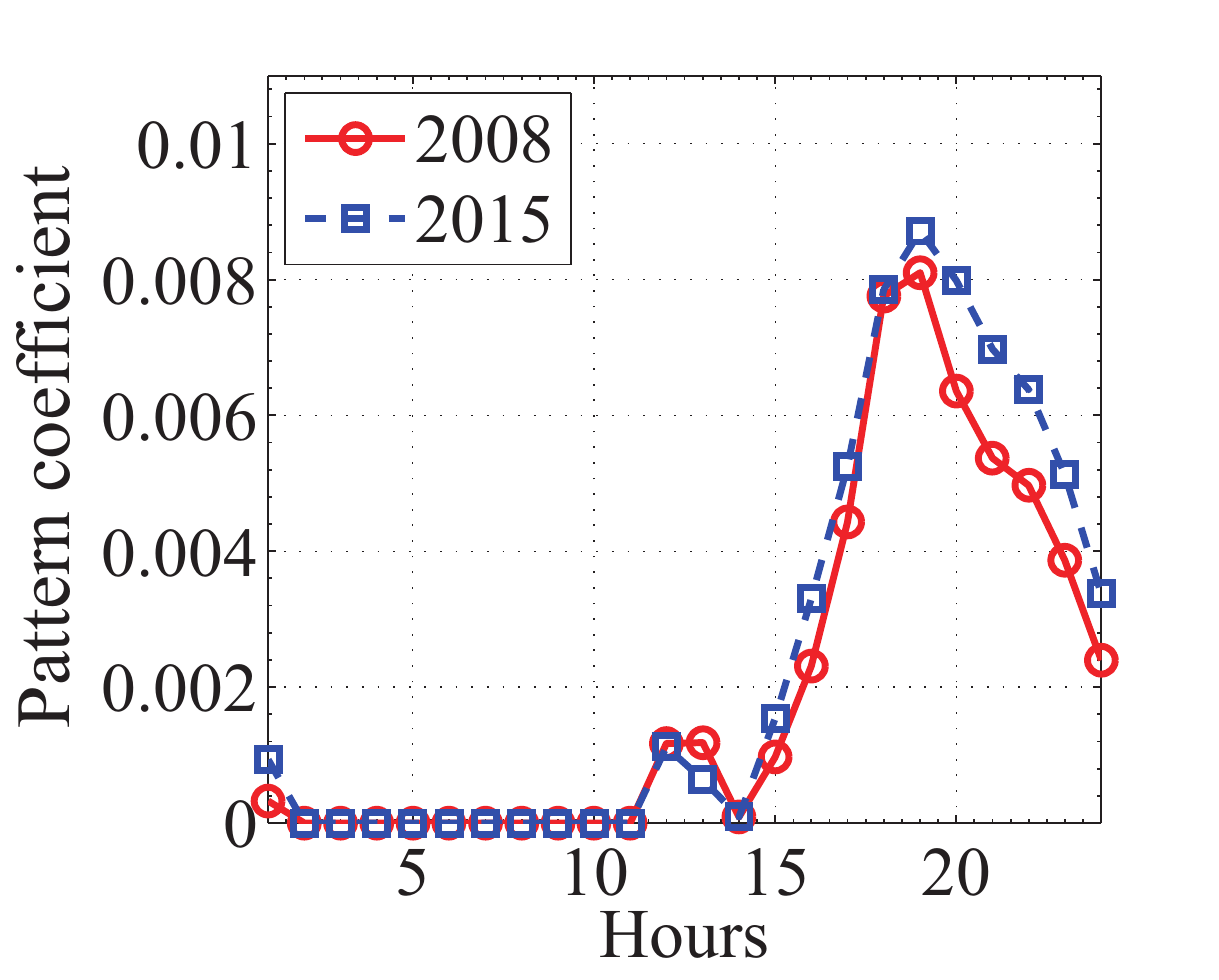}}~ \subfigure[Night]{\label{fig:night}\includegraphics[height=35mm,width=0.42\columnwidth]{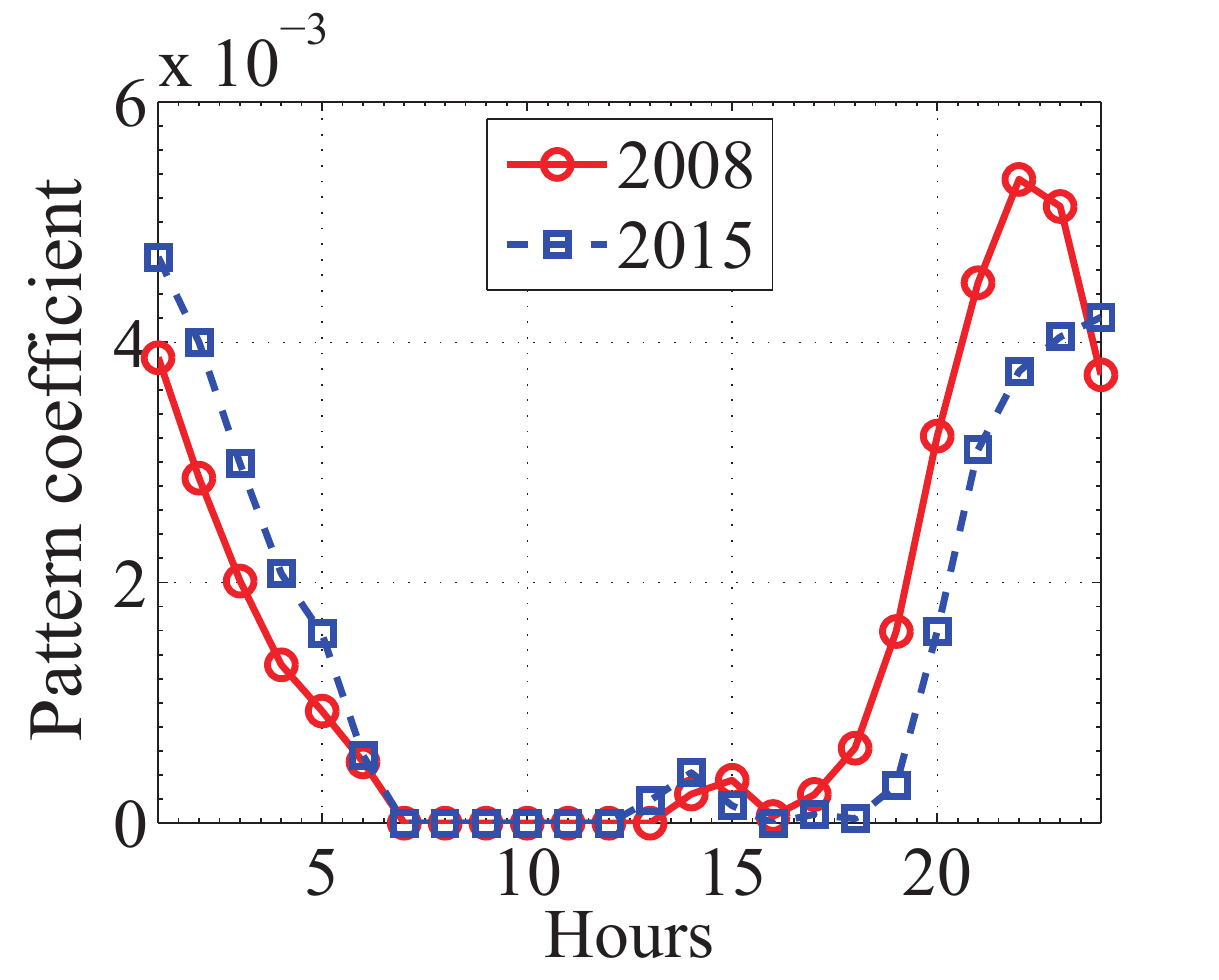}}~
	\end{center}
	\caption{The temporal patterns comparison between 2008 and 2012.}
	\label{fig:temporal_pattern_comparison}
\end{figure*}

{\revision Here, we describe the temporal patterns discovered from Beijing taxi traffic in 2008 and 2015. To facilitate comparison, we first introduce a normalization scheme to the projection matrix $\mathbf{T}$. Specifically, for the $k$-th pattern, we define a mask matrix as $\mathbf{Y}^k\in \mathbb{R}^{N\times K}$, where the element $y^k_{xi} = 1$ when $i = k$, and $0$ otherwise. We use the mask matrix to construct a data tensor as
\begin{equation}\label{eq:component}
\boldsymbol{\mathcal{\tilde{R}}}^k = \boldsymbol{\mathcal{C}} \times_{o} \mathbf{O} \times_{d} \mathbf{D} \times_{t}  \left(\mathbf{T} \odot \mathbf{Y}^k \right).
\end{equation}
In Eq.~\eqref{eq:component}, the elements of $\mathbf{T}$ corresponding to the patterns $\neg k$ are multiplied by zero, so $\boldsymbol{\mathcal{\tilde{R}}}^k$ only contains the components of the pattern $k$. Therefore, the physical meaning of $\boldsymbol{\mathcal{\tilde{R}}}^k$ is a component tensor corresponding to the $k$-th temporal pattern of the data tensor $\tensor{R}$.  Using $\boldsymbol{\mathcal{\tilde{R}}}^k$, we define the {\em energy} of the temporal pattern $k$ as
\begin{equation}%\small
u_k = \frac{\|\boldsymbol{\mathcal{\tilde{R}}}^k\|_1}{M \times M \times N} = \frac{\sum_{x=1}^M \sum_{y=1}^M \sum_{z=1}^N |\tilde{r}_{xyz}^k| }{M \times M \times N}.
\end{equation}
The physical meaning of the {\em energy} $u_k$ is a normalized size of the components corresponding to the temporal pattern $k$.

In the experiments, we define the re-scaled pattern coefficient $\tilde{t}_{zk}$ as
\begin{equation}\label{eq:coefficient}%\small
\tilde{t}_{zk} = {\frac{t_{zk}}{\sum_{n=1}^{N} t_{nk}}} \times u_k.
\end{equation}
The physical meaning of $\tilde{{t}}_{zk}$ is the energy of the temporal pattern $k$ at the time slice $z$. The vector $\tilde{\mathbf{t}}_{:k}$ is the distribution of $u_k$ over the $N$ time slices, and $\sum_{z=1}^N \tilde{t}_{zk} = u_k$. We compare the re-scaled pattern coefficients of different years to demonstrate the changes of temporal patterns of resident mobility from 2008 and 2015.}

\begin{figure*}[t]\centering
	\begin{center}\centering
		%\subfigure[2008 Origin]{\label{fig:O08} \includegraphics[width=0.44\columnwidth]{fig/O2008-eps-converted-to.pdf}}~~~~
		\subfigure[2008 DSP's by NR-cNTF] {\label{fig:D08}\includegraphics[width=0.5\columnwidth]{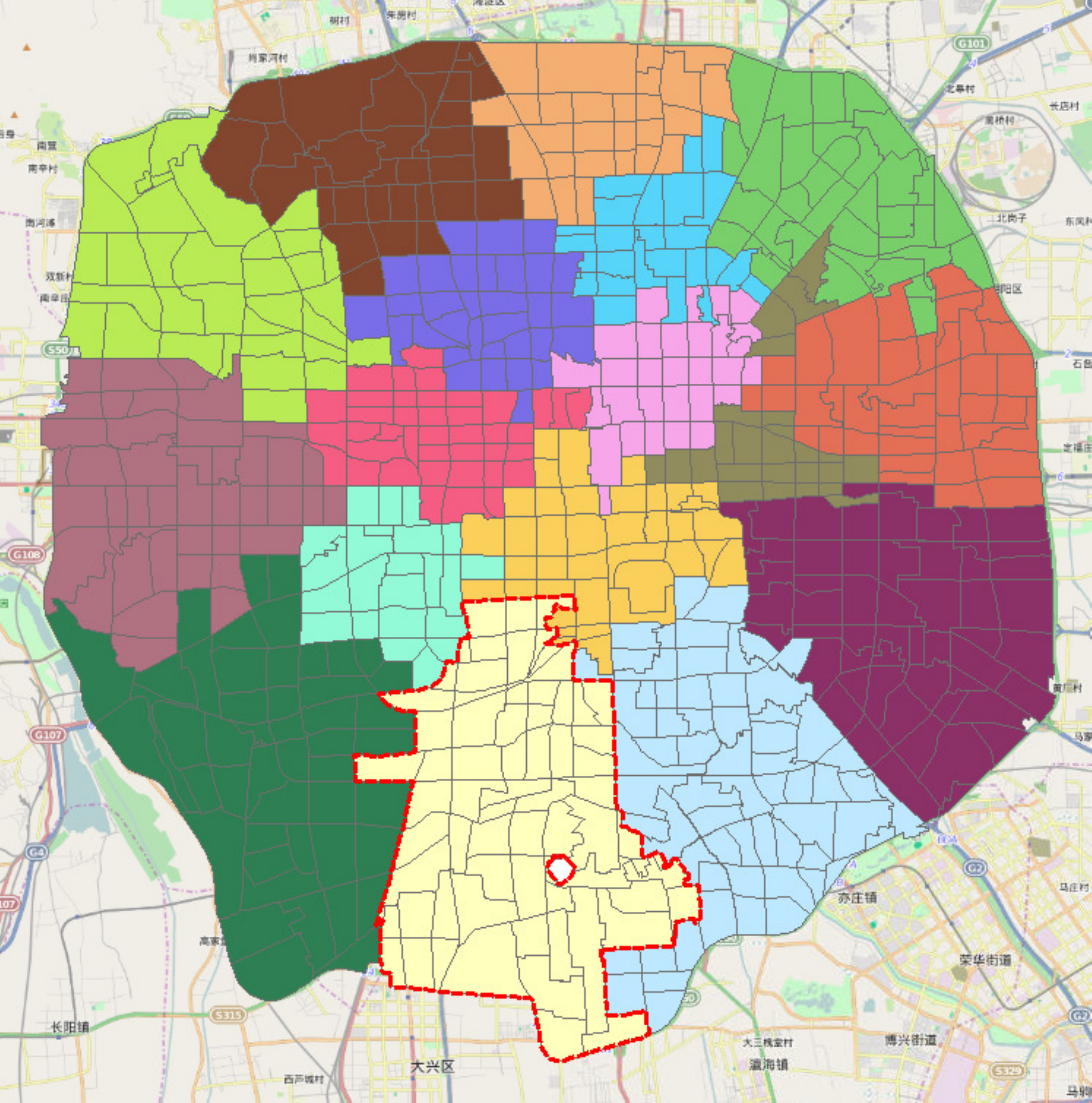}}~~~~~~
		%\subfigure[2015 Origin]{\label{fig:O15} \includegraphics[width=0.45\columnwidth]{fig/O2015-eps-converted-to.pdf}}~~~~
		\subfigure[2015 DSP's by NR-cNTF] {\label{fig:D15}\includegraphics[width=0.5\columnwidth]{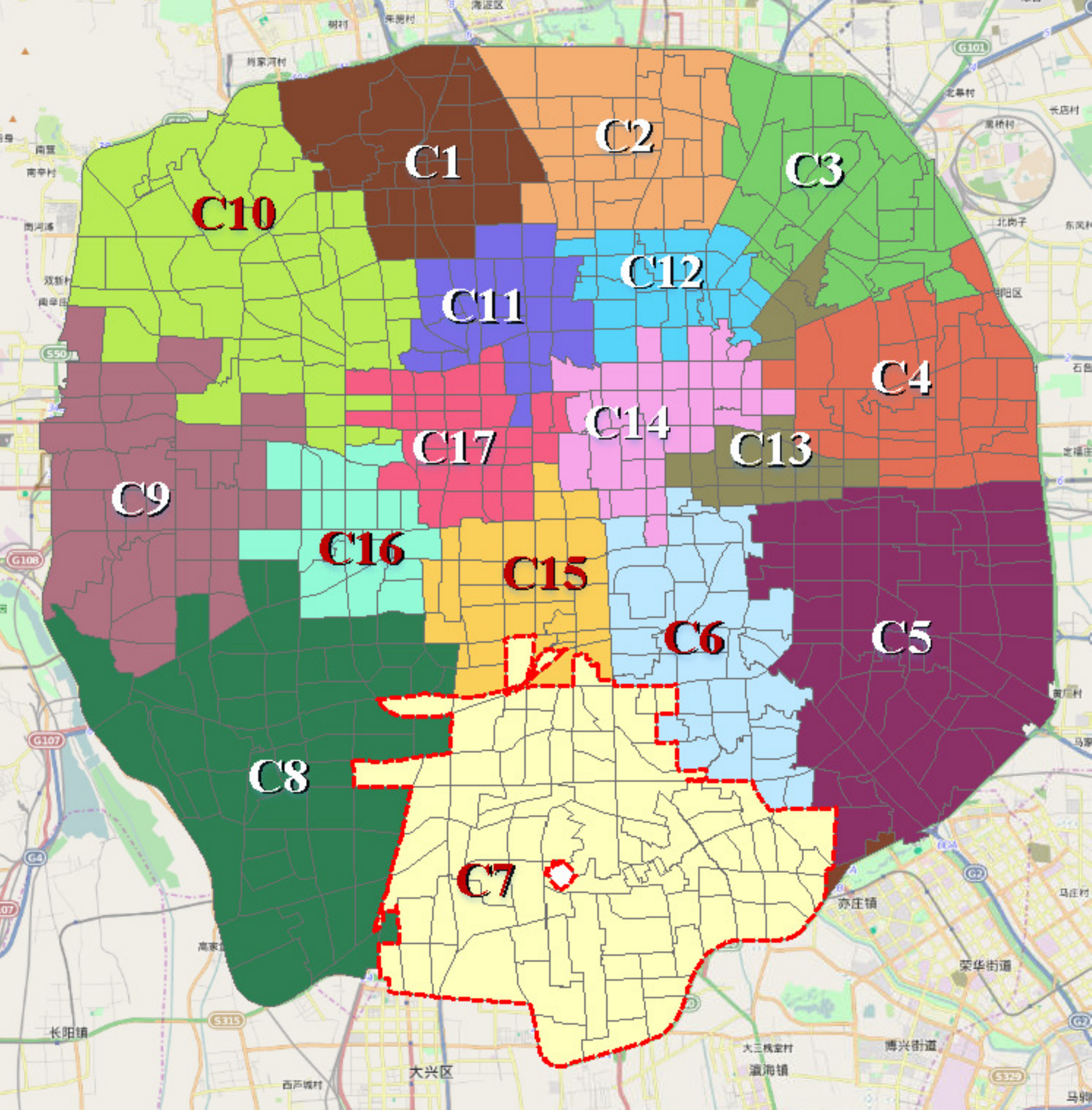}}~~~~~~
		\subfigure[2008 DSP's by cNTF] {\label{fig:2008D_cntf}\includegraphics[width=0.55\columnwidth]{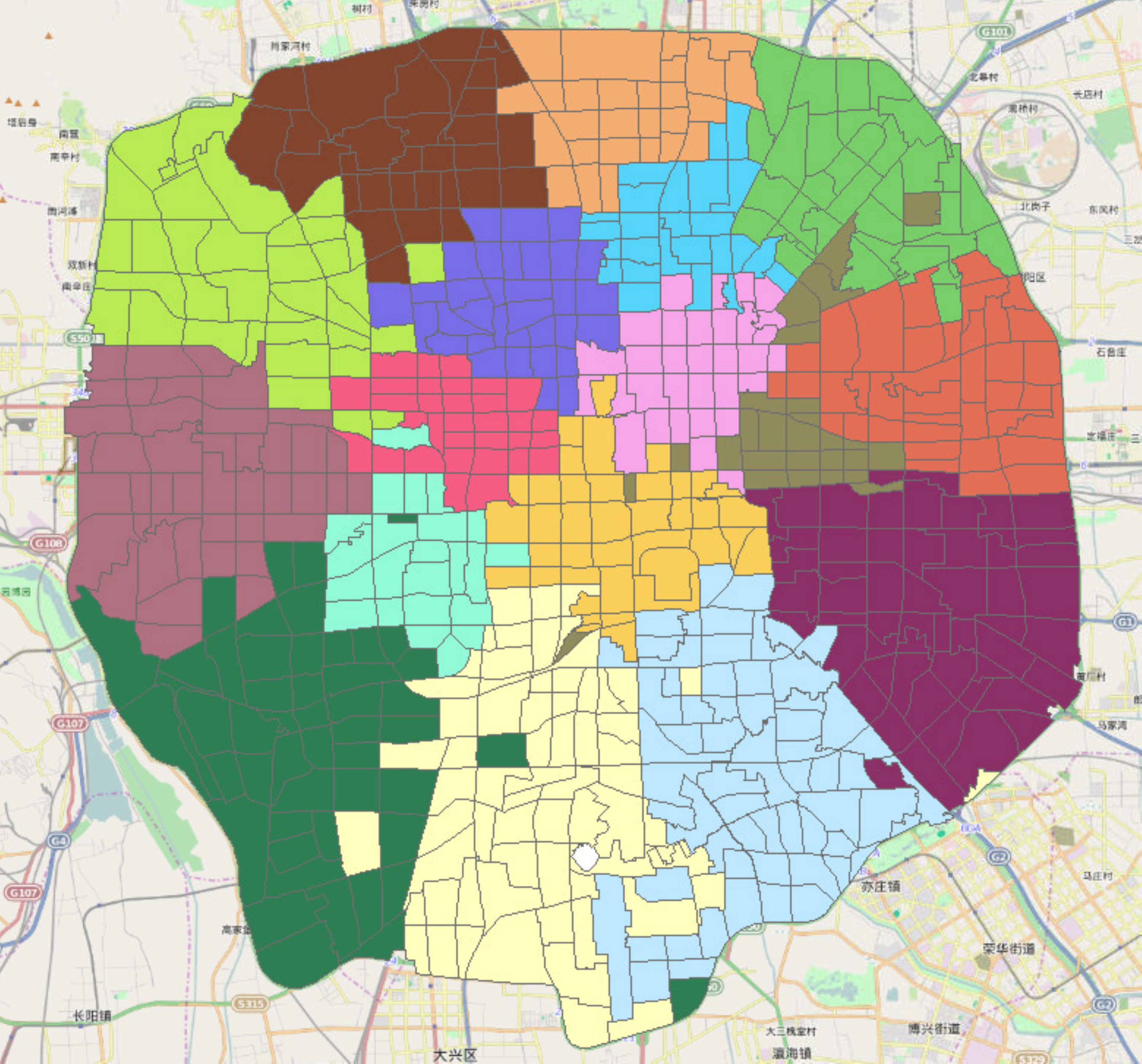}}
	\end{center}
%	\vspace{-0.3cm}
	\caption{Destination spatial patterns in 2008 and 2015.}
	\label{fig:spatial_patterns}
%	\vspace{-0.2cm}
\end{figure*}

%To facilitate comparison, we first introduce a normalization scheme to the projection matrix $\mathbf{T}$. That is, the re-scaled pattern coefficient $t_{zk}$ to
%\begin{equation}\small
%\tilde{t}_{zk} = \frac{t_{zk}}{\sum_{r=1}^{N} t_{rk}} \times u_k,
%\end{equation}
%where $u_k$ is the {\em energy} of the temporal pattern $\mathbf{t}_{:k}$ as
%\begin{equation}\small
%u_k = \frac{\sum_{x=1}^M \sum_{y=1}^M \sum_{z=1}^N \tilde{x}_{xyz}^k }{M \times M \times N}.
%\end{equation}
%$\tilde{x}^k_{xyz}$ is the ($x,y,z$) element of the reconstructed tensor $\boldsymbol{\mathcal{\tilde{X}}}^k = \boldsymbol{\mathcal{C}} \times_{o} \mathbf{O} \times_{d} \mathbf{D} \times_{t}  \left(\mathbf{T} \odot \mathbf{Y}_k \right)$, where $\mathbf{Y}_k\in \mathbb{R}^{N\times K}$ is a mask matrix with an all-one vector as the $k$th column and all-zero vectors as the other columns. {\color{blue} More detail explanation about {\em energy} and the re-scaled pattern coefficient of temporal patterns could be found in Section 4 of the supplementary materials.}

Fig.~\ref{fig:temporal_patterns} shows the four temporal patterns, which indeed correspond to four rhythms of urban traffic:
\begin{itemize}
	\item {\em P1: Morning Peak}, with an active range roughly from 6:00 to 11:00.
	\item {\em P2: Midday}, with an active range roughly from 9:00 to 18:00.
	\item {\em P3: Evening Peak}, with an active range roughly from 16:00 to 24:00.
	\item {\em P4: Night}, with an active range roughly from 20:00 to 3:00 of the next day.
\end{itemize}

To further reveal the evolution of temporal patterns from 2008 to 2015, we plot comparative diagram for each pattern of the two yeas in Fig.~\ref{fig:temporal_pattern_comparison}. The first observation is that the intensity of the morning pattern was decreased significantly from 2008 to 2015 (see Fig.~\ref{fig:morning}), whereas the evening pattern seems much more stable (see Fig.~\ref{fig:evening}). We believe the reduction of the morning peak via taxies is due to the rapid development of the metro system in Beijing. During the period from 2008 to 2015, the Beijing metro increased the mileage from 198km to 631km, which is particularly suitable for the time-rigid morning commute but has less impact to the evening commute with relatively flexible time.

Another observation is that the intensity of the midday pattern was increased during the seven years (see Fig.~\ref{fig:midday}). The main part of travel volume in the midday pattern consists of business travels from one workplace to another, whose destinations are random in essence and therefore cannot count heavily on public transportation systems like metros. Moreover, the fast-rising income in China in recent years might also contribute to the more spending on the relatively expensive taxi service.

The most interesting observation is that the peak time of the night pattern in 2015 came about two hours later than that in 2008 (Fig.~\ref{fig:night}). {\revision This implies that residents tend to have more travels in the midnight in recent years. The reasons behind this could be complicated, which might include some lifestyle changes in Beijing, such as the more colorful nightlife or the higher overtime working pressures.}

To sum up, the NR-cNTF model well captures the temporal patterns hidden inside the Beijing taxi traffic. The evolution of these patterns further unveils the development of Beijing metros and the changes of lifestyle.

\begin{figure*}[t]\centering
	\begin{center}\centering
		\subfigure[2008 Morning Peak]{\label{fig:c08-1} \includegraphics[width=0.4\columnwidth]{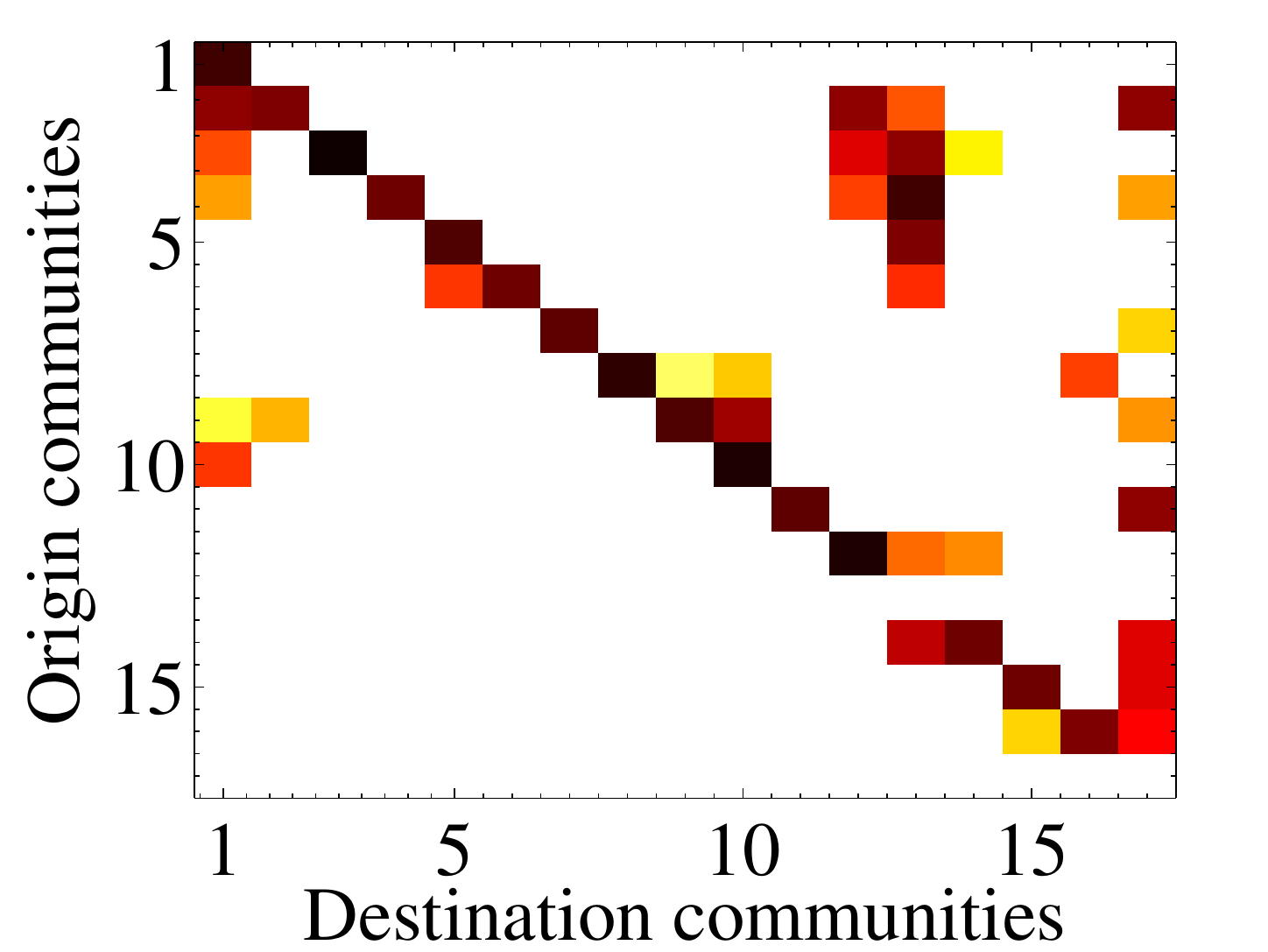}}
		\subfigure[2008 Midday]{\label{fig:c08-2}
			\includegraphics[width=0.4\columnwidth]{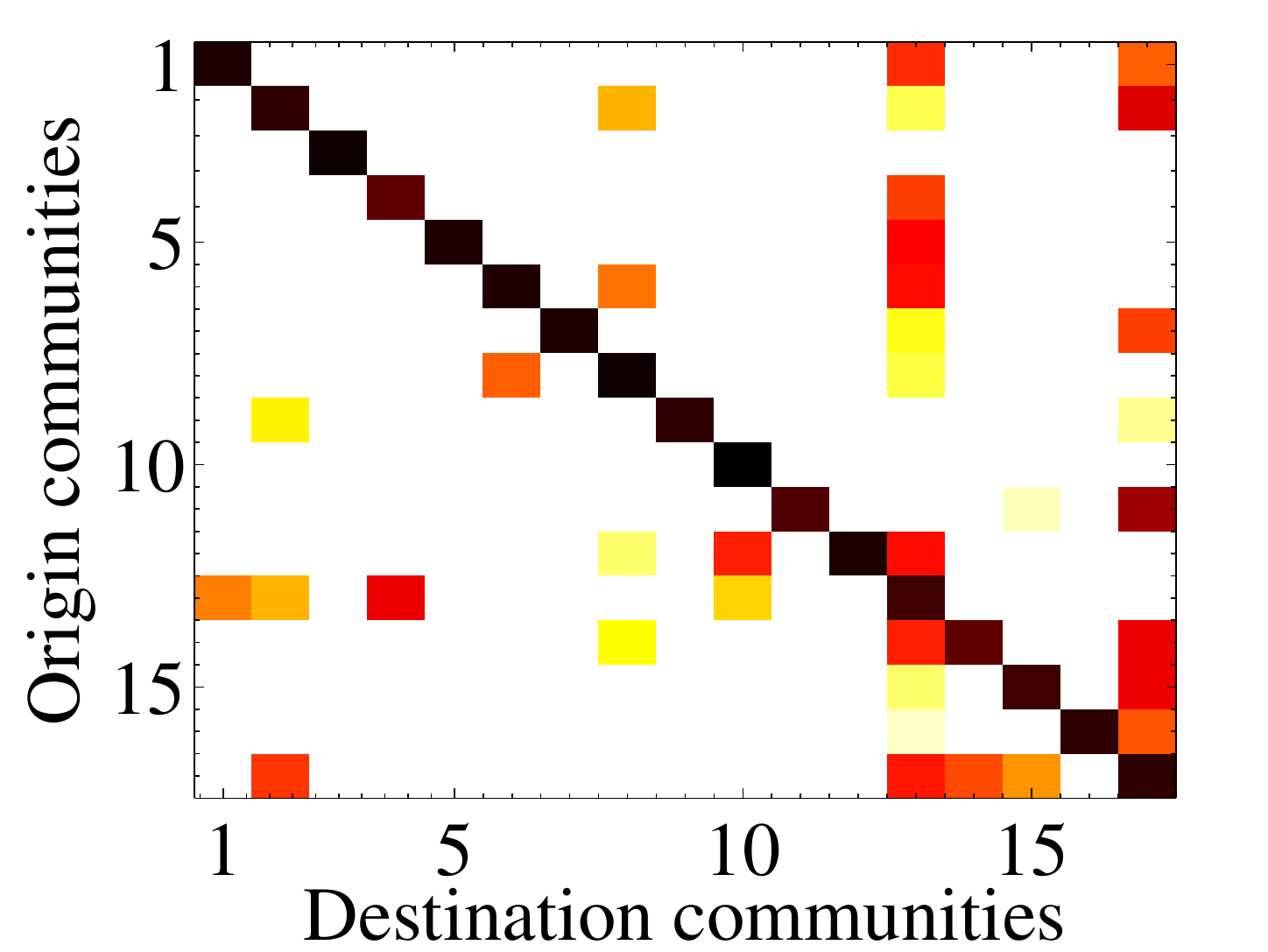}}
		\subfigure[2008 Evening Peak]{\label{fig:c08-3} \includegraphics[width=0.4\columnwidth]{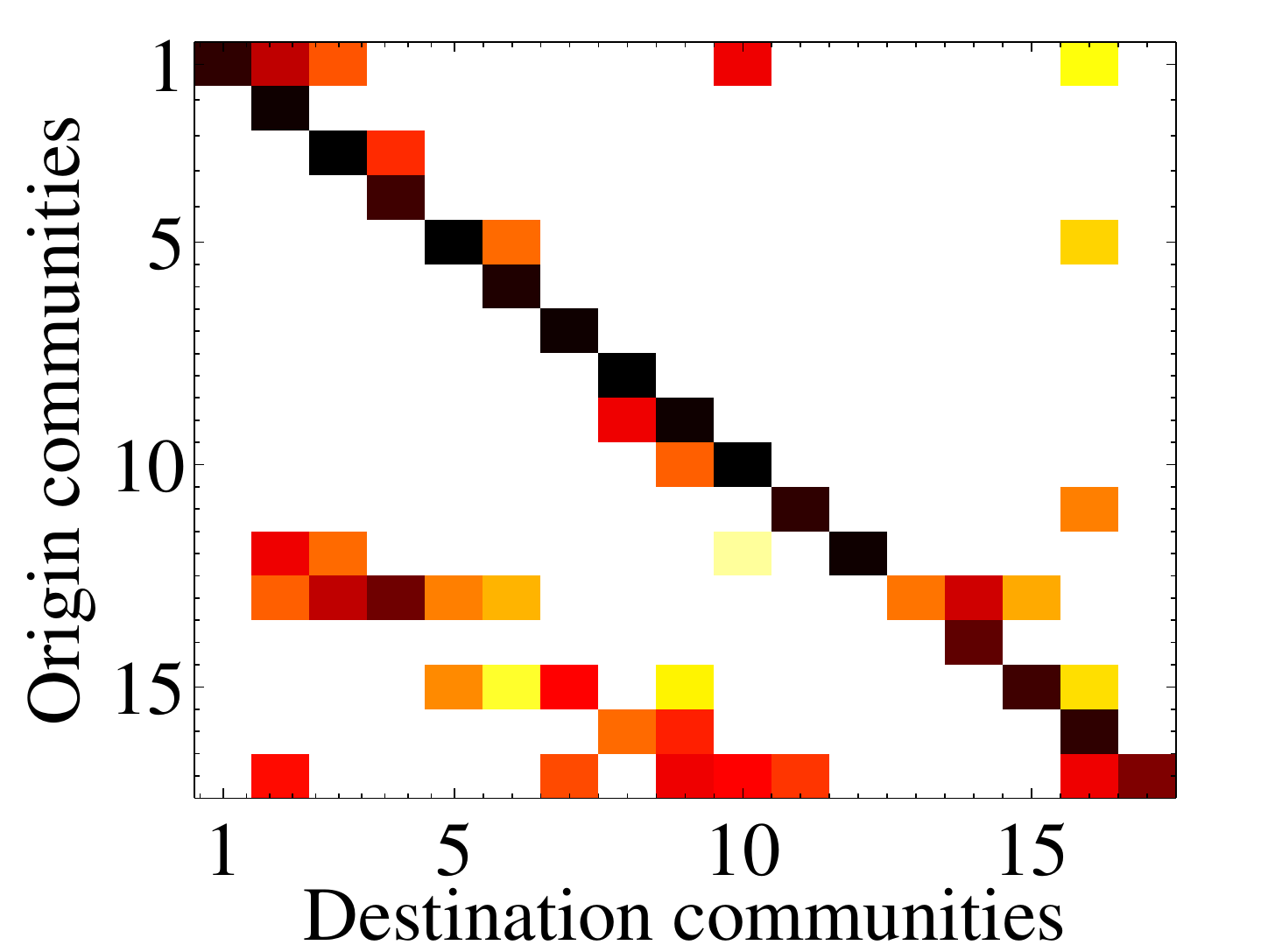}}
		\subfigure[2008 Night]{\label{fig:c08-4}
			\includegraphics[width=0.4\columnwidth]{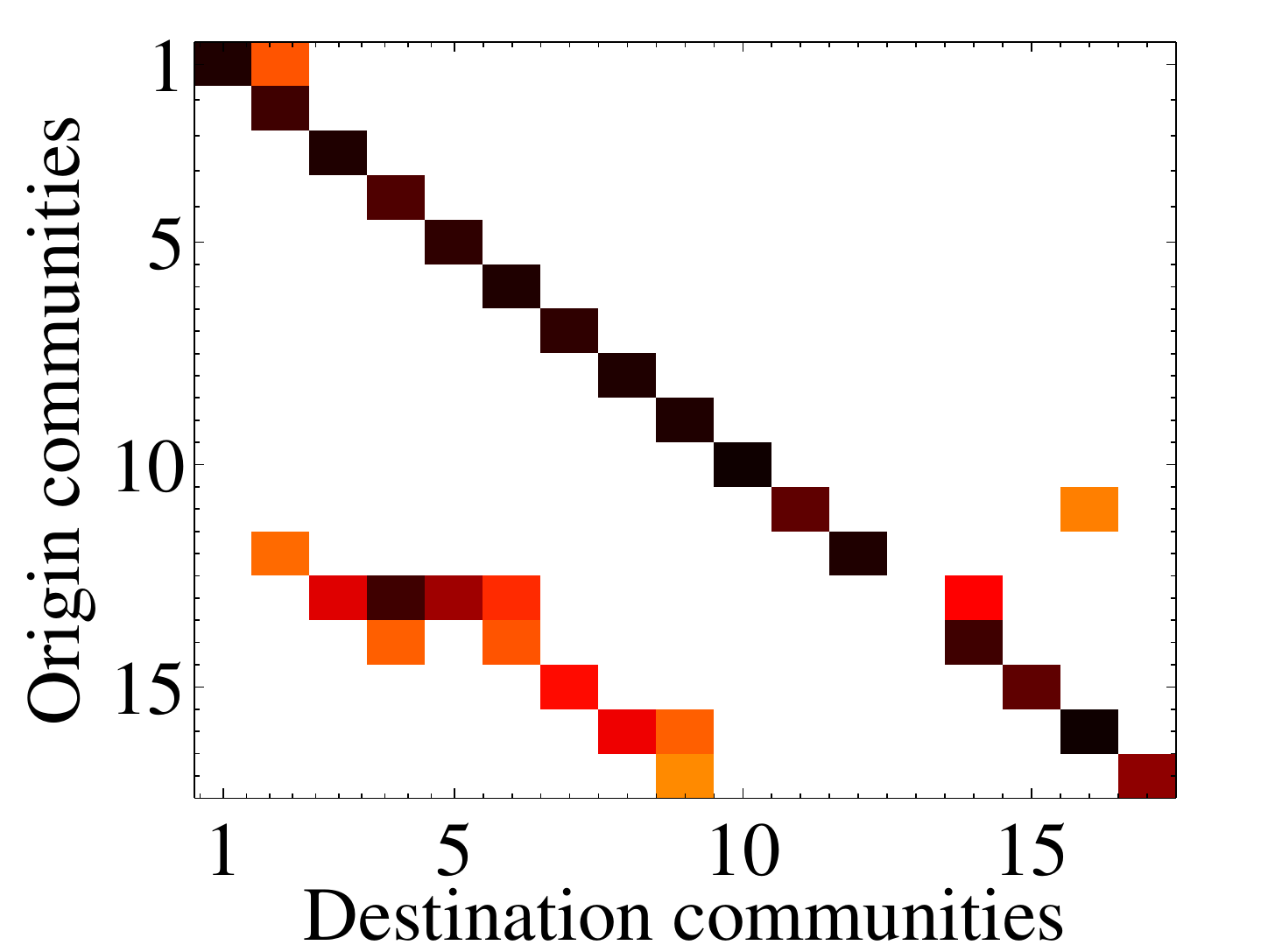}}\\
		\subfigure[2015 Morning Peak]{\label{fig:c015-1} \includegraphics[width=0.4\columnwidth]{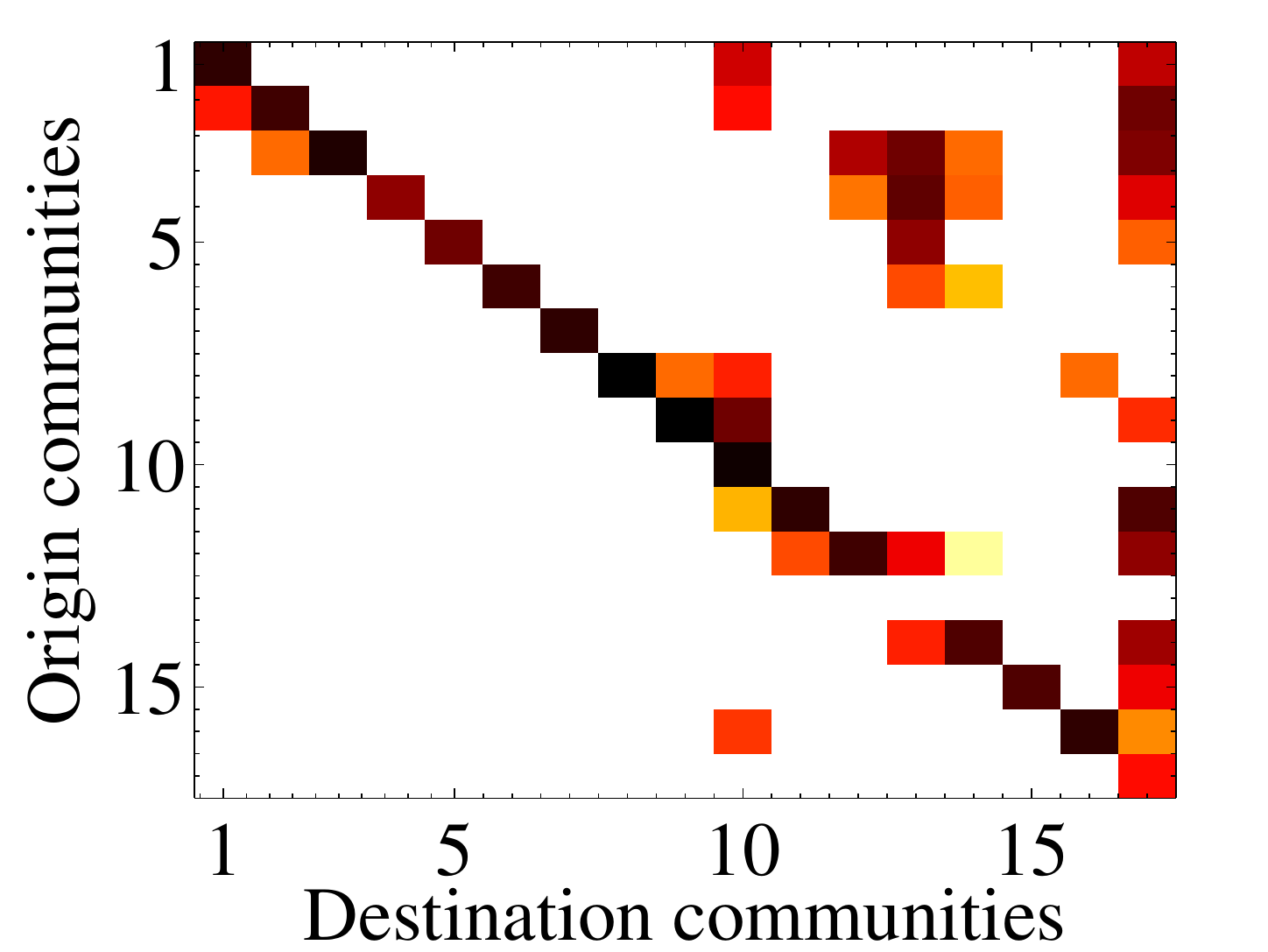}}
		\subfigure[2015 Midday]{\label{fig:c015-2}
			\includegraphics[width=0.4\columnwidth]{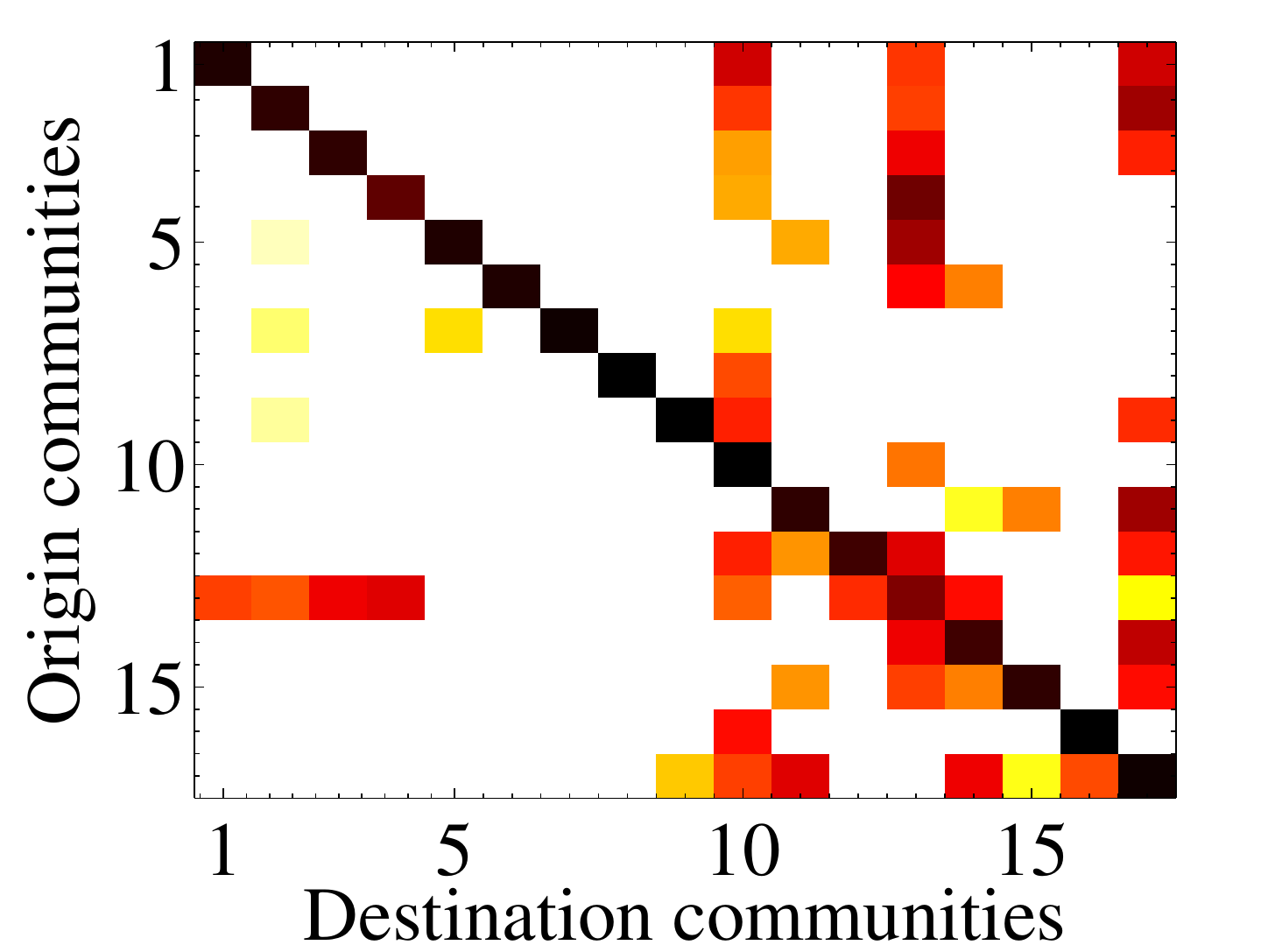}}
		\subfigure[2015 Evening Peak]{\label{fig:c015-3} \includegraphics[width=0.4\columnwidth]{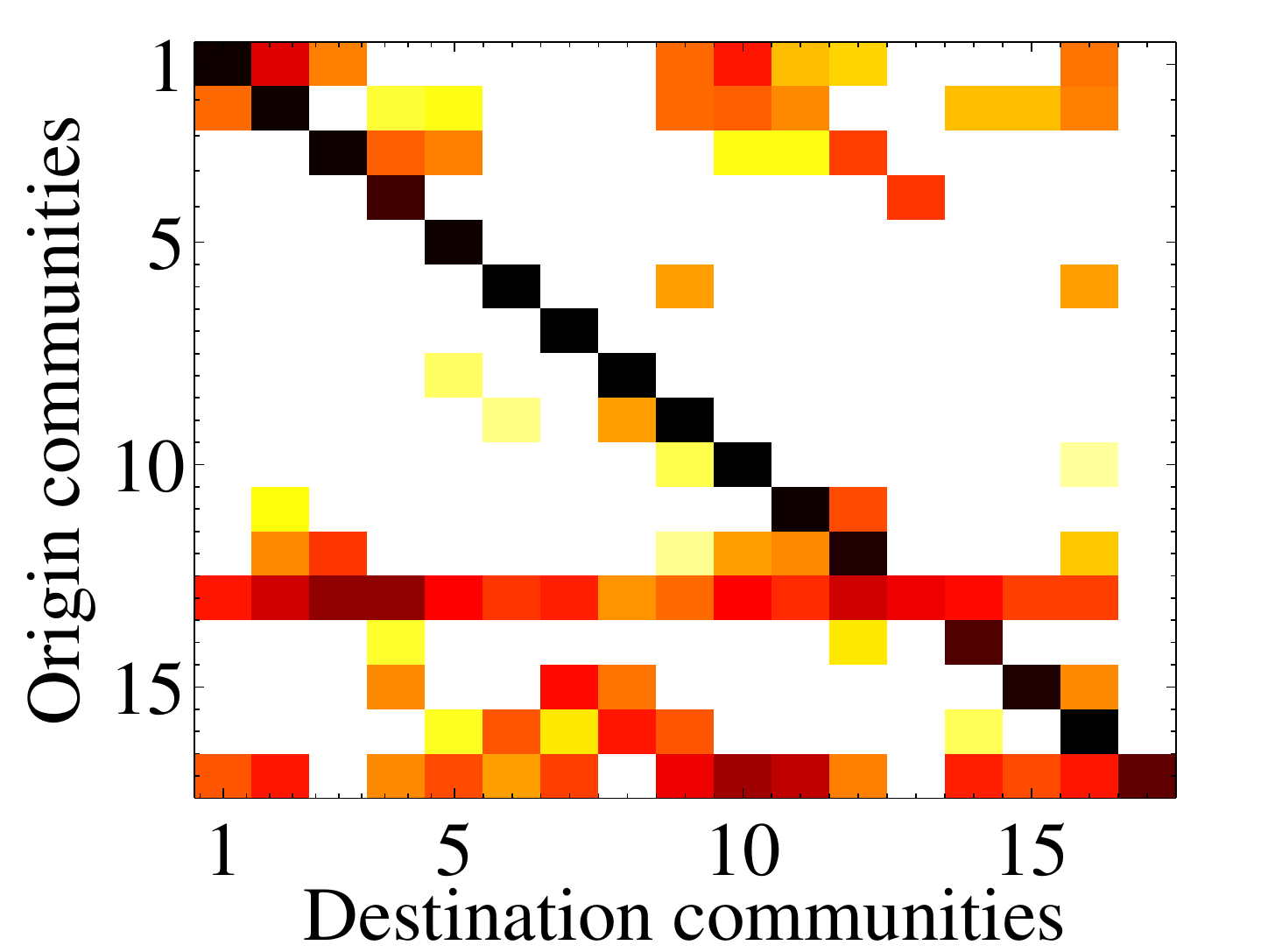}}
		\subfigure[2015 Night]{\label{fig:c015-4}
			\includegraphics[width=0.4\columnwidth]{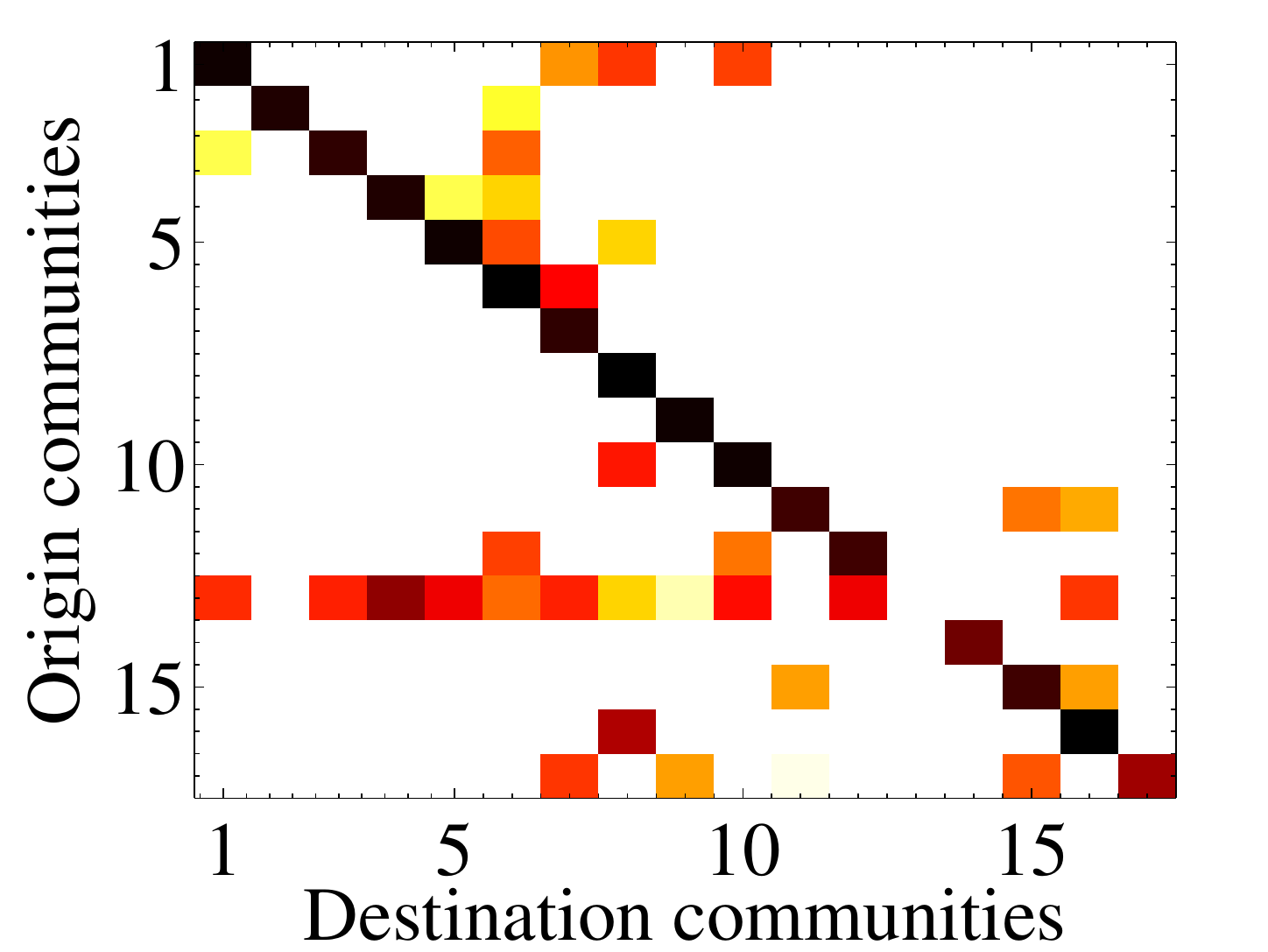}}
	\end{center}
%	\vspace{-0.3cm}
	\caption{Dynamic patterns in 2008 and 2015.}
%	\vspace{-0.2cm}
	\label{fig:core_patterns}
\end{figure*}

\subsection{Discovery of Spatial Patterns}
Here, we explore the spatial patterns discovered by NR-cNTF. Given any origin or destination pattern $\mathbf{v}_{:i}$ (see Def.~1 in Sect.~\ref{subsec:pattern}), we first obtain the corresponding urban community $\mathcal{SP}_i$ (see Sect.~\ref{subsec:neighbor}). We adopt the ``crisp partition'' assumption so that an urban zone will be assigned to one and only one urban community. As a result, among the $I = J = 20$ patterns in our experiment, we obtain 17 urban communities, and the rest three are empty and omitted. Note that we only use destination spatial patterns (DSP) for illustration below. The origin spatial patterns have the similar results, we don't put them in the paper for concision.

%{\color{red}and put the similar results of origin spatial patterns to the supplemental material for concision.}

{
\begin{figure}[t!]
	\centering
	\subfigure[\revision The Ring Roads in Beijing]{\label{fig:ringroads}
	\includegraphics[width=0.4\columnwidth]{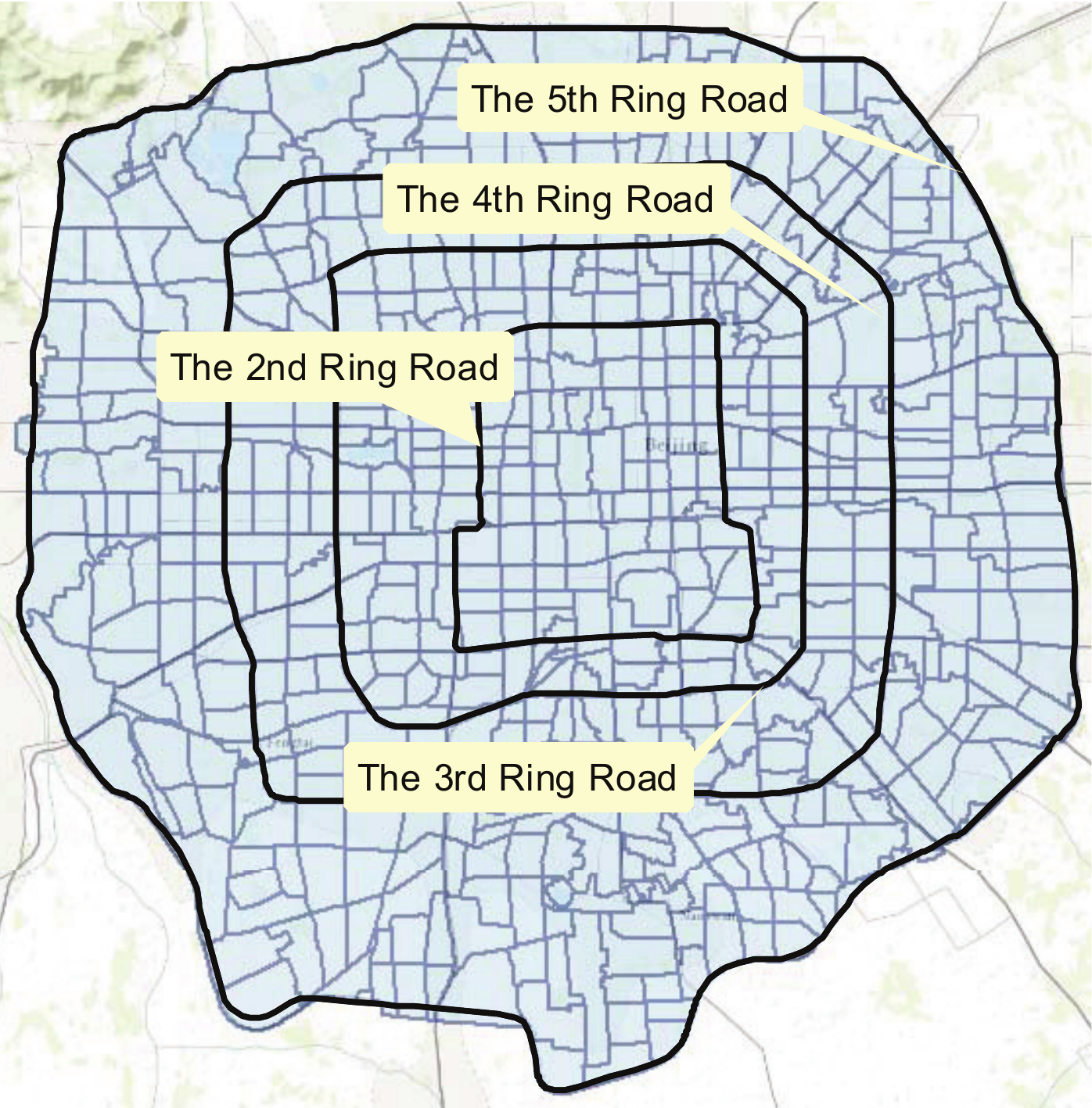}}~~~~~
	\subfigure[\revision The Trunk Roads in Beijing]{\label{fig:roadnet}
		\includegraphics[width=0.43\columnwidth]{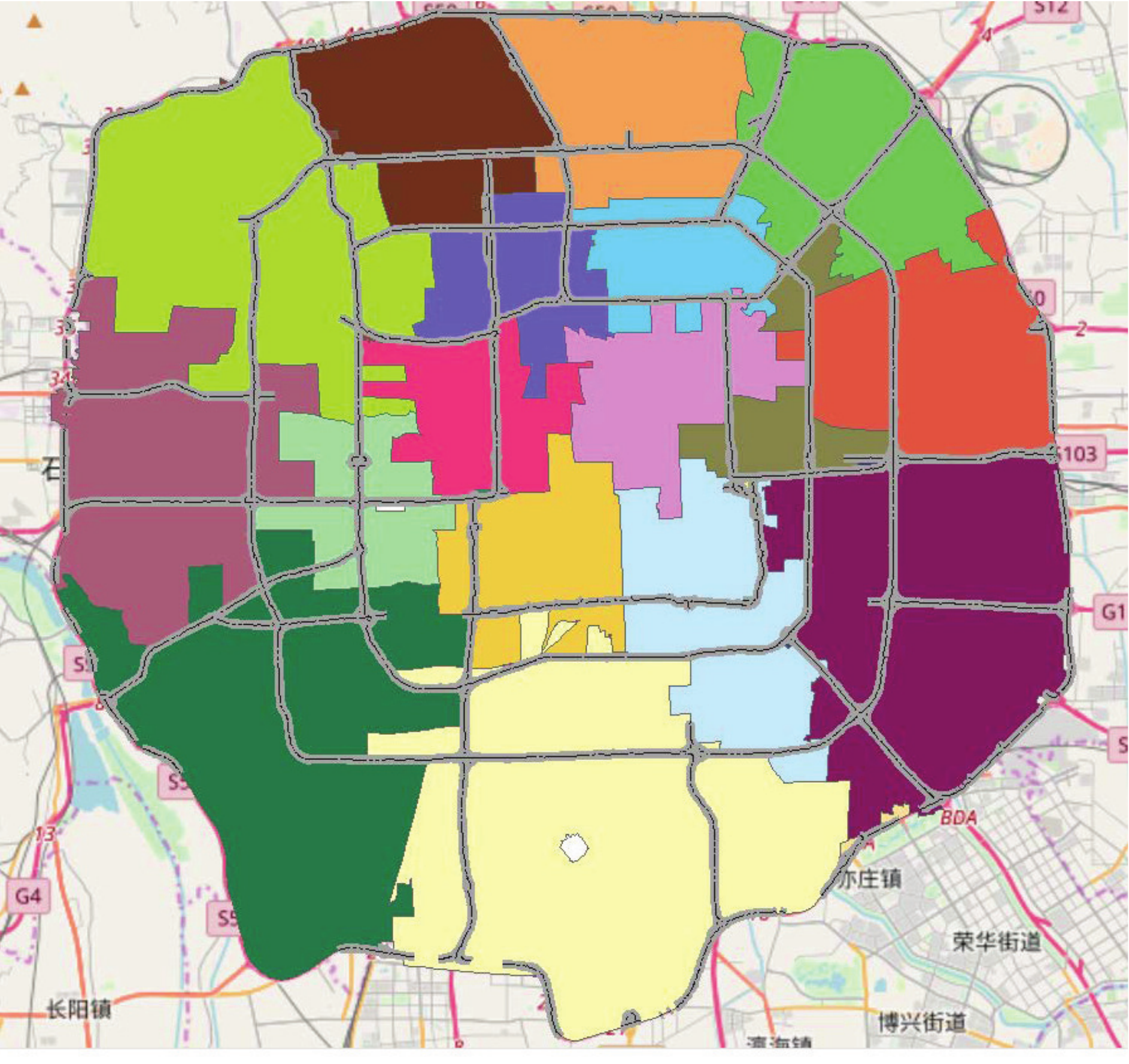}}
	\caption{\revision The urban communities and trunk roads in Beijing.}\label{fig:map}
\end{figure}
}

Fig.~\ref{fig:D08} and Fig.~\ref{fig:D15} visualize the urban communities corresponding to the destination spatial patterns found in 2008 and 2015, respectively. As can be seen, each urban community (filled with a same color) identified by NR-cNTF contains urban zones geographically adjacent to at least one zone in the same community, which agrees with our intuition about functional zoning of a city. In contrast, Fig.~\ref{fig:2008D_cntf} shows the 2008 urban communities found by cNTF without neighboring regulation, whose functionalities are less clear due to the geographical discontinuity. For the convenience of discussion, we numbered the communities in Fig.~\ref{fig:D15} from 1 to 17.

{\revision A general observation from Fig.~\ref{fig:spatial_patterns} is that the spatial communities of Beijing radially surround the center of Beijing. This character of spatial communities has close relations with the trunk road network structure of Beijing. Fig.~\ref{fig:ringroads} shows there are four concentric ring roads surrounding the center of Beijing. As reported in~\cite{beijing_ring_roads}, the ring roads provide a basic framework for the city's overall spatial pattern. Affected by the ring roads, we can see that the communities discovered in Fig.~\ref{fig:spatial_patterns} also constitute two concentric circles surrounding the center of the Beijing city. Specifically, the communities C1-C10 form the outer circle, and C11-C17 form the inner circle. Fig.~\ref{fig:roadnet} plots the trunk road network of Beijing over the communities, from which we can see that many boundaries of the communities overlap with the trunk roads, indicating that the spatial patterns of residential mobility in Beijing are deeply shaped by the urban trunk road network.}

{\revision Another observation from Fig.~\ref{fig:spatial_patterns} is the interesting evolution of some urban communities in recent years. Let us take a closer look on community C7 located in the south of Beijing, which has an obvious expansion trend from 2008 to 2015. That is, some urban zones that belonged to C6 in 2008 were ``absorbed'' by C7 in 2015. To understand this, we should trace back to the so-called South Beijing Development Plan (SBDP) issued in 2008, which is a government investment plan in south areas of Beijing, with an executive period from 2010 to 2015 and a total investment of nearly 62.9 billion USD (more information about SBDP could be found in {\em Supplementary Materials}). The purpose of SBDP is to narrow the development gap between the lagging-behind southern region and other areas of the city. It is interesting that the communities C6 and C7 are just in the investment region of the plan (see Fig.~2 in {\em Supplementary Materials} for the evidence). The evolution of C6 and C7 from 2008 to 2015 essentially reflects the great impact of huge economic investments to the real-life development of a city.}

%That is, after the 2008 Beijing Olympic Game, the Beijing government started an investment plan, {\it i.e.}, SBDP, to promote the development of south Beijing, which is far from the Olympic venues and therefore did not get enough government investments before 2008. In this plan, 30 billion dollars were to be invested to the C7 area to improve the infrastructures of this area.
%
%And the evolution of C7 from 2008 to 2015 right reflects the great appeal of huge economic investments. {\revision More information about SBDP could be found in the supplementary materials.}

To sum up, the above results justify the effectiveness of our NR-cNTF model in uncovering latent and geographically adjacent spatial patterns, as well as their inconspicuous evolutions in recent years.

\begin{figure*}[t]\centering
	\begin{center}\centering
		\subfigure[Inter-Community Traffic]{\label{fig:inter_traffic} \includegraphics[width=0.5\columnwidth]{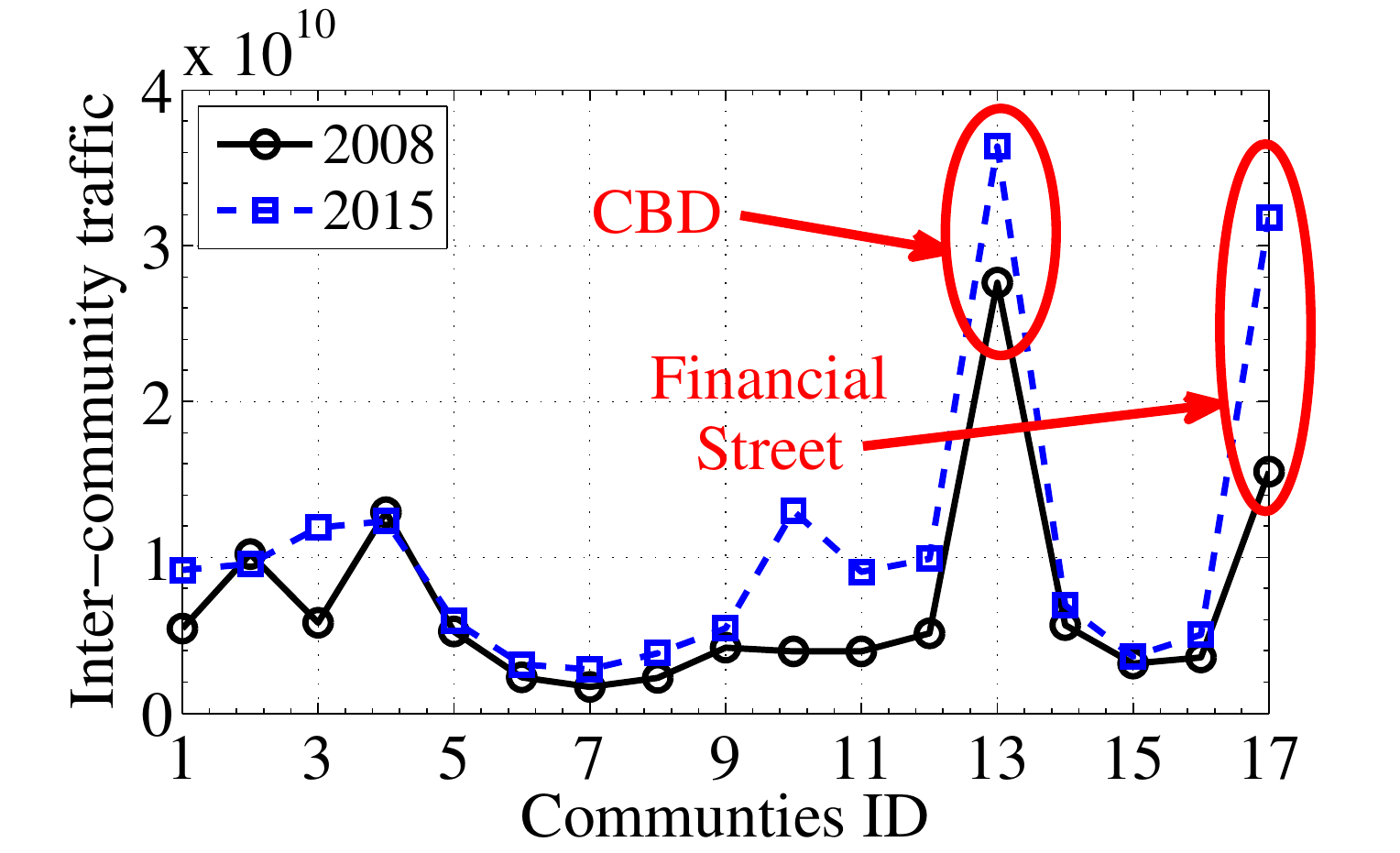}}~~~~~~~~
		\subfigure[Inter-Community Traffic Growth]{\label{fig:inter_growth}\includegraphics[width=0.5\columnwidth]{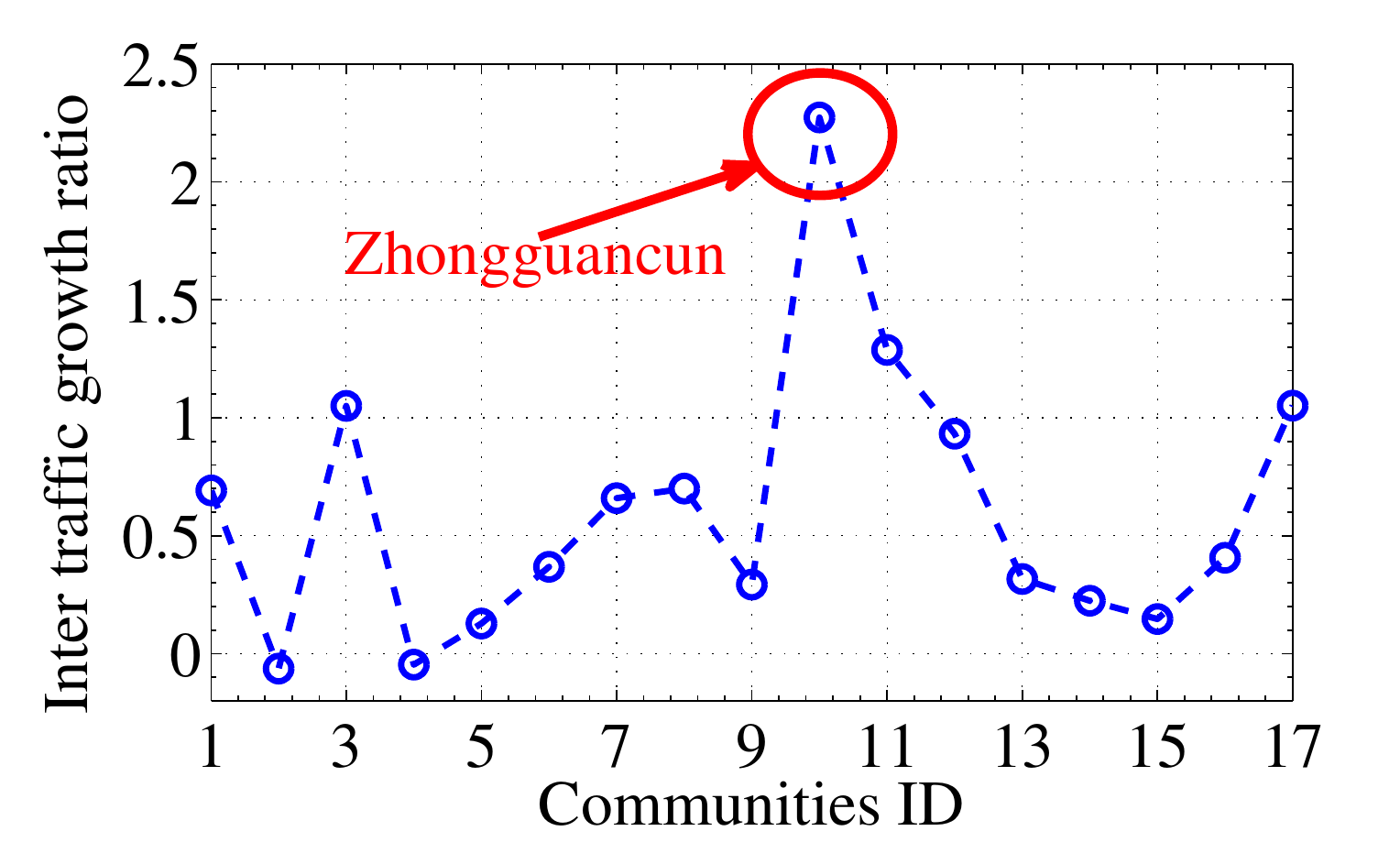}}~~~~~~
		\subfigure[Intra-Community Traffic]{\label{fig:intra_traffic}\includegraphics[width=0.52\columnwidth]{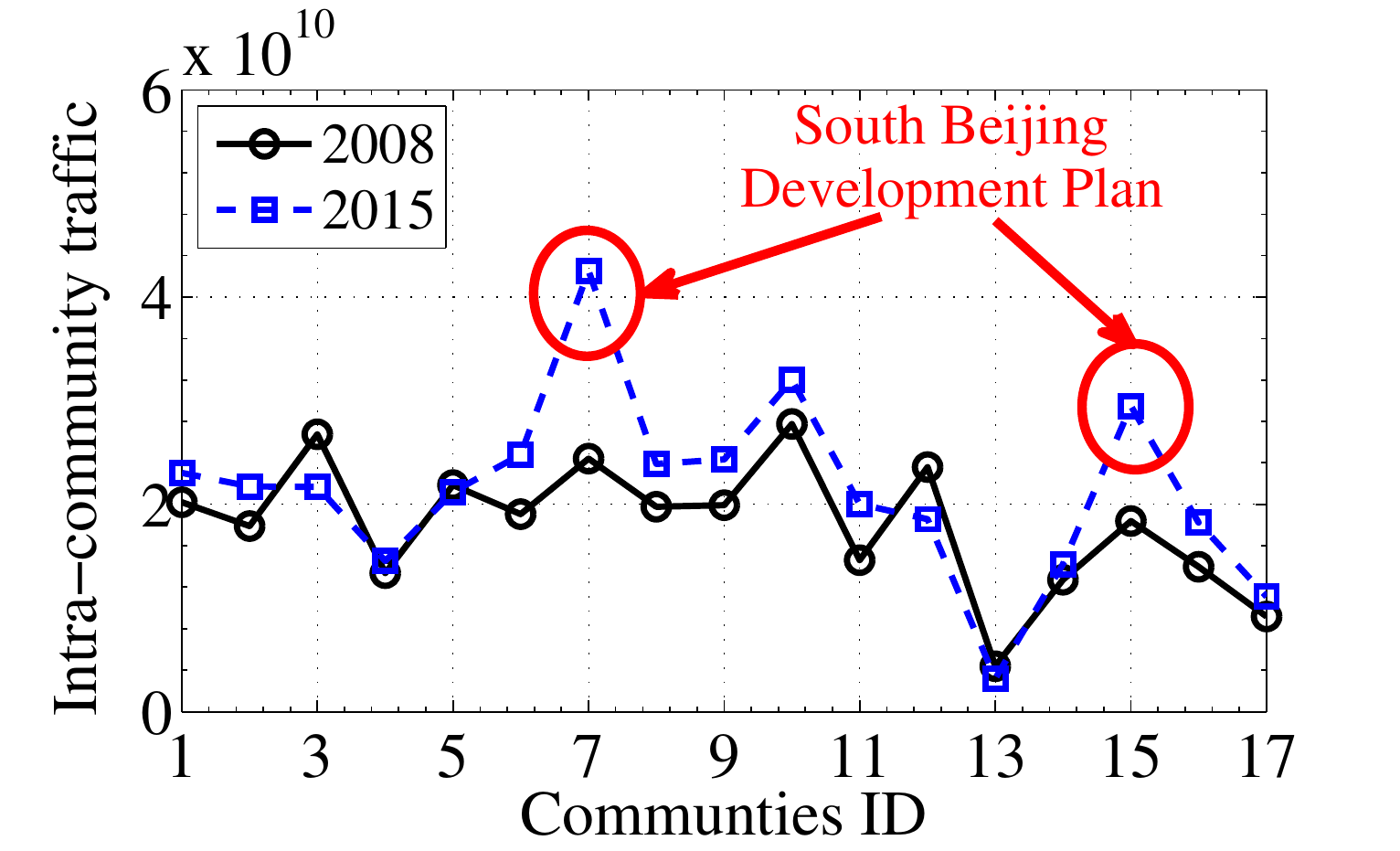}}
	\end{center}
%	\vspace{-0.3cm}
	\caption{Inter- and intra-community traffic intensities.}
	\label{fig:traffic_case}
%	\vspace{-0.2cm}
\end{figure*}

\subsection{Discovery of Urban Dynamics among Patterns}

\eat{
	\begin{figure}[t]\centering
		\begin{center}\centering
			\subfigure[Traffic comparison for morning peak]{\label{fig:13Morning} \includegraphics[width=0.45\columnwidth]{fig/Morning-eps-converted-to.pdf}}~
			\subfigure[Traffic comparison for evening peak]{\label{fig:13Evening}\includegraphics[width=0.45\columnwidth]{fig/Evening-eps-converted-to.pdf}}~
		\end{center}
		\caption{Dynamic patterns from/to the Beijing CBD community.}
		\label{fig:cbd_patterns_map}
	\end{figure}
}

Here, we use the core tensor $\tensor{C}$ to explore the urban dynamics, {\it i.e.}, the interactions among spatial and temporal patterns. We first observe the slice $\mathbf{C}_{::k}$ of $\tensor{C}$, which reveals the traffic intensity from every origin communities to every destination ones given the temporal pattern $k$, {\it i.e.}, a community level origin-destination (OD) matrix in rhythm $k$.

Fig.~\ref{fig:core_patterns} visualizes the community OD-matrices in the morning peak, midday, evening peak and night rhythms of 2008 and 2015. A darker color indicates a higher traffic intensity. As can be seen, most energies of the OD-matrices are concentrated in their diagonal lines, implying that most of taxi travels in Beijing actually happened within the same community with relatively short distances. Moreover, the travel demands across communities have a {\it tidal} phenomenon. That is, in the morning peak, people flowed out from many communities ({\it i.e.}, residential areas) and flowed in a few ones ({\it i.e.}, working areas), and the situation was just the reverse in the evening peak and night rhythms. This implies that while the residential areas in Beijing are very dispersed, the workplaces are relatively concentrated. Indeed, it seems from Fig.~\ref{fig:c015-1} that C10, C13 and C17 are the three ``most attractive'' workplaces in Beijing, which are actually well-known as the Zhongguancun area\footnote{\url{https://en.wikipedia.org/wiki/Zhongguancun}}, Beijing Central Business District (CBD)\footnote{\url{https://en.wikipedia.org/wiki/Beijing_central_business_district}}, and Beijing Financial Street\footnote{\url{https://en.wikipedia.org/wiki/Beijing_Financial_Street}}, respectively. From this aspect, NR-cNTF indeed generates high-quality patterns for urban dynamics understanding.

We then explore the evolution of traffic intensities from 2008 to 2015 in Beijing. For the comparison purpose, we first concentrate the energies of projection matrices into the core tensor as $c'_{ijk} = c_{ijk}\cdot\sum_{x} o_{xi} \cdot\sum_{y} d_{yj} \cdot\sum_z t_{zk}$. The total intensity of inter-community traffic for a community $x$ is then calculated as $I_x^{inter} = \sum_{i\neq x} \sum_{k} c'_{ixk} + \sum_{j\neq x} \sum_{k} c'_{xjk}$, and the intra-community traffic intensity  for $x$ is given by $I_x^{intra} = \sum_{k} c'_{xxk}$. Along this line, we can quantify the daily increments of inter- and intra-community traffic intensities from 2008 to 2015, as shown in Fig.~\ref{fig:traffic_case}.

\begin{figure*}[t!]\centering
	\begin{center}\centering
		\subfigure[2008 Morning Peak]{\label{fig:08M} \includegraphics[width=0.4\columnwidth]{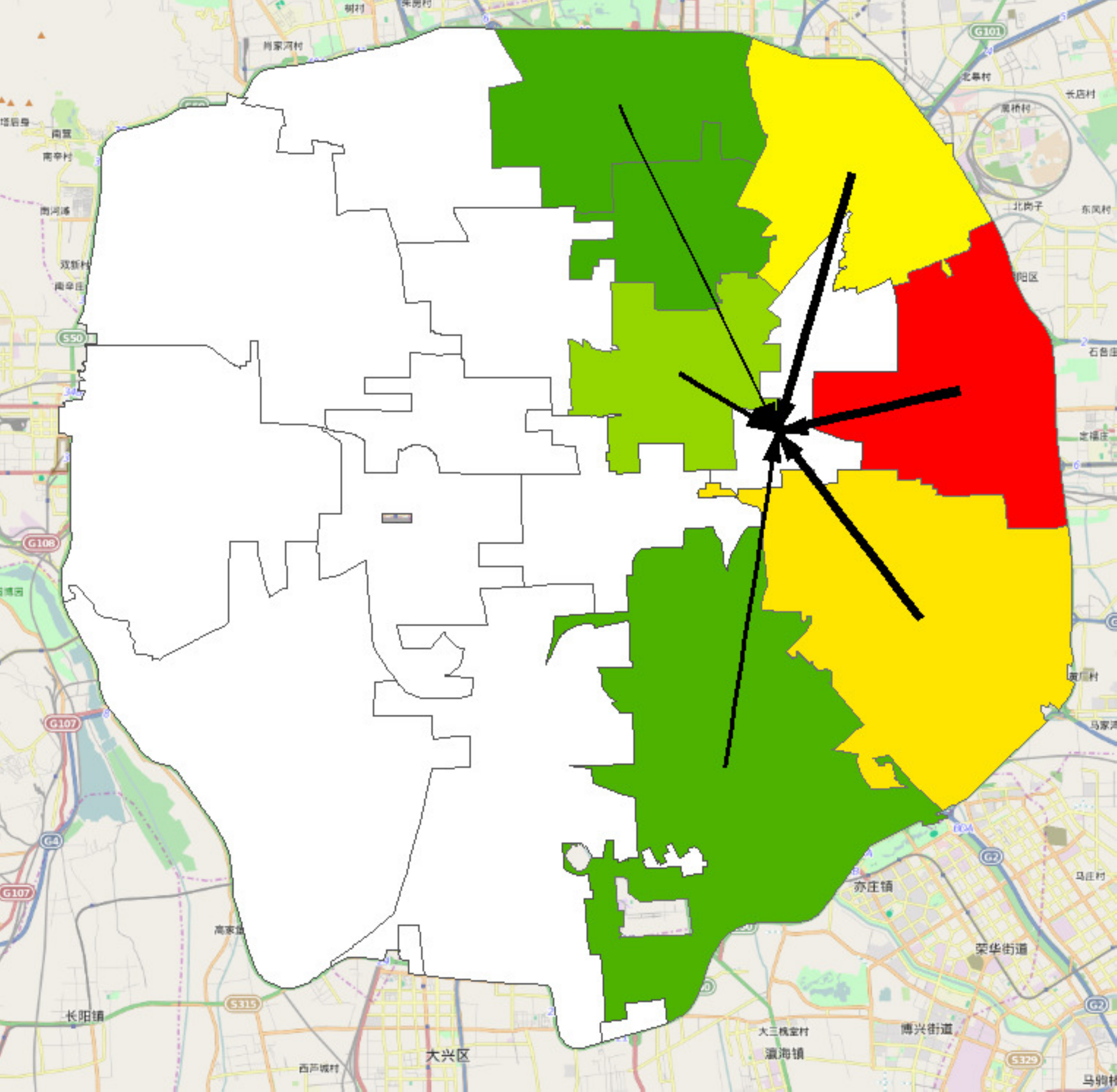}}~~~~~~
		\subfigure[2008 Evening Peak]{\label{fig:08E}\includegraphics[width=0.4\columnwidth]{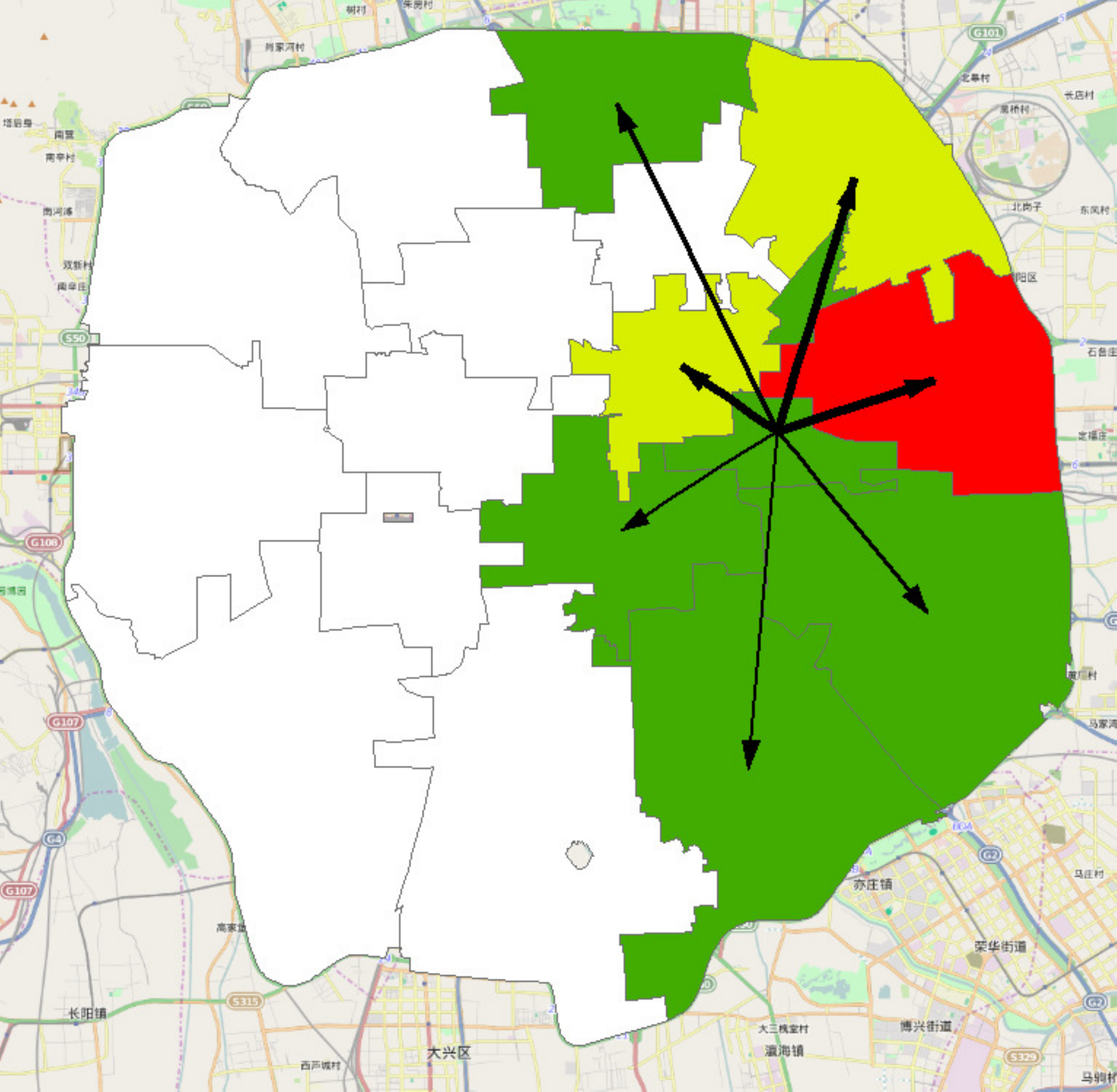}}~~~~~~
		\subfigure[2015 Morning Peak]{\label{fig:15M} \includegraphics[width=0.4\columnwidth]{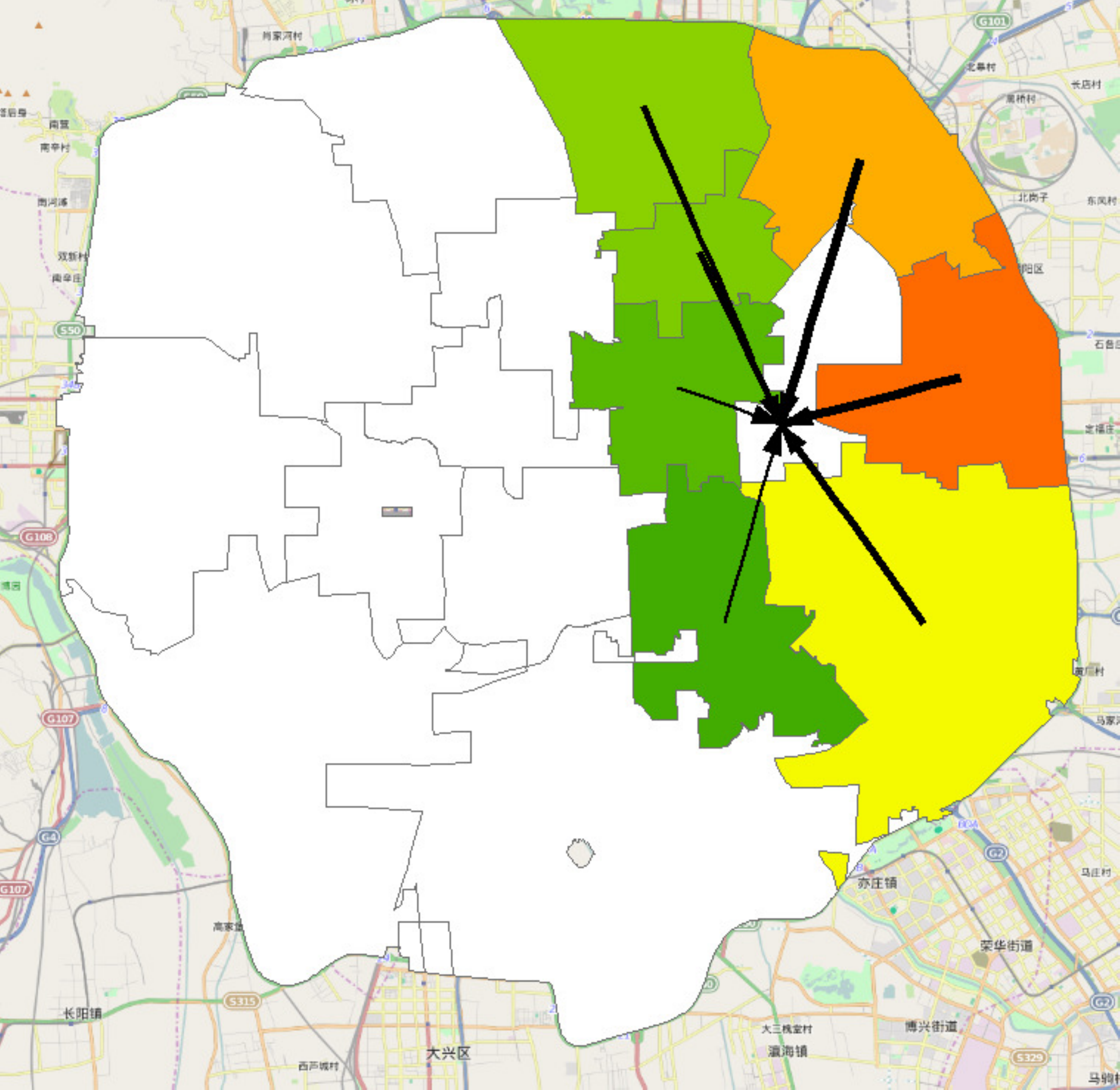}}~~~~~~
		\subfigure[2015 Evening Peak]{\label{fig:15E}\includegraphics[width=0.4\columnwidth]{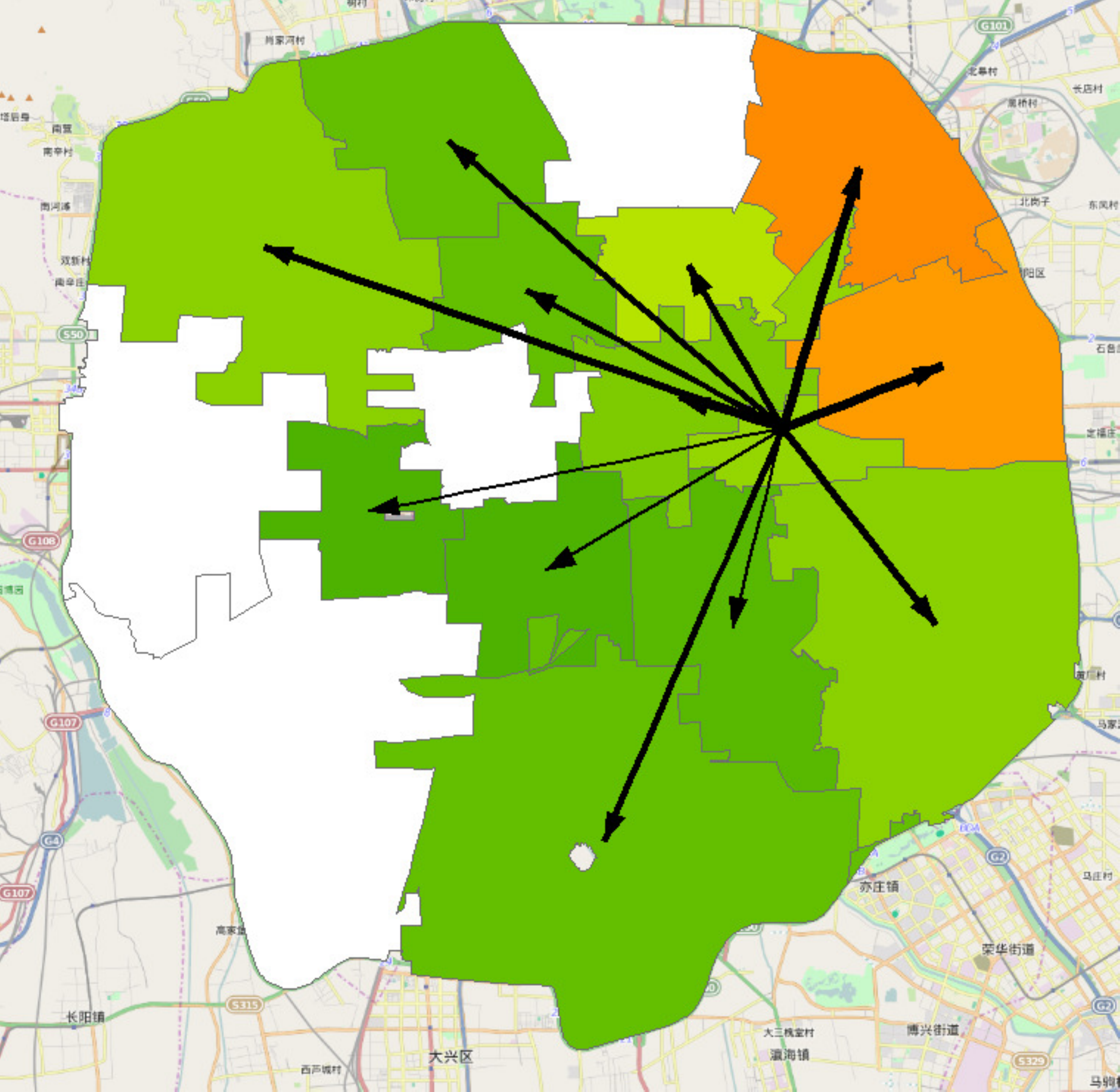}}
	\end{center}
	\vspace{-0.3cm}
	\caption{Dynamic patterns from and to the CBD community.}
	\label{fig:cbd_patterns_map}
	\vspace{-0.2cm}
\end{figure*}

From Fig.~\ref{fig:inter_traffic}, it is obvious that the inter-community traffic increased from 2008 to 2015 for almost all communities, with C10 (Zhongguancun area), C13 (CBD area) and C17 (Financial Street area) being the most significant ones. In particular, as shown in Fig.~\ref{fig:inter_growth}, the Zhongguancun area, a technology hub of Beijing and well-known as the ``Chinese Silicon Valley'', gains a highest growth ratio during the seven years, which coincides with the developing priority of Beijing with high-tech industries preference.

Fig.~\ref{fig:intra_traffic} depicts the intra-community traffic intensity of each community from 2008 to 2015. It is interesting that C7 and C15 emerged as the top-2 communities with highest growth in internal traffic. Recall that these two communities are located in the south of the Beijing city, and have benefited from the 30 billion dollar investment of the South Beijing Development Plan. The significant growth of internal traffic implies that these two communities are gaining more active economics, and perhaps are enjoying more sustainable developing pattern --- residents can work and rest interchangeably within a small distance. This indeed recommends a potential solution to mitigating the ``big city disease'' of Beijing: to promote industries and housing in a same community or close ones. This job-housing balance thinking, however, was not the primary choice of Beijing in the past several decades. The development of the CBD area, which we will discuss below, is just the epitome.

In Fig.~\ref{fig:cbd_patterns_map}, we study the dynamic patterns of a particular community: the CBD area (C13), which is the central business district of Beijing and shapes the lifestyle of the city deeply. In the figure, the color of a community indicates the traffic intensity of that community from or to the CBD community: the redder the stronger, and the arrows indicate traffic directions between communities. %For example, in the Fig~\ref{fig:08M}, the arrow to CBD with the reddest community color is from community 4, i.e. the community is east adjacent with CBD, which indicates that, in the morning peak of 2008, the strongest traffic pattern to the CBD community was from the community 4.
As shown in Fig.~\ref{fig:cbd_patterns_map}, CBD is a pure business area, with residents flowing in in the morning and flowing out in the evening. Similar situations can be found from the Zhongguancun (C10) and the Financial Street (C17) communities. This indeed reflects the severe job-housing imbalance in Beijing, which contributes a lot to the city disease such as traffic congestion. Nevertheless, it is more interesting to find the pattern evolution of CBD from 2008 to 2015. From Fig.~\ref{fig:08M} and Fig.~\ref{fig:08E}, we can find the nearly symmetric incoming and outgoing flows between the CBD community and the communities surrounding CBD in 2008. This symmetry, however, disappeared in 2015, where the outflows from CBD in the evening spread over more communities than that in the morning (see Fig.~\ref{fig:15M} and Fig.~\ref{fig:15E}). We believe it is Fig.~\ref{fig:15E} rather than Fig.~\ref{fig:15M} that revealed all the housing communities for CBD. The possible reason is, for residents living in remote communities, the {\it long-term}, {\it timely} and {\it economic} way commuting to CBD in the morning is to take metro rather than taxi. From this angle, we can conclude that the job-housing imbalance gets even worse with the rapid development of the CBD area from 2008 to 2015.

To sum up, the evolution of urban dynamics indicates the rapid development of Beijing city in recent years. The development pattern, however, is still worrying for the job-housing imbalance status quo, although the southern area has showed some positive changes.

%taxi is generally regarded as a short-trip transportation means,

%in the morning, the traffic patterns going to the CBD community were mainly from the communities surrounding CBD. In the evening peak, the traffic patterns of leaving CBD are also mainly to go to these surrounding communities. In the evening peak of 2015, the traffic patterns that flow out from CBD flow into more communities. This phenomenon implies the activity range of nightlife for residents worked in CBD increased from 2008 to 2015.

\subsection{Quantitative Evaluation}
%\subsection{Model Evaluation by Traffic Prediction}

In this subsection, we evaluate our NR-cNTF model by comparing its data tensor reconstruction error with that of some baseline models, for further explaining why NR-cNTF can work well for understanding the Beijing city. Following the tradition of tensor factorization based studies~\cite{travel_time,fema}, the {\em Root Mean Square Error} defined in Eq.~\eqref{eq:rmse} is used as an indicator of quality.

%given a quantitative evaluation as a supplement of above urban dynamics pattern discovery experiments. In the quantitative evaluation, we use {\em Root Mean Square Error} defined in~\eqref{eq:rmse} as an indicator to compare the data tensor reconstruction error of the NR-cNTF model and baseline models. It means a model could fit a urban scenario gracefully if the model could reconstruct the data tensor of the urban scenario with very low errors. This evaluation method is widely used in tensor factorization based studies~\cite{travel_time,fema}.

In the experiments, we define a sampling tensor $\tensor{S}\in \mathbb{R}^{M\times M \times N}$, in which the element $s_{xyz} = 1$ when the traffic volume form zone $x$ to zone $y$ in time slice $z$ was sampled, otherwise un-sampled. We then rewrite the objective function in Eq.~\eqref{eq:main_obj} as
\begin{equation}\label{eq:main_obj_with_S}
\begin{aligned}
\arg& \min_{\tensor{C},\mathbf{O},\mathbf{D},\mathbf{T} \geq 0}~  {\mathcal{J}}=\| \tensor{S} \odot \left( \tensor{R} - \tensor{C} \times_{o} \mathbf{O} \times_{d} \mathbf{D} \times_{t} \mathbf{T} \right) \|_{F}^2 \\
&+ \alpha \| \mathbf{W} - \mathbf{OO}^\top \|_{F}^2 + \beta \| \mathbf{W} - \mathbf{DD}^\top \|_{F}^2 \\
&+\gamma \left\|\mathbf{O}\right\|_1+\delta \|\mathbf{D}\|_1+\epsilon \|\mathbf{T}\|_1+\varepsilon\|\tensor{C}\|_1. \end{aligned}
\end{equation}
The reconstruction error between $\tensor{R}$ and the reconstructed tensor $\hat{\tensor{R}} = \tensor{C} \times_{o} \mathbf{O} \times_{d} \mathbf{D} \times_{t} \mathbf{T}$ is calculated using Eq.~\eqref{eq:rmse}.

%The element $\hat{r}_{xyz}$ in the restructured tensor $\hat{\tensor{R}} = \tensor{C} \times_{o} \mathbf{O} \times_{d} \mathbf{D} \times_{t} \mathbf{T}$ is a prediction value of $\hat{r}_{xyz}$ with ${s}_{xyz} = 0$.

We compare the reconstruction error of NR-cNTF with that of the following baseline methods:
\begin{itemize}
	\item \textbf{Tucker}: Non-negative Tucker Factorization, of which the objective function is
	\begin{equation}\small
	\begin{aligned}
	\arg& \min_{\tensor{C},\mathbf{O},\mathbf{D},\mathbf{T}} \left \| \tensor{S} \odot \left( \tensor{R} - \tensor{C} \times_o \mathbf{O} \times_d \mathbf{D} \times_t \mathbf{T}\right)  \right \|_F^2\\
	&+\gamma \left\|\mathbf{O}\right\|_1+\delta \|\mathbf{D}\|_1+\epsilon \|\mathbf{T}\|_1+\varepsilon\|\tensor{C}\|_1.
	\end{aligned}
	\end{equation}
	Compared with our method, \emph{Tucker} does not consider urban context and neighboring regularization.
	\item \textbf{CP}: Non-negative CP Factorization, which supposes a joint latent space for each mode by solving an objective function as
	\begin{equation}\small
	\begin{split}
	\arg& \min_{\mathbf{O},\mathbf{D},\mathbf{T}} \left \|\tensor{S} \odot \left( \tensor{R} - \sum_{m}{\mathbf{o}_{:m} \circ \mathbf{d}_{:m} \circ \mathbf{t}_{:m}}\right) \right\|^2_F, \\
	&+\gamma \left\|\mathbf{O}\right\|_1+\delta \|\mathbf{D}\|_1+\epsilon \|\mathbf{T}\|_1,
	\end{split}
	\end{equation}
where operator $\circ$ represents the vector outer product. In the CP factorization, the latent factor dimensionality for both the spatial and temporal patterns are the same. As a result, we set the number of latent factors $m=4$ or $m=20$. The former is the same as the number of temporal patterns for NR-cNTF, and the latter is in accordance with that of spatial patterns.
	
\item \textbf{rCP}: Regularized Non-negative CP Factorization, which is a CP factorization with the urban context-aware regularization. The objective function is
	\begin{equation}
	\begin{split}
	\arg& \min_{\mathbf{O},\mathbf{D},\mathbf{T}} \left\|\tensor{S} \odot \left( \tensor{R} - \sum_{m}{\mathbf{o}_{:m} \circ \mathbf{d}_{:m} \circ \mathbf{t}_{:m}}\right) \right\|^2_F \\
	&+\alpha\left\|\mathbf{W} - \mathbf{O}\mathbf{O}^{\top}\right\|_F^2  + \beta\left\|\mathbf{W} - \mathbf{D}\mathbf{D}^{\top}\right\|_F^2\\
	&+\gamma \left\|\mathbf{O}\right\|_1+\delta \|\mathbf{D}\|_1+\epsilon \|\mathbf{T}\|_1.
	\end{split}
	\end{equation}
\end{itemize}

In our experiments, we compared the methods on the data tensor of 2015. The sampling rate varied from 50\% to 90\%. The average $RMSE$ values of ten times repeated experiments are reported in Table \ref{table:performance}. From the table, we have the following observations:
\begin{itemize}
  \item Both NR-cNTF and cNTF performed much better than the baseline methods, indicating the general superiority of the proposed methods.
  \item NR-cNTF performed nearly the same as cNTF, indicating that the neighboring regularization improves the interpretability of spatial patterns at the very low cost of model deviation from real-world data.
  \item NR-cNTF/cNTF performed generally better than Tucker, indicating the distinct value of urban contexts for tensor factorization.
  \item NR-cNTF/cNTF/Tucker performed generally better than rCP4/CP4/rCP20/CP20, implying the advantage of employing Tucker rather than CP based methods. This is not unusual, since the core tensor generated by Tucker factorization contains important information about urban dynamic patterns and improves the model interpretability.
\end{itemize}

In summary, besides the superior interpretability, NR-cNTF also shows excellent performance in quantitative evaluation on tensor factorization, by employing core tensor, neighboring regulation, and urban contexts. As a natural corollary, NR-cNTF could be used for urban traffic volume prediction when the elements of a data tensor are only partially available.

\begin{table}\small
	%\begin{spacing}{1.5}
		%\footnotesize
		\footnotesize
		%\vspace{-0.2cm}
		\caption{Tensor Reconstruction Performance by RMSE} \label{table:performance}
		\vspace{-0.5cm}
		\begin{center}
			\begin{tabular}{c|c|c|c|c|c}
			\toprule
			&  50\%  & 60\%  & 70\% & 80\% & 90\%  \\ \midrule
			\bf{NR-cNTF}  & {0.351}	&\bf{0.344}	&\bf{0.343}  & \bf{0.342}	&\bf{0.341}	   \\
			\bf{cNTF}     & \bf{0.350}	&{0.345}	&\bf{0.343}  & \bf{0.342}	&\bf{0.341}	   \\
			Tucker        & 0.357     	&0.356	&0.353  & 0.351	&0.350	  \\
			rCP-20         & 0.351	&0.349	&0.349  & 0.347	&0.347	   \\
			rCP-4          & 0.403	        &0.401	&0.400  &0.398 	&0.396 	   \\
			CP-20          & 0.353	&0.352	&0.349  & 0.348	&0.346   \\
			CP-4           & 0.405	&0.403	&0.401  & 0.401	&0.400	   \\\bottomrule
			\end{tabular}
		\end{center}
%		\vspace{-0.3cm}
%	\end{spacing}
\end{table}

%%%%%%%%%%%%%%%%%%%%
%{\color{red}\bf End my modification --- Jingyuan 20170507}
%%%%%%%%%%%%%%%%%%%%

\section{Related Work}
\label{sec:relatedwork}

%\subsection{Mining of Urban Human Mobility Data}

Mining knowledge from human mobility data generated in urban areas has attracted many researchers' interests in recent years~\cite{urban_computing,pangang_survey}. Various types of ``social sensors'', such as cell phones~\cite{cellphone}, GPS terminals~\cite{pangang_survey}, and smart bus/metro cards~\cite{pnas1}, have been adopted to record mobility information of urban residents, based on which many successful applications have emerged for intelligent transportation~\cite{tdrive,fcs_bike}, environmental protection~\cite{uair}, urban planning~\cite{planning}, urban emergency~\cite{song2013modeling}, {\it etc}. An excellent survey from an urban computing perspective can be found in~\cite{urban_computing}, while \cite{pangang_survey} provides a survey from a social and community dynamics perspective.

Among the abundant methods for human mobility data mining, tensor factorization/decomposition, like CANDECOMP/PARAFAC (CP)~\cite{CP} and Tucker factorizations~\cite{tucker}, gains particular interests for its distinct ability in modeling multi-aspect heterogeneous big data. Indeed, in city scenarios data samples are always involved with many aspects, such as time, space, human, urban contexts and so on, and therefore are very suitable for tensor factorization based data mining methods~\cite{urban_computing}. Typical applications of tensor factorization could be classified into two categories. The first category is to reconstruct tensors for predicting unknown values in multi-aspect data sets, such as completing missing traffic data~\cite{tensor_its}, inferring urban gas consumption~\cite{gas_consum}, predicting travel time~\cite{travel_time}, recommending social tags~\cite{unified_recom}, movies~\cite{movie_recom} and sightseeing locations~\cite{UCLAF,UCLAF_J}, and so on.

In recent years, more and more works focused on mining explainable latent factors from multi-aspect urban data sets, which form the second category of applications. The focal point here is to use tensor factorization to discover latent lower-dimensional factors from higher-dimensional multi-aspect data sets. For instance, Metafac~\cite{Metafac} used CP factorizations to extract latent community structures from various social networks, and~\cite{multiview} proposed a multi-view data clustering and partitioning method based on Tucker factorization. Our study in this paper also falls in this category, with some most related works as follows.

The study~\cite{PLOSONE_matrix} used a non-negative matrix factorization, {\it i.e.}, a second-order tensor factorization, to model taxi trip data, and discovered the latent factors corresponding to three rhythms of resident's daily life. Similarly, matrix factorizations were used for understanding the operational behaviors of taxicabs in cities~\cite{Kang2016Understanding}. In the inspiring work, \cite{apweb} adopted a regularized non-negative Tucker decomposition (rNTD) to discover residents' mobility patterns in Beijing from an origin-destination-time tensor. Following this idea, \cite{TRB} proposed a probabilistic tensor factorization method to find mobility patterns of public transaction system passengers from an origin-destination-time-type tensor. CitySpectrum~\cite{cityspectrum} used CP factorizations to mine joint time-day-location patterns of residents after the Great East Japan Earthquake. Some more complex algorithms include NTCoF~\cite{ntcof}, which is a non-negative tensor co-factorization algorithm for urban events detection from bike trip and check-in data, and HTM~\cite{HTM}, which is a hybrid tensor model and uses ACS-tucker decomposition to detect events from traffic data. In recent years, many dynamic tensor factorization algorithms were proposed for time series and stream data mining. For instance, Dynamic Tensor Analysis~\cite{DTA} extended Tucker factorization to process dynamic and stream high-order data, the Facets model~\cite{facets} combined dynamic graphical models with tensor factorizations for mining co-evolving high-order time series, and FEMA~\cite{fema} was a flexible evolutionary tensor factorization algorithm to mine dynamic behavioral patterns of multi-facet data sets.

Despite of the wide existence of related works mentioned above, our study in this paper has its own uniqueness. Unlike the previous works, we focus on understanding urban dynamics from multiple aspects, including spatial, temporal, as well as spatio-temporal interactions, with still a pursue to long-term evolution patterns. The results indeed bring some important managerial insights and suggestions to city development of Beijing. The proposed NR-cNTF model takes Tucker factorization as a basic framework, which compared with CP and matrix factorization based models~\cite{PLOSONE_matrix,Kang2016Understanding,cityspectrum,HTM} has better interpretability for adopting a core tensor to model relations among latent factors. Compared with the existing Tucker factorization based methods~\cite{TRB,tensor_its,urban_computing}, NR-cNTF incorporates urban contexts and neighboring regulation, which improve both the accuracy and interpretability of Tucker factorization greatly. Moreover, we proposed a pipeline initialization approach to analyze the evolution of urban dynamics across several years, which is simple yet practical.

\section{Conclusion}
\label{sec:conclusion}
In this paper, we proposed a POI context-aware nonnegative tensor factorization model with neighboring regulation (NR-cNTF) for urban dynamics discovery. A simple pipeline initialization method was also introduced to NR-cNTF to facilitate evolution analysis of the dynamics. Experiments on Beijing taxi trajectory and POI data demonstrated the high-quality of the spatial, temporal and spatio-temporal patterns generated by NR-cNTF for city-disease diagnosing and urban planning. The comparative studies with some baselines on traffic prediction further justified the advantage of NR-cNTF in adopting urban contexts and neighboring regulation.

%\section*{Acknowledgments}
%
%Dr. J. Wang's work was partially supported by the National Key Research
%and Development Program of China (No.2016YFC1000307), the National Natural Science Foundation of China (NSFC) (61572059, 61202426) and the Science and Technology Project of Beijing. Prof. J. Wu's work was partially supported by the National Natural Science
%Foundation of China (NSFC) (71531001, 71725002, U1636210). Mr. Z.Wang's work was partially supported by the Science and Technology Project of Beijing (Z181100003518001).

\bibliography{tensor}
\bibliographystyle{IEEEtran}

\newpage

\end{document}